\documentclass[runningheads]{llncs}

\usepackage{eccv}

\usepackage{eccvabbrv}

\usepackage{graphicx}
\usepackage{booktabs}
\usepackage{multirow}

\usepackage[accsupp]{axessibility}  

\usepackage[pagebackref,breaklinks,colorlinks,citecolor=eccvblue]{hyperref}

\usepackage{orcidlink}

\begin{document}

\title{S3-SLAM: Sparse Tri-plane Encoding for Neural Implicit SLAM} 

\titlerunning{S3-SLAM: Sparse Tri-plane Encoding for Neural Implicit SLAM}


\author{Zhiyao Zhang$^{1}$ \and
Yunzhou Zhang$^{1}$\thanks {Corresponding author}
\and Yanmin Wu$^{2}$ \and 
Bin Zhao$^{1}$ \and
Xingshuo Wang$^{1}$ \and
Rui Tian$^{1}$}

\authorrunning{Zhiyao Zhang \and
Yunzhou Zhang \and Yanmin Wu et al.}

\institute{$^1$Northeastern University \quad $^2$Peking University
}

\maketitle


\vspace{-6mm}
\begin{abstract}
  With the emergence of Neural Radiance Fields (NeRF), neural implicit representations have gained widespread applications across various domains, including simultaneous localization and mapping. However, current neural implicit SLAM faces a challenging trade-off problem between performance and the number of parameters. To address this problem, we propose sparse tri-plane encoding, which efficiently achieves scene reconstruction at resolutions up to 512 using only 2$\sim$4\% of the commonly used tri-plane parameters (reduced from 100MB to 2$\sim$4MB). On this basis, we design S3-SLAM to achieve rapid and high-quality tracking and mapping through sparsifying plane parameters and integrating orthogonal features of tri-plane. Furthermore, we develop hierarchical bundle adjustment to achieve globally consistent geometric structures and reconstruct high-resolution appearance. Experimental results demonstrate that our approach achieves competitive tracking and scene reconstruction with minimal parameters on three datasets. \textbf{Source code will soon be available.}
  \vspace{-4mm}
\end{abstract}

\section{Introduction}
\label{sec:intro}
Dense RGB-D SLAM is a fundamental task in computer vision, playing a crucial role in applications such as augmented reality, scene editing, robot navigation, and intelligent agent positioning and planning. Traditional RGB-D SLAM\cite{mur2015orb,mur2017orb2,schops2019bad} relies on manually designed features, making it difficult to learn the inductive biases of scenes in a data-driven manner and thus unable to reconstruct scenes with infinite resolution. Neural implicit SLAM, drawing inspiration from the volume rendering of NeRF and spatial mapping encoding, has gained popularity for reconstructing high-fidelity and high-resolution scenes using only multi-layer perceptrons (MLP).

Existing neural implicit SLAM\cite{sucar2021imap,zhu2022nice,yang2022voxfusion,johari2023eslam,wang2023coslam,sandstrom2023pointslam} excels in reconstructing high-quality scenes and accurately predicting camera poses. However, current state-of-the-art approaches face a trade-off problem between the number of parameters and model performance. For instance, methods like Vox-Fusion\cite{yang2022voxfusion} and Co-SLAM\cite{wang2023coslam} have fewer parameters but lose some detailed appearance information. Alternatively, methods such as NICE-SLAM\cite{zhu2022nice}, ESLAM\cite{johari2023eslam}, and Point-SLAM\cite{sandstrom2023pointslam} require more parameters and face challenges related to fast convergence. Notable efforts to enhance the efficiency of neural implicit representations include techniques such as sparse voxel octrees\cite{yang2022voxfusion}, tri-planes\cite{gao2022get3d}, dense grids\cite{sun2022dvgo,zhu2022nice}, hash grids\cite{muller2022instantngp}, and tensor decomposition\cite{chen2022tensorf,gao2023strivec}. However, these methods still experience a rapid increase in parameter count when dealing with resolutions higher than 512.


\begin{figure*}[t]
  \centering
  \captionsetup{type=figure}
  \includegraphics[width=0.96\linewidth]{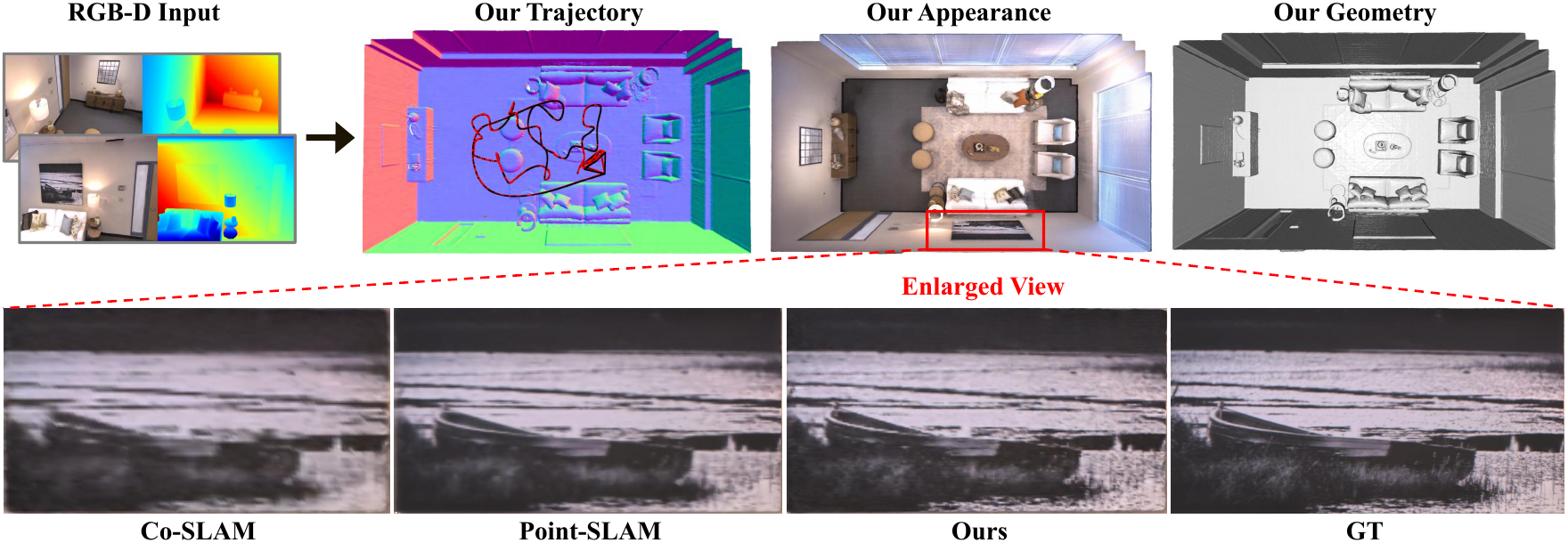}
  \vspace{-2mm}
  \captionof{figure}{We propose S3-SLAM, a neural implicit SLAM that applies our designed sparse tri-plane encoding. S3-SLAM can estimate accurate camera poses and reconstruct high-fidelity scenes with minimal memory footprint. The parameter count of our sparse tri-plane encoding is only 2$\sim$4\% of the tri-plane\cite{gao2022get3d} (reduced from 100MB to 2$\sim$4MB).}
  \vspace{-6mm}
  \label{fig:cover}
\end{figure*}%

In the design of neural implicit SLAM scene representations, NICE-SLAM\cite{zhu2022nice} successfully reduces the parameter count by employing a hierarchical feature grid but still faces significant computational overhead. 
By employing joint encoding to sparsify the scene representation, Co-SLAM\cite{wang2023coslam} can reduce the parameter count but sacrifice the fine-level appearance of the scene.
ESLAM\cite{johari2023eslam} utilizes tri-plane encoding for accurate camera pose estimation and high-quality mapping. However, the high-resolution scene representation of the tri-plane results in a huge parameter count, making it difficult to rapidly learn the induction bias of detailed appearance in scenes. Point-SLAM\cite{sandstrom2023pointslam} can reconstruct the fine-level appearance, but the parameter count of neural point cloud features gradually increases during SLAM running. 
We believe that neural implicit SLAM requires a novel scene representation that can comprehensively balance efficiency and parameter number to improve SLAM performance further and reduce memory footprint. Thus, we pose the question: \textbf{How to reconstruct fine-level scene geometry and appearance while estimating camera pose with minimal parameters?}

To address the challenges above, we propose S3-SLAM, leveraging our novel sparse tri-plane encoding to achieve rapid iteration and parameter sparsity. 
As shown in \cref{fig:cover}, our approach achieves the high-resolution scene reconstruction and accurate pose estimation in a few iterations. Our key idea is to employ spatial hash function to maintain sparse 2D hash-grid plane features at minimal memory footprint. By introducing the orthogonal projection properties of tri-plane, we achieve a compact and global scene representation. Additionally, we adopt neural networks to learn inductive biases of scenes, enabling interpolating unseen views and high-fidelity reconstruction from our sparse tri-plane encoding. Furthermore, we observe that existing bundle adjustment in neural implicit SLAM does not adequately consider global and local geometric consistency. Consequently, we introduce a hierarchical bundle adjustment (HBA) method to preserve global geometric consistency while achieving the reconstruction of high-resolution local appearance and geometry. 

The main contributions can be summarized as follows:
\begin{itemize}
\item 
We propose a compact and efficient sparse tri-plane encoding to address the trade-off challenges in neural implicit representations between parameter quantity and reconstruction quality. Sparse tri-plane encoding significantly reduces model memory consumption by sparsifying orthogonal plane features using hash-grids.
\item 
We develop S3-SLAM, a neural implicit SLAM that applies our sparse tri-plane encoding, enabling accurate camera pose estimation and scene reconstruction with minimal iteration. For higher-resolution representation, we only require 2$\sim$4\% of the parameters in regular tri-plane encoding by introducing multi-resolution sparse tri-plane to represent complex scenes.
\item
We design hierarchical bundle adjustment (HBA) in S3-SLAM to refine local appearance and ensure global geometry consistency, enabling high-quality appearance reconstruction and accurate pose estimation.
\item
Experimental results demonstrate that S3-SLAM achieves accurate and robust camera tracking with minimal parameters and iterations, simultaneously achieving high-fidelity scene reconstruction. 
\end{itemize}

\section{Related Work}

\textbf{Neural Radiance Fields.}
NeRF\cite{mildenhall2020nerf} utilizes positional encoding\cite{tancik2020fourier} to enable neural networks to acquire high-frequency information and inductive biases of scene appearance for synthesizing photorealistic novel views using volume rendering. 
Subsequent works address aliasing artifacts\cite{barron2021mip}, poor rendering quality in unbounded scenes\cite{barron2022mip, tancik2022block}, or blurry rendering in dynamic scenes\cite{li2021nsff, pumarola2021dnerf, gao2021dynamic, peng2023dynamaps}.
Due to the absence of constraints on surface geometry, the above density-based methods cannot produce high-quality scene geometry. Some neural implicit surface reconstruction approaches\cite{wang2021neus, yariv2021volsdf, yu2022monosdf, ren2023volrecon} address the challenge of scene geometry generation by introducing additional constraints such as SDF, depth, and normals.

\noindent\noindent
\textbf{Neural Scene Representations.} 
Since the emergence of NeRF, the exploration of neural scene representations have been greatly promoted. Neural scene representations aim to efficiently encode the scene, enabling NeRF to reconstruct scenes faster and more effectively. Subsequent works utilize spatial data structures like grids\cite{sun2022dvgo, fridovich2022plenoxels}, hash tables\cite{muller2022instantngp}, and octrees\cite{liu2020nvsf} to store 3D sampled point features, reducing queries of neural networks and accelerating training of NeRF. Additionally, TensoRF\cite{chen2022tensorf} and Strivec\cite{gao2023strivec} employ tensor decomposition to transform dense grids into linear combinations of low-rank vectors and matrices, significantly reducing parameters. In contrast to prior excellent research, our study sparsifies tri-plane and integrates multi-level orthogonal features to achieve sparse tri-plane encoding, resulting in a more compact representation to reconstruct complex scenes.

\noindent\noindent
\textbf{Neural Implicit SLAM.} 
Recently, the development of neural implicit representations has led to the popularity of NeRF in SLAM. RGB-D neural implicit SLAM approaches\cite{sucar2021imap, zhu2022nice, johari2023eslam} use NeRF or advanced neural implicit representations to achieve consistent geometry and accurate pose estimation. Point-SLAM\cite{sandstrom2023pointslam} encodes scene features with point clouds, achieving fine-grained geometry and color reconstruction. Co-SLAM\cite{wang2023coslam} adopts coordinated and parametric encoding, creating a coherent prior for the MLP to reconstruct a complete scene geometry. Compared to these outstanding methods, our approach utilizes our novel scene representation to achieve higher-resolution appearance and geometry reconstruction, significantly reducing memory footprint while enhancing performance. We design global-to-local HBA optimization while ensuring global geometric consistency to reconstruct local scene appearance and camera tracking more accurately.

\section{Compact and Efficient Scene Representation}
Tri-plane\cite{gao2022get3d} demonstrates remarkable proficiency in representing geometric consistency but faces challenges in cubical memory growing due to dense feature layouts, limiting its effectiveness in SLAM. Therefore, we propose a sparse tri-plane encoding, which achieves compact and efficient axis-aligned feature plane representations of the scene by sparsifying plane parameters. Furthermore, the neural implicit representation of S3-SLAM consists of multi-resolution sparse tri-plane encoding and MLP decoders. 
Moreover, we suggest employing sparse tri-plane encoding to represent geometry and appearance separately, thereby minimizing the forgetting issue caused by the higher frequency of neural network updates in color compared to geometry.

\subsection{Sparse Tri-plane}
\begin{figure}[!t]
  \centering
   \includegraphics[width=0.55\linewidth]{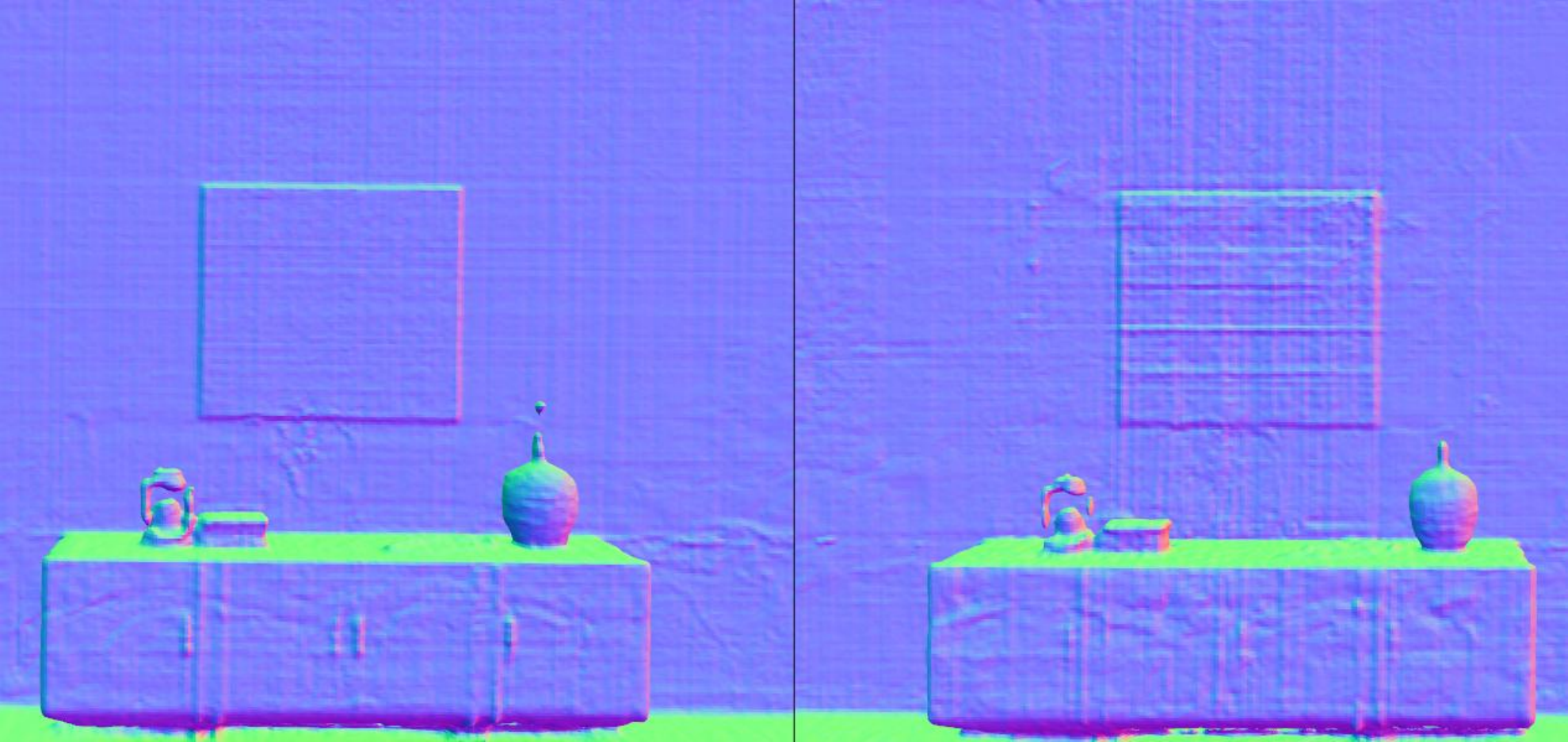}

   \caption{
  In our design, the hash-grid\cite{muller2022instantngp} (left) generates smoother and more coherent surface geometry than permutohedral lattice-grid\cite{rosu2023permutosdf} (right).}
  \vspace{-6mm}
   \label{fig:hp}
\end{figure}
We set up three mutually orthogonal planes, project 3D points $p$ onto these planes, and then apply sparse parametric encoding to these projected points. Currently, multi-resolution hash encoding\cite{muller2022instantngp} and permutohedral lattice encoding\cite{rosu2023permutosdf} are popular techniques in the sparse parametric encoding. As shown in \cref{fig:hp}, multi-resolution hash encoding can generate smoother surfaces in our sparsification design. 
Therefore, we represent projection planes as 2D square hash grid planes $H_{xy}$, $H_{xz}$, $H_{yz}$. When hashing the indices of grid plane vertices $\mathbf{x}$, we obtain the vertex indices $h_\mathbf{x}$ through the hash function:
\begin{equation}
    h_\mathbf{x} = \left(\bigoplus\limits_{i=1}^2 x_i \pi_i \right) \hspace{1mm} mod \hspace{1mm} 2^T
    \label{eq:eq1}
\end{equation}
where $\pi_1=1$, $\pi_2=2654435761$,  $T$ is a hyperparameter ranging from 14 to 24, $x_i$ represents the coordinate value of the $i$-th dimension of grid vertices $\mathbf{x}$.
The spatial hash function maps vertex indices to a hash table of length $2^T$, limiting the maximum number of vertices requiring updates and achieving sparse single-plane encoding. Specifically, our sparse tri-plane encoding encodes the projected points $p_{xy}$, $p_{xz}$, $p_{yz}$ of the 3D point 
$p$ through 2D hash grid planes. Afterwards, we concatenate the encoded features to obtain the final sparse tri-plane encoding $f_\theta$:\looseness=-1
\begin{equation}
    \begin{split}
        f_\theta(p) = H_{xy}(p_{xy}) \oplus H_{xz}(p_{xz}) \oplus H_{yz}(p_{yz})
    \end{split}
    \label{eq:eq2}
\end{equation}

Moreover, owing to the efficient feature maintenance by hash tables, we successfully enhance the compactness of sparse tri-plane encoding while achieving sparse feature representation of the tri-plane. Sparse tri-plane encoding makes capturing high-frequency geometry and appearance easier for the neural network.

\subsection{Multi-resolution Sparse Tri-plane}
\begin{figure*}[ht]
  \centering
  \includegraphics[width=1\linewidth]{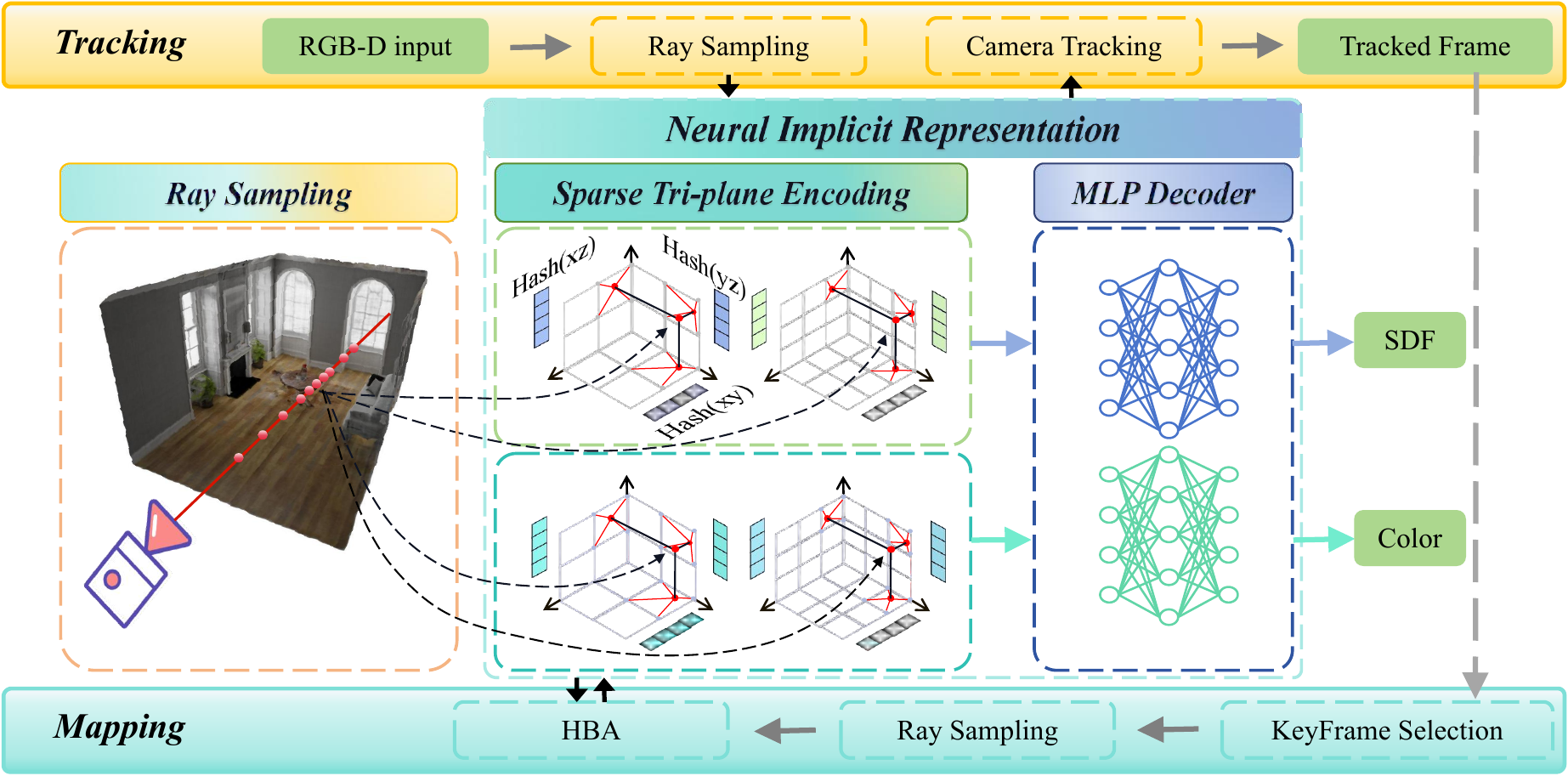}

  \caption{Our approach consists of neural implicit representation, tracking, and mapping. Given RGB-D input stream, S3-SLAM utilizes predicted poses and ray sampling to obtain 3D sampled points. Our sparse tri-plane encoding hashes multi-resolution plane features of sampled points to represent geometry and appearance compactly and efficiently. We represent $H_{xy}$, $H_{xz}$, and $H_{yz}$ of \cref{eq:eq2} using Hash(xy), Hash(xz), and Hash(yz). Our decoder decodes these encodings into SDF and color. During tracking and mapping, we adopt re-rendering loss of \cref{eq:loss} to update parameters of S3-SLAM.}
  \vspace{-4mm}
  \label{fig:framework}
\end{figure*}
We design a multi-resolution sparse tri-plane encoding to represent complex scenes more effectively. Inspired by multi-resolution hash encoding, as shown in \cref{fig:framework},  we construct multi-resolution tri-planes using multi-level 2D hash grid planes and a resolution growth factor. Subsequently, we apply our sparse tri-plane encoding to each tri-plane and concatenate the resulting features to achieve a multi-resolution sparse tri-plane encoding. The approach enables our scene representation to incorporate multi-level features, effectively representing fine-level appearance and geometry.  
To reconstruct a high-quality scene, we employ two multi-resolution sparse tri-plane encodings to encode the scene geometry and appearance, sacrificing some speed but ensuring more accurate appearance reconstruction. 

\subsection{Decoder Architecture} 
We employ tiny MLPs as our decoders due to sparse parameters of our representation. Our decoder architecture consists of SDF decoder and color decoder. The SDF decoder with two hidden layers containing 32 neurons and ReLU as the non-linear activation function. Similarly, we use the same configuration for the color decoder, with the only difference being adding a Sigmoid activation function as the output layer. The application of tiny MLPs ensures that our method does not significantly increase the parameter count. The decoder decodes our multi-resolution sparse tri-plane encoding into appearance and accurate TSDF representation.

\section{Color and Geometry Rendering}
Given the camera center $\mathbf{o}$ and direction $\mathbf{d}$, we obtain sampled points along the ray $\mathbf{r}=\mathbf{o}+t_i \mathbf{d}, i \in [1,M]$. Regarding selecting sampling distance $t_i$, NeuS\cite{wang2021neus} employs a hierarchical sampling approach for precise surface reconstruction but comes with significant computational overhead. Instant-NGP\cite{muller2022instantngp} adopts uniform sampling and inverse space transform to adapt in the unbounded scene, but making it unsuitable for reconstructing surface mesh. We strike a balance between these two sampling methods and adopt depth-guided sampling.

\subsection{Depth-guided Sampling} 
We consider depth observations $D^{gt}$ as the surface and uniformly sample two-thirds of the points near the surface in the range of $[D^{gt}-1.2\times tr, D^{gt}+1.2\times tr]$ to ensure sufficient training near the surface. We denote the truncated distance of the TSDF as $tr$, which is a hyperparameter. We also apply uniform sampling to the remaining one-third of the rays to reduce artifacts in empty regions.

\subsection{Neural RGB-D Loss Function}
To obtain color $\mathbf{C_r}$ and depth $D_r$ of the ray $\mathbf{r}$, we employ the rendering functions:
\vspace{-5mm}
\begin{equation}
    \begin{split}
        w_i = \sigma(-\frac{s_i}{tr})&\sigma(\frac{s_i}{tr}), \quad 
        \mathbf{C_r} = \frac{1}{M} 
        \sum_{i=1}^{M} w_i \cdot \mathbf{c_i}, \quad D_r = \frac{1}{M} 
        \sum_{i=1}^{M} w_i \cdot t_i
        \label{render}
    \end{split}
\end{equation}
where $\sigma$ is the Sigmoid function, 
color $\mathbf{c_i}$ and SDF $s_i$ of $i$-th sampled point on the ray $r$ are predicted by decoder. We randomly sample the $N$ rays. To obtain the set of rays $R$ with valid depth observations, we filter out sampled rays where the ground truth depth is either zero or excessively large, thereby reducing the impact of depth noise. We apply the Mean Square Error (MSE) between rendering (color $\mathbf{C}_{j}$, depth $D_{r}$ and SDF $s_p$) and ground truth as the loss functions: 
\vspace{-2mm}
\begin{equation}
    \begin{split}
    \mathcal{L}_{c} = \frac{1}{N} 
    \sum_{j=1}^{N}&
    \left ( \mathbf{C}_{j} - \mathbf{C}_{j}^{gt} \right )^2 , \quad \mathcal{L}_{d} = \frac{1}{\left | R \right |} 
    \sum_{r \in R}^{}
    \left ( D_{r} - D_{r}^{gt} \right )^2 \\
    \mathcal{L}_{sdf}& = \frac{1}{\left | R \right |} 
    \sum_{r \in R}^{}
    \frac{1}{|{S}_{r}^{tr}|} 
    \sum_{p \in {S}_{r}^{tr}}^{}
    \left ( s_p - (D_{r}^{gt} - t_p) \right )^2 \\
    &\mathcal{L}_{fs} = \frac{1}{\left | R \right |} 
    \sum_{r \in R}^{}
    \frac{1}{|{S}_{r}^{fs}|}
    \sum_{p \in {S}_{r}^{fs}}^{}
    \left ( s_p - tr \right )^2 
    \end{split}
    \label{eq:loss}
\end{equation}
where $\mathbf{C}^{gt}_j$ and $D^{gt}_r$ represents the color and depth ground truth. ${S}_{r}^{tr}$ is truncated surface region $(\left | D_{r}^{gt} - t_p \right | \le tr)$, ${S}_{r}^{fs}$ is the region of the camera origin center to the positive truncated surface $(D_{r}^{gt} - t_p > tr)$.
We design final loss function as follows:
\begin{equation}
    \begin{split}
        \mathcal{L} = \lambda_{c} \mathcal{L}_{c} + \lambda_{d} \mathcal{L}_{d} + 
        \lambda_{sdf} \mathcal{L}_{sdf} +
        \lambda_{fs} \mathcal{L}_{fs}
        \label{render}
    \end{split}
\end{equation}

\section{Tracking and Mapping}
\label{sec:blind}

\subsection{Camera Tracking}
Our tracking operates independently without relying on Bundle Adjustment (BA). Tracking utilizes a constant velocity motion model to initialize camera pose. As \cref{eq:loss}, our tracking minimizes the Mean Squared Error between rendered and ground truth images and diminishes the shortest distance between the RGB-D point cloud and the geometric surface (equivalent to the predicted SDF) through gradient backpropagation for updating the camera pose.

\subsection{Keyframe Selection Strategy}
Given the predicted current camera pose $\mathbf{T_{cur}}$, the pose of the previous reference keyframe $\mathbf{T_{ref}}$, and the camera intrinsics $\mathbf{K}$, we determine the keyframe sequence based on the projection relationship between image frames. Specifically, we back-project the pixels $u_{cur}$ of the current frame into the world coordinate system and then project the 3D points onto the previous reference keyframe pixels $u_{ref}$.
\begin{equation}
    \begin{split}
    u_{ref} = \mathbf{K}\mathbf{T_{ref}}\mathbf{{T_{cur}}^{-1}}\mathbf{K^{-1}}u_{cur}
    \end{split}
    \label{eq:ekeyframe_eq}
\end{equation}
Subsequently, we exclude points not within the range of normalized image coordinates. If the proportion of points successfully projected onto the pixel plane of the reference keyframe falls below our selected threshold, we identify the current frame as a keyframe.

\subsection{Hierarchical Bundle Adjustment}
We design a hierarchical bundle adjustment (HBA) method to enhance appearance quality, ensuring global structural consistency while reconstructing high-resolution scene appearance. In every iteration of HBA, we sample a small number of rays across all keyframes to maintain global consistency. For ensuring local consistency, we establish a local sliding window, and sample more rays from keyframes within sliding window to thoroughly estimate local camera poses and reconstruct fine-level scene geometry and appearance. In our S3-SLAM, the global sample rays constitute 10\% of the overall sampled rays, while the local sample rays are determined by weighting each keyframe based on its best loss. We ensure that the weighted sum of losses for all keyframes within the sliding window is normalized to one. Furthermore, to prevent our sampling strategy from being overly influenced by training losses, potentially causing local divergence, we set a minimum proportion of sampled rays for each frame at 10\% of the total rays. The strategy ensures that frames with higher losses receive more training emphasis, enhancing the precision of local appearance details while mitigating the impact of training divergence on the sampling strategy.

\section{Results}

\subsection{Experimental Setup}
\textbf{Datasets.}
We evaluate our S3-SLAM method on three datasets. Firstly, one synthetic dataset, Replica\cite{straub2019replica}, comprise 8 scenes, respectively. Additionally, two real datasets, ScanNet\cite{dai2017scannet} and TUM RGB-D\cite{sturm12tum}, include 6 and 3 scenes, respectively, with significant rotations and depth noise. The ground truth poses for ScanNet are derived from BundleFusion\cite{dai2017bundlefusion}, while those for TUM RGB-D are obtained from a motion capture system.

\noindent\noindent
\textbf{Baselines.}
To assess the performance of our method in scene reconstruction and camera tracking, we primarily use NICE-SLAM\cite{zhu2022nice}, Co-SLAM\cite{wang2023coslam}, and Point-SLAM\cite{sandstrom2023pointslam} as baselines. Furthermore, we evaluate all baselines using our hardware device for a fair performance evaluation.

\noindent\noindent
\textbf{Metrics.}
For reconstruction quality evaluation, we evaluate our method on Replica\cite{straub2019replica} using the strategy proposed by Co-SLAM\cite{wang2023coslam}, which involves a virtual camera process strategy and removes the unobserved regions of  any camera frustum. Specifically, we randomly sample 200,000 points from the reconstructed and the ground truth mesh to calculate three metrics.
The accuracy (Acc.) represents the distance from the sampled points on the reconstructed mesh to their nearest points on the ground truth mesh. 
The completion (Comp.) represents the distance from the sampled points on the ground truth mesh to their nearest points on the reconstructed mesh.
The completion ratio (Comp. Ratio) represents the percentage of sampled points with completion smaller than 5cm. Additionally, we adopt Depth L1 as depth evaluation metric for the scene mesh. In the tracking evaluation, we employ ATE RMSE as the metrics for pose estimation.

\noindent\noindent
\textbf{Implementation Details.}
Our hardware includes a single NVIDIA GeForce RTX 3090 GPU and an Intel Core i7-11700 (16 cores @ 2.50GHz). Our sparse tri-plane encoding comprises 16 levels of 2D hash-grids, with the length of a hash table in each level being $2^{18}$. Additionally, we perform tracking with 1024 rays and mapping with 2048 rays, including the 20-length window in each HBA iteration. The finest size of the square grid of 2D hash-grid is 2cm, and we set 10cm as the truncated distance for the TSDF.

\subsection{Evaluation of Tracking and Reconstruction}

\begin{figure}[!t]
          \rotatebox[origin=clt]{90}{\scriptsize{Co-SLAM\cite{wang2023coslam}}}
          \begin{subfigure}{0.47\linewidth}
            \includegraphics[width=1\textwidth]{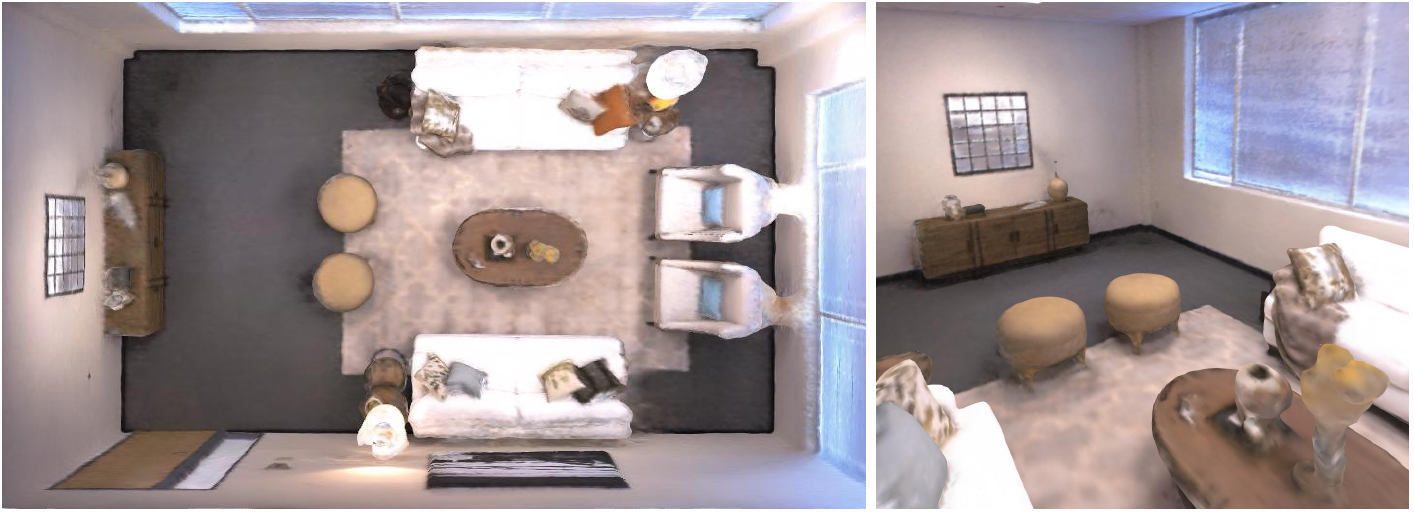}
          \end{subfigure}
          \rotatebox[origin=lb]{90}{\scriptsize{Point-SLAM\cite{sandstrom2023pointslam}}}
          \begin{subfigure}{0.47\linewidth}
                \includegraphics[width=1\textwidth]{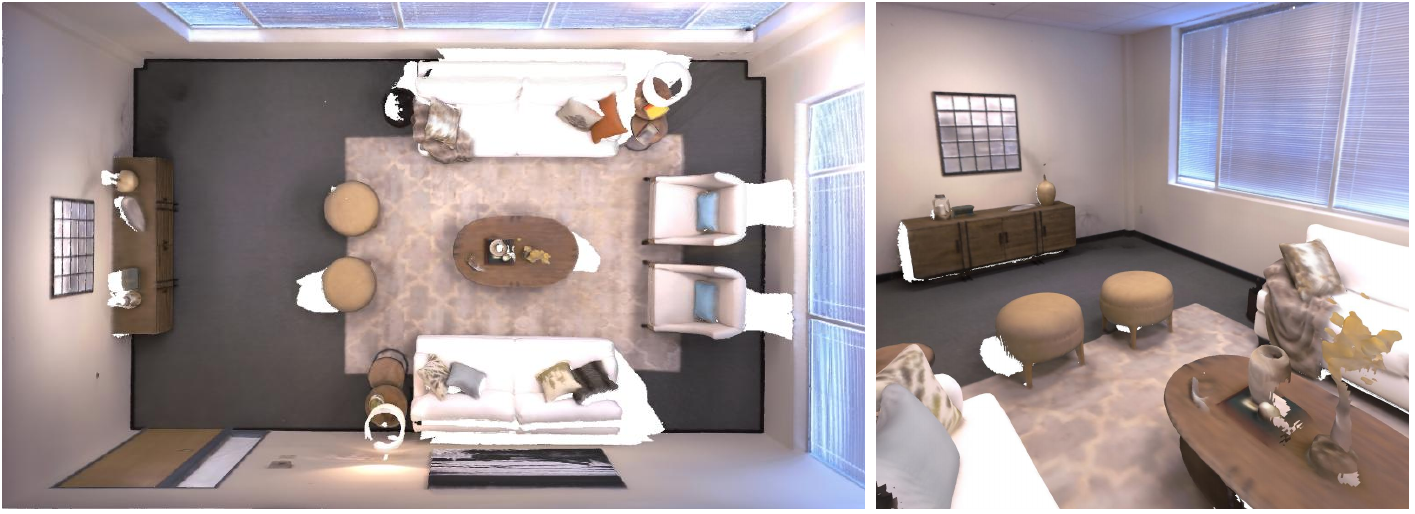}
          \end{subfigure}
          \rotatebox[origin=lb]{90}{\footnotesize{\hspace{6mm} Ours}}
          \begin{subfigure}{0.47\linewidth}
                \includegraphics[width=1\textwidth]{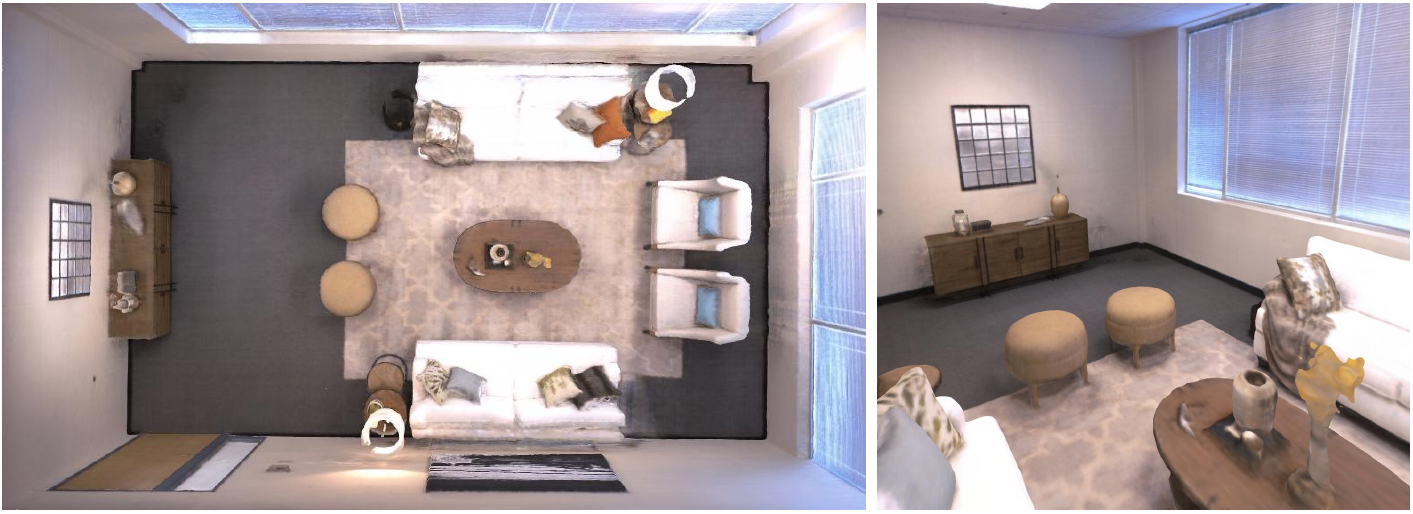}
          \end{subfigure}
          \rotatebox[origin=lt]{90}{\footnotesize{\hspace{5mm} GT}}
          \begin{subfigure}{0.47\linewidth}
                \includegraphics[width=1\textwidth]{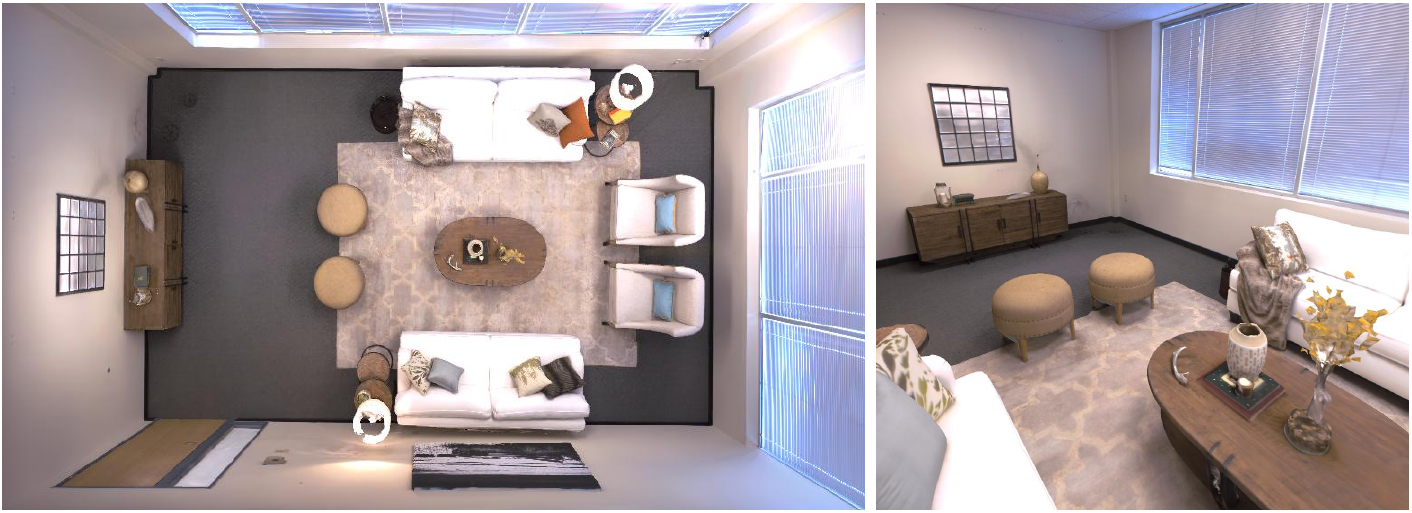}
          \end{subfigure}

          \rotatebox[origin=lt]{90}{\scriptsize{Co-SLAM\cite{wang2023coslam}}}
          \begin{subfigure}{0.47\linewidth}
                \includegraphics[width=1\textwidth]{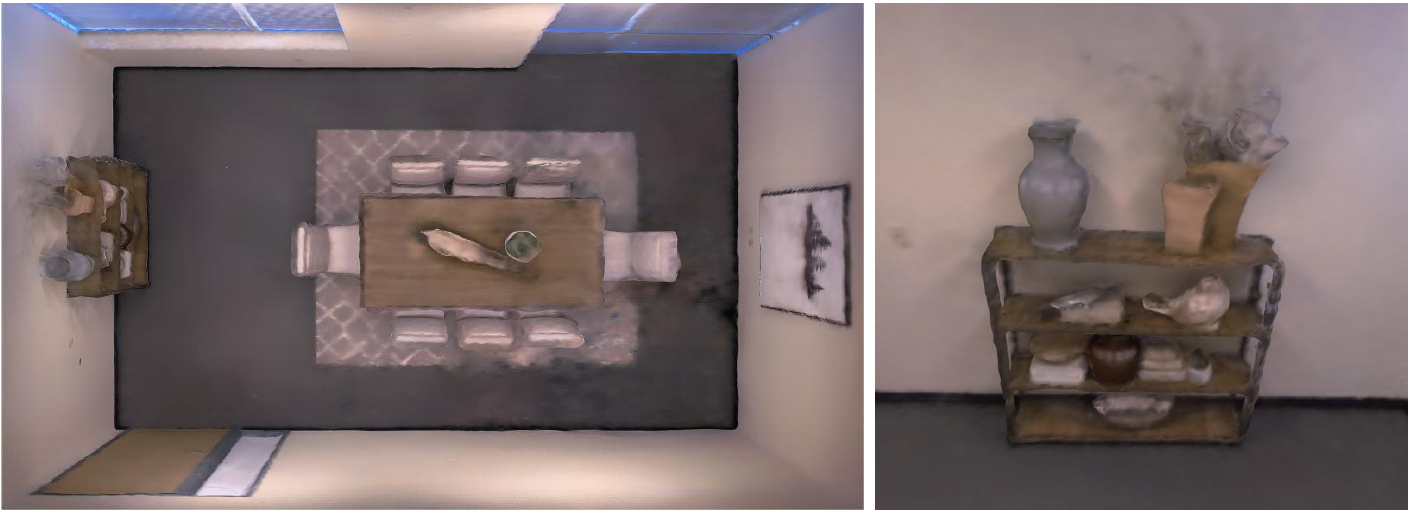}
          \end{subfigure}
          \rotatebox[origin=lb]{90}{\scriptsize{Point-SLAM\cite{sandstrom2023pointslam}}}
          \begin{subfigure}{0.47\linewidth}
                \includegraphics[width=1\textwidth]{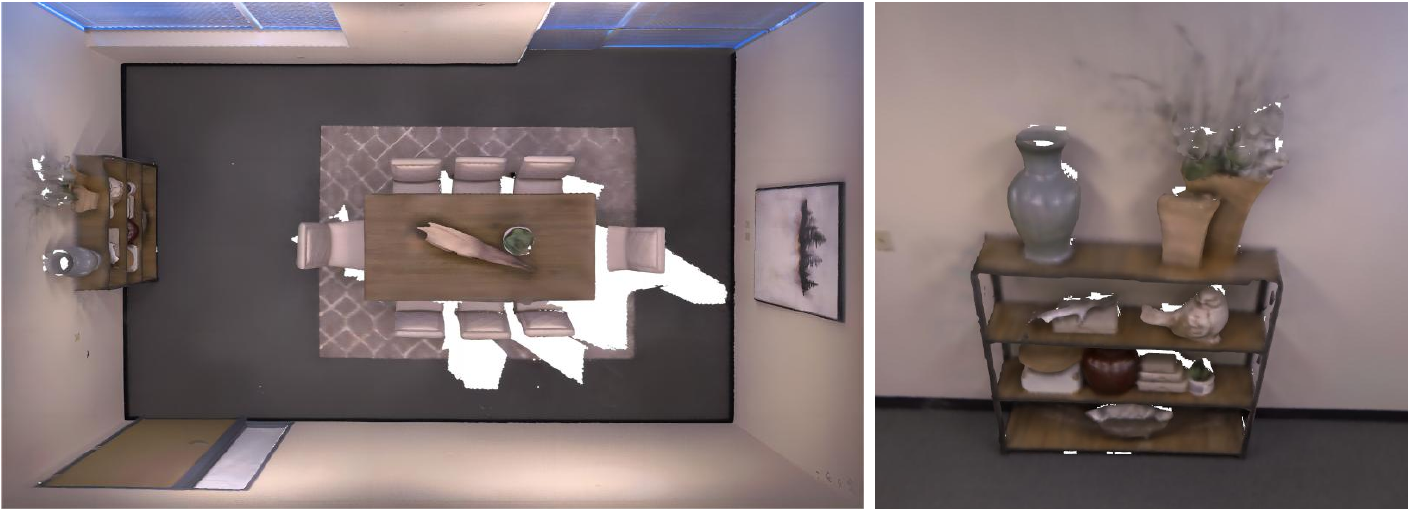}
          \end{subfigure}
          \rotatebox[origin=lt]{90}{\footnotesize{\hspace{4mm} Ours}}
          \begin{subfigure}{0.47\linewidth}
                \includegraphics[width=1\textwidth]{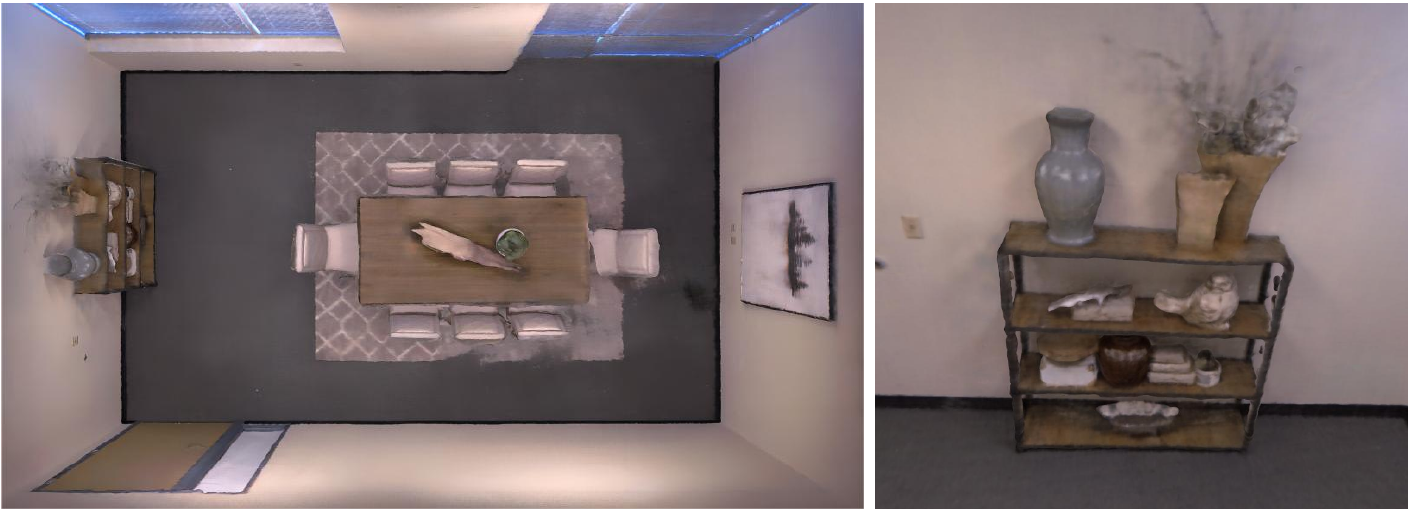}
          \end{subfigure}
          \rotatebox[origin=lt]{90}{\footnotesize{\hspace{5mm} GT}}
          \begin{subfigure}{0.47\linewidth}
                \includegraphics[width=1\textwidth]{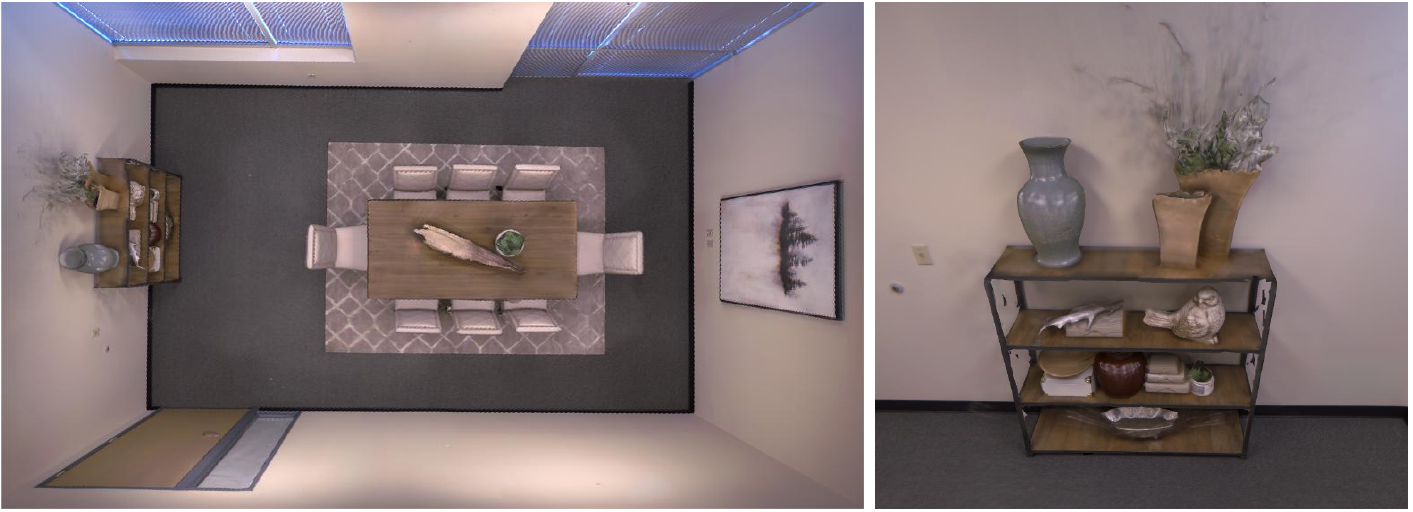}
          \end{subfigure}

          \rotatebox[origin=lt]{90}{\scriptsize{Co-SLAM\cite{wang2023coslam}}}
          \begin{subfigure}{0.47\linewidth}
                \includegraphics[width=1\textwidth]{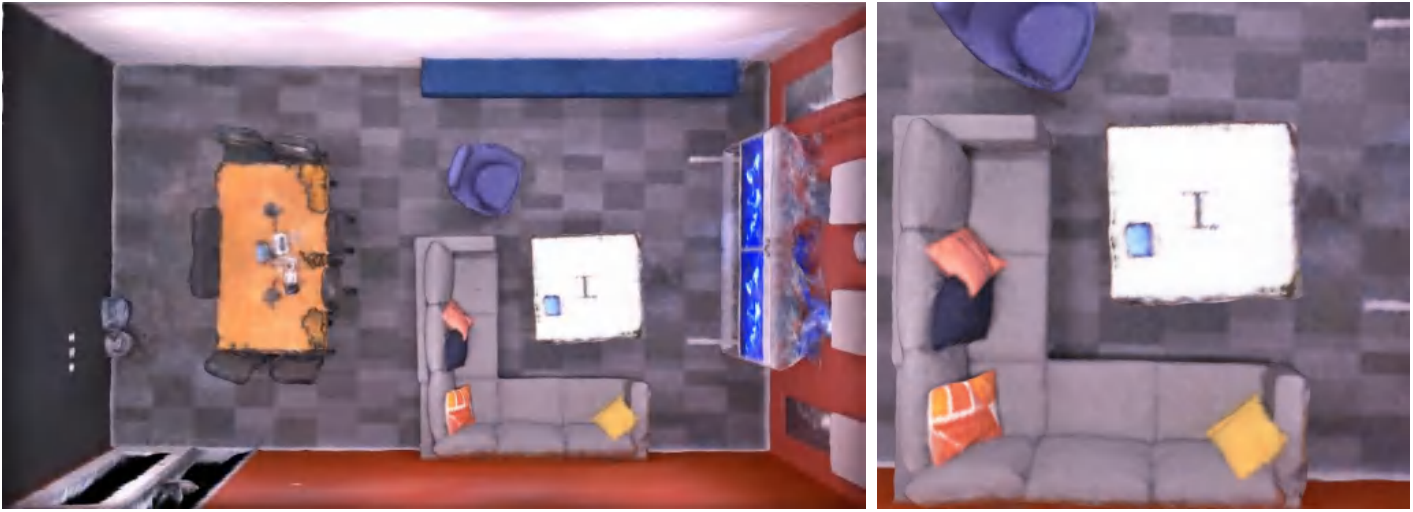}
          \end{subfigure}
          \rotatebox[origin=lb]{90}{\scriptsize{Point-SLAM\cite{sandstrom2023pointslam}}}
          \begin{subfigure}{0.47\linewidth}
                \includegraphics[width=1\textwidth]{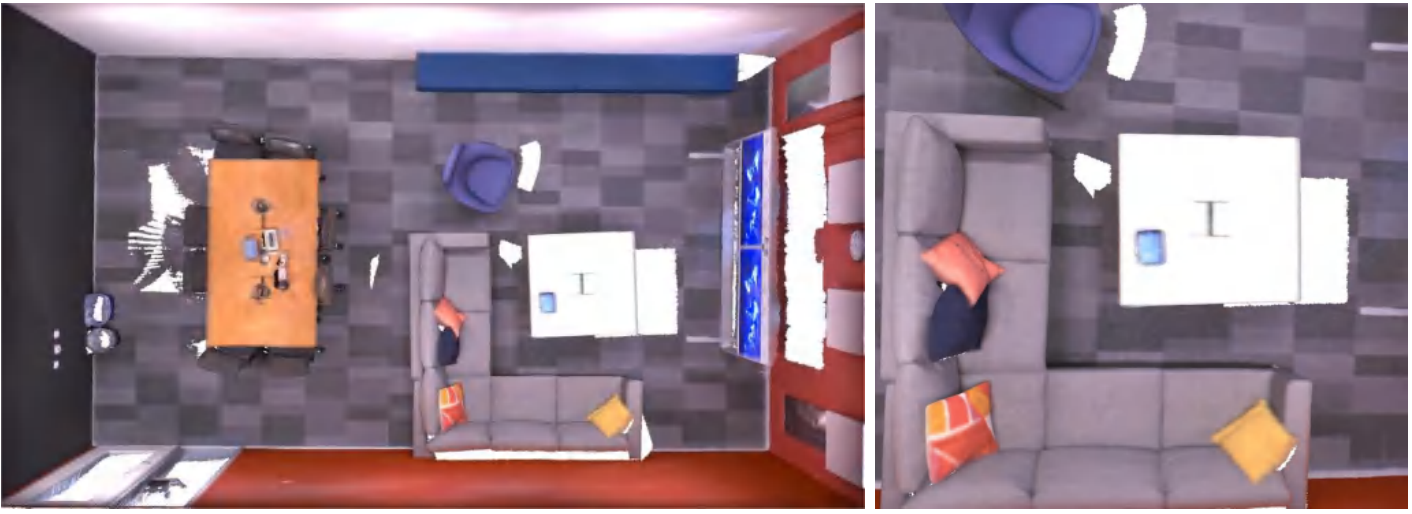}
          \end{subfigure}
          \rotatebox[origin=lt]{90}{\footnotesize{\hspace{4mm} Ours}}
          \begin{subfigure}{0.47\linewidth}
                \includegraphics[width=1\textwidth]{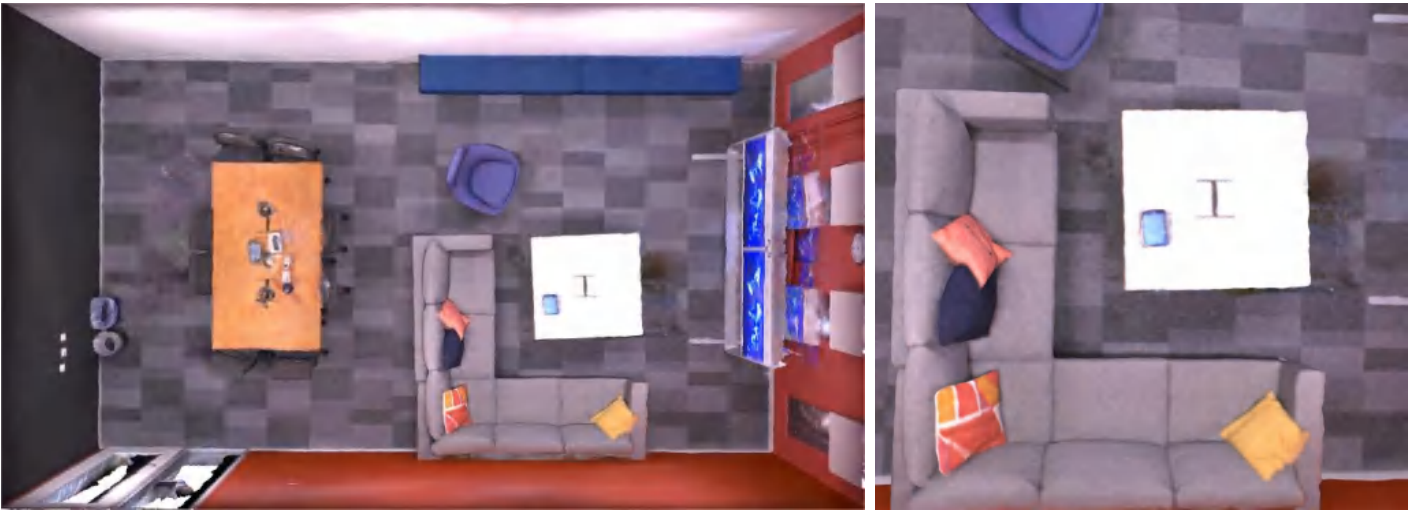}
          \end{subfigure}
          \rotatebox[origin=lt]{90}{\footnotesize{\hspace{5mm} GT}}
          \begin{subfigure}{0.47\linewidth}
                \includegraphics[width=1\textwidth]{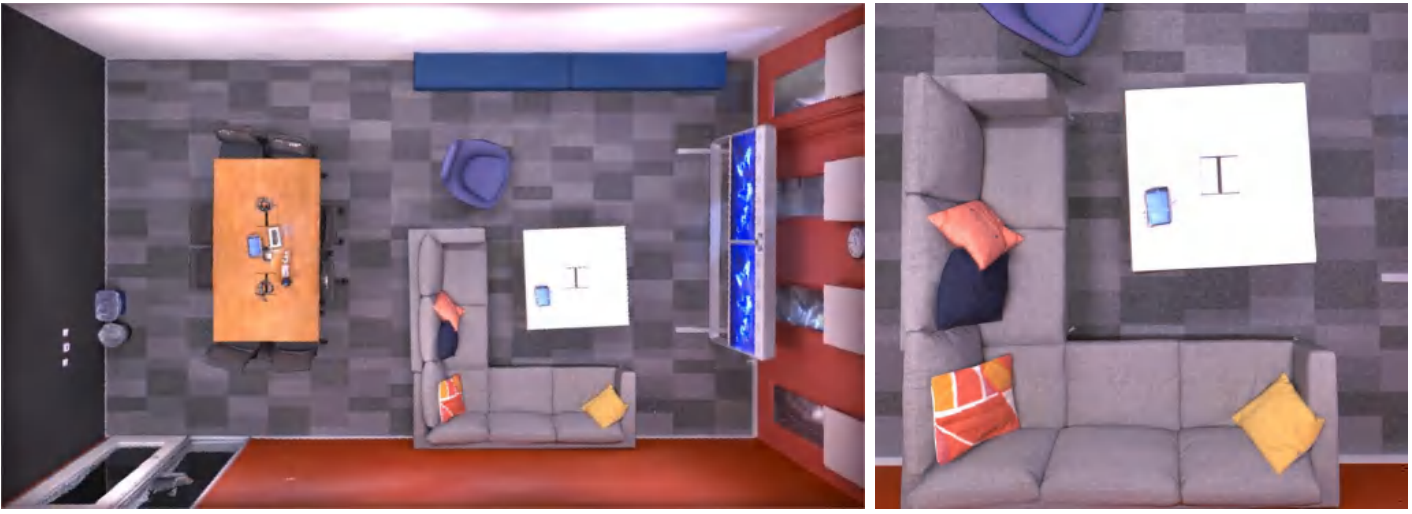}
          \end{subfigure}

  \vspace{-3mm}
  \caption{Qualitative comparison of the Replica\cite{straub2019replica} dataset shows S3-SLAM can reconstruct complete geometry of unknown viewpoints while reconstructing fine-level appearance. S3-SLAM employs sparse tri-plane encoding, achieving higher-resolution scene reconstruction.}
  \label{fig:eval_replica}
  \vspace{-2mm}
\end{figure}

\noindent\noindent
\textbf{Evaluation on Replica\cite{straub2019replica}.}
We evaluate our reconstruction performance on the Replica dataset. 
As shown in \cref{tab:replica_tracking}, our method achieves the highest quality reconstruction with the few number of iterations. As shown in \cref{fig:eval_replica}, Co-SLAM\cite{wang2023coslam} generates smooth geometry but struggles to reconstruct sharp contours and thin structures. Additionally, Co-SLAM faces challenges in effectively completing certain corners due to limitations in joint encoding. Point-SLAM\cite{sandstrom2023pointslam} refines geometry effectively but struggles to generate complete geometry, displaying numerous geometric holes. In contrast, our S3-SLAM leverages sparse tri-plane encoding to reconstruct high-fidelity appearance, completing nearly all scene geometry without compromising much accuracy.

\begin{table}[t]
      \scriptsize
      \centering
            \setlength{\tabcolsep}{1.7mm}{
            \begin{tabular}{lccccc}
            \toprule
            Method & Depth L1 (cm)$\downarrow$ & Acc. (cm)$\downarrow$ & Comp. (cm)$\downarrow$ & Comp. Ratio (\%)$\uparrow$ \\
            \midrule
            NICE-SLAM\cite{zhu2022nice} & 1.90 & 2.37 & 2.64 & 91.13 \\
            Co-SLAM \cite{wang2023coslam} & 1.51 & 2.10 & 2.08 & 93.44 \\
            Point-SLAM\cite{sandstrom2023pointslam} & 3.92 & \textbf{1.26} & 2.95 & 89.10 \\
            Ours & \textbf{0.98} & 2.11 & \textbf{1.66} & \textbf{96.71} \\
            \bottomrule
            \end{tabular}}
      \caption{
            Quantitative comparison results on the Replica\cite{straub2019replica} dataset, where each metric represents the average value across all scenes in the datasets. Our S3-SLAM achieves higher-quality reconstruction with the fewest iterations.
      }
      \vspace{-2mm}
      \label{tab:replica_tracking}
\end{table}

\begin{figure*}[!t]
  \centering
  \captionsetup[subfigure]{labelformat=empty}

  \rotatebox[origin=lb]{90}{\fontsize{9pt}{1pt} \hspace{8mm} \ttfamily {scene0000}}
  \begin{subfigure}{0.23\linewidth}
        \centering
        \captionsetup{justification=raggedright}
        \caption{ \scriptsize{Co-SLAM\cite{wang2023coslam}}}
        \includegraphics[width=1\textwidth]{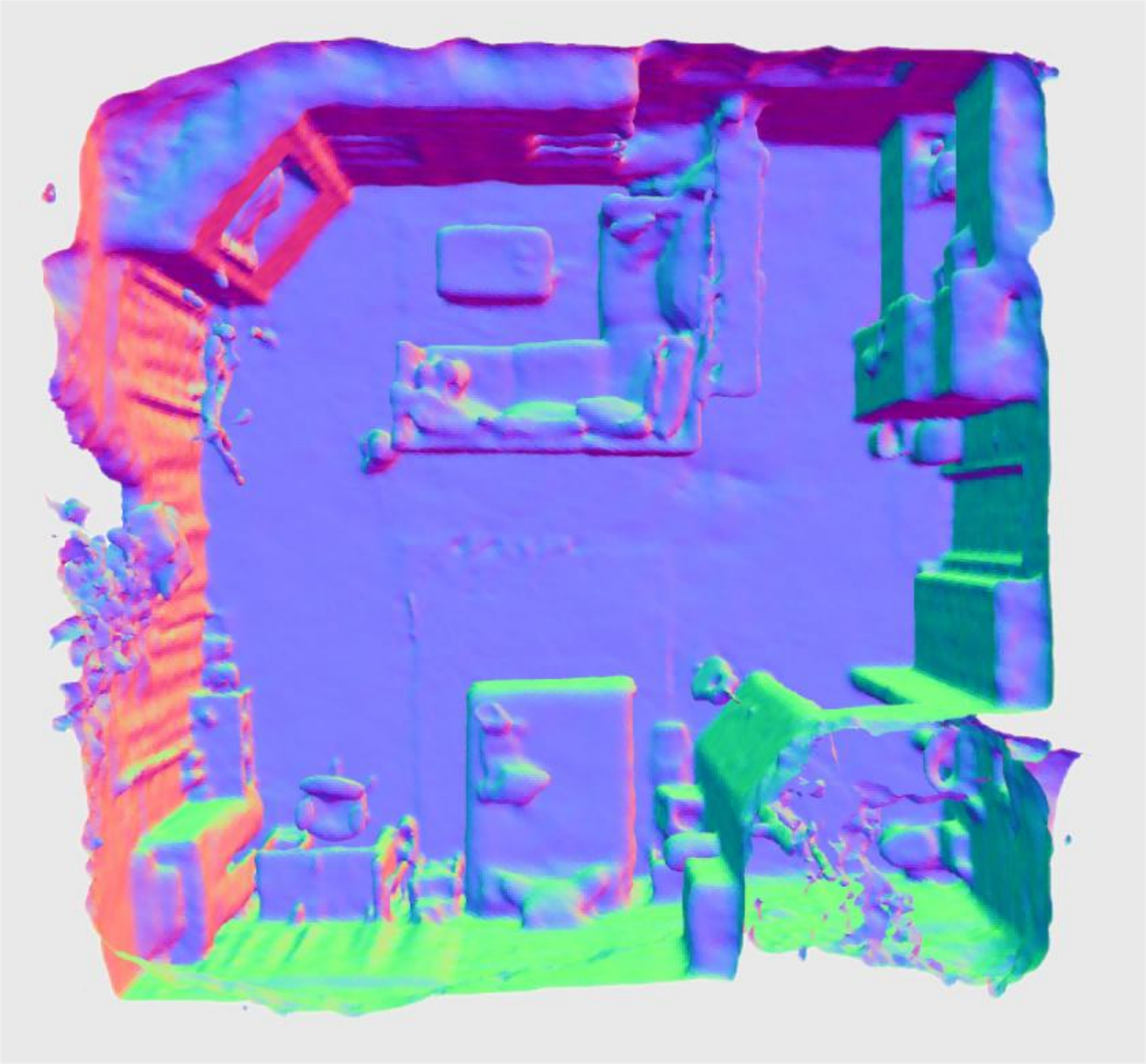}
        \label{000_coslam}
  \end{subfigure}
  \begin{subfigure}{0.23\linewidth}
        \centering
        \captionsetup{justification=raggedright}
        \caption{\scriptsize{Point-SLAM\cite{sandstrom2023pointslam}}}
        \includegraphics[width=1\textwidth]{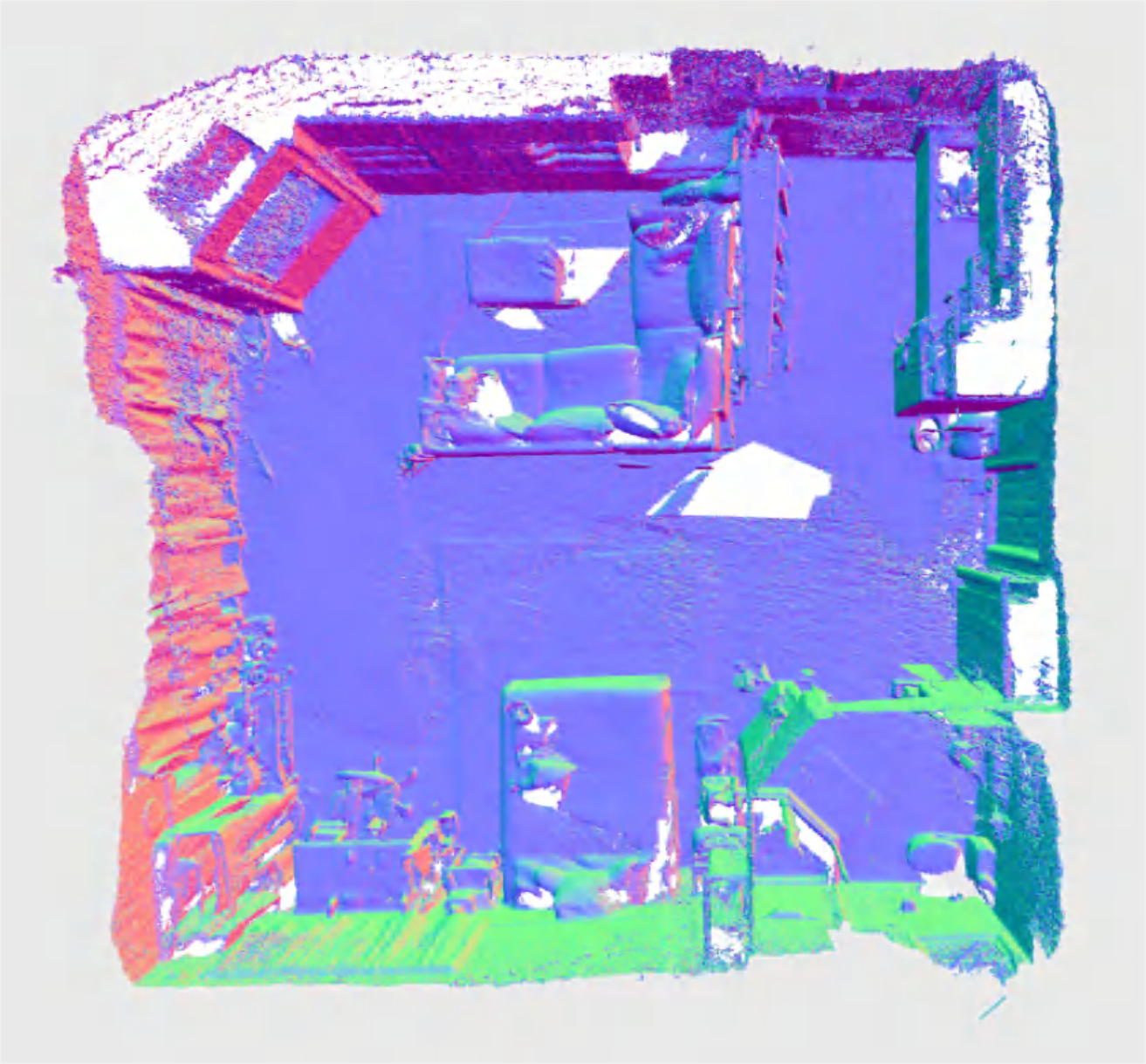}
        \label{000_pointslam}
  \end{subfigure}
  \begin{subfigure}{0.23\linewidth}
        \captionsetup{justification=raggedright}
        \caption{\scriptsize{Ours}}
        \includegraphics[width=1\textwidth]{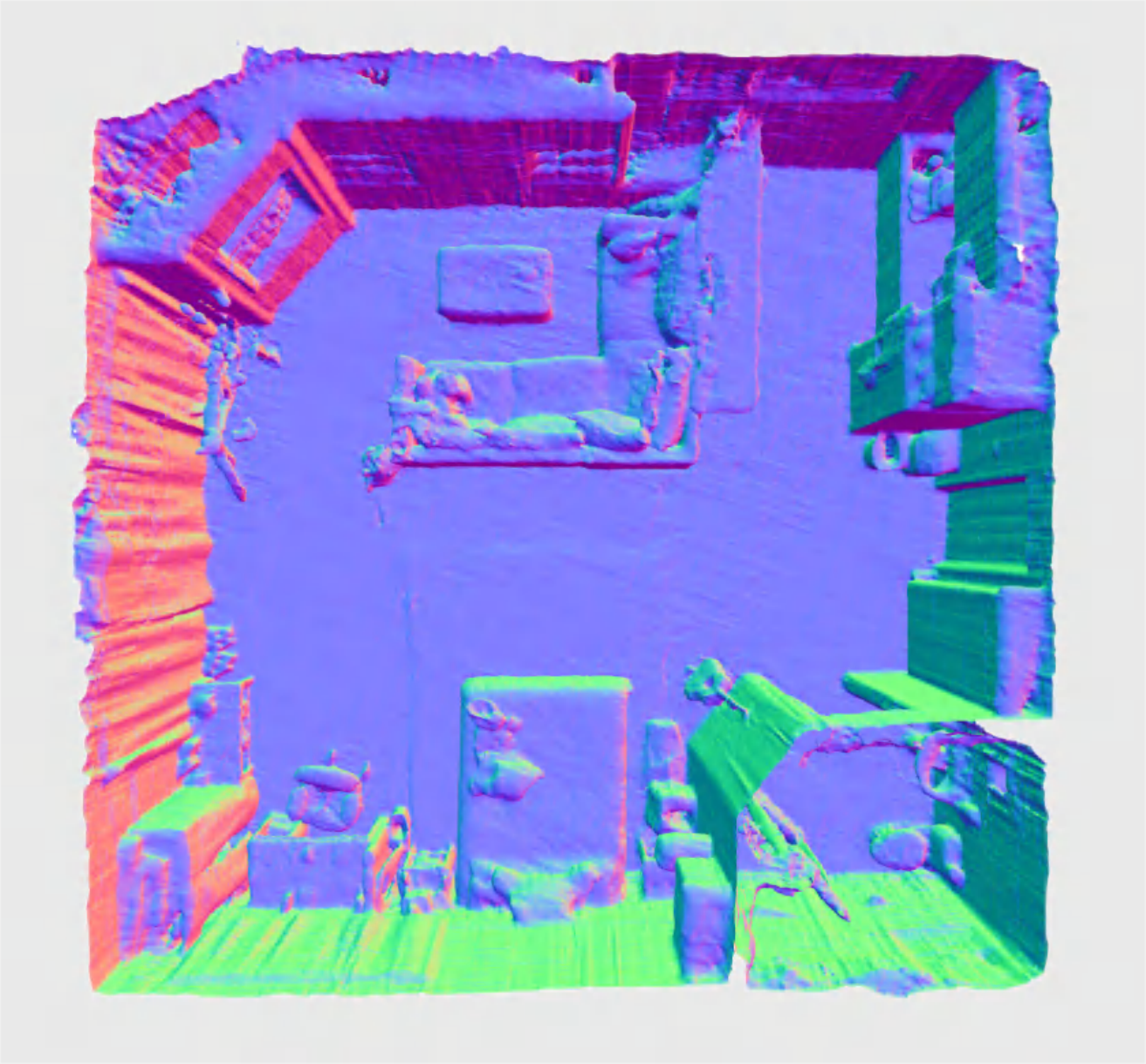}
        \label{000_ours}
  \end{subfigure}
  \begin{subfigure}{0.23\linewidth}
        \captionsetup{justification=raggedright}
        \caption{\scriptsize{GT}}
        \includegraphics[width=1\textwidth]{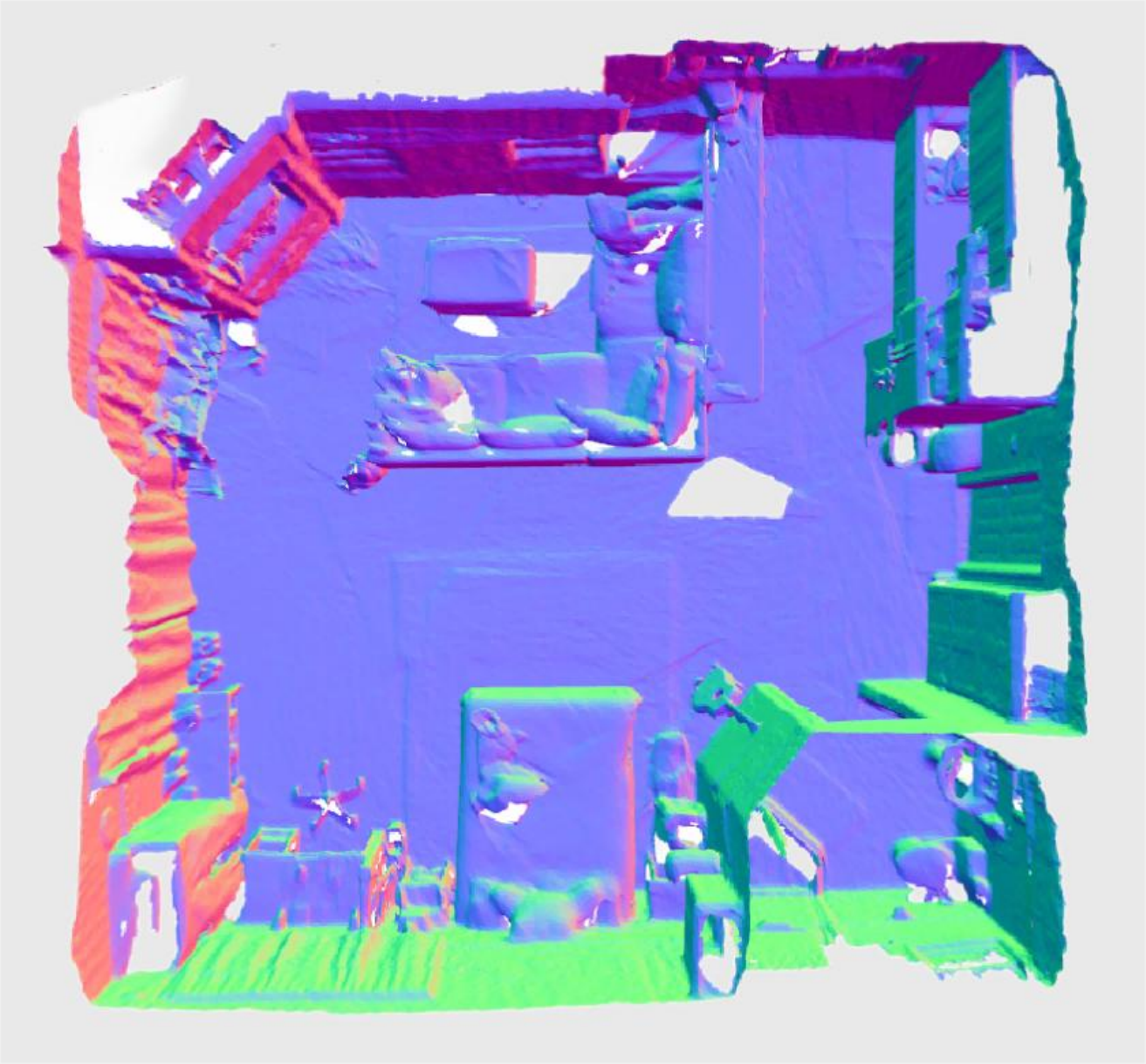}
        \label{000_gt}
  \end{subfigure}
  
  \vspace{-4mm}
  \rotatebox[origin=lb]{90}{\fontsize{9pt}{1pt} \hspace{6mm} \ttfamily {scene0169}}
  \begin{subfigure}{0.23\linewidth}
        \includegraphics[width=1\textwidth]{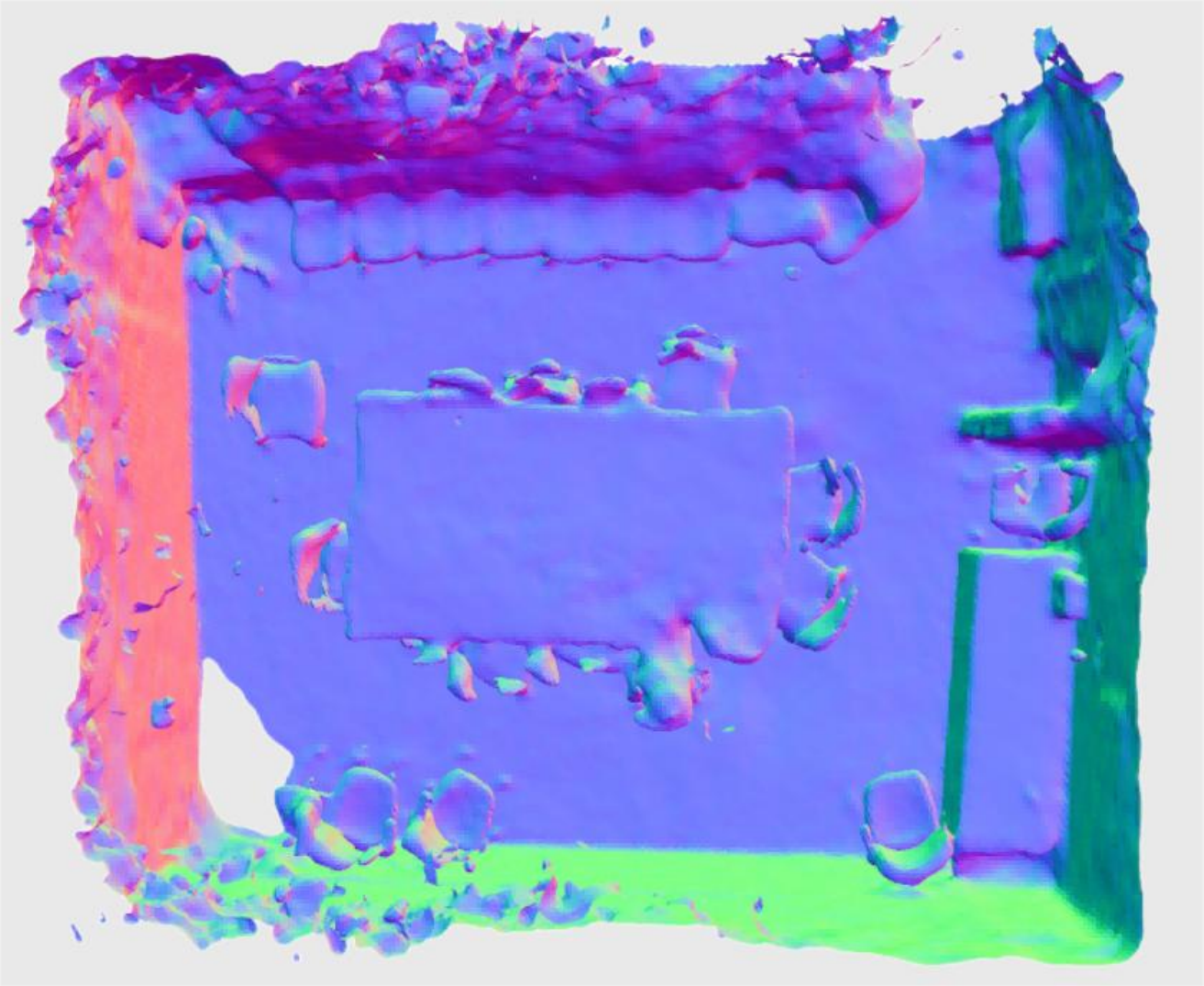}
        \label{169_coslam}
  \end{subfigure}
  \begin{subfigure}{0.23\linewidth}
        \includegraphics[width=1\textwidth]{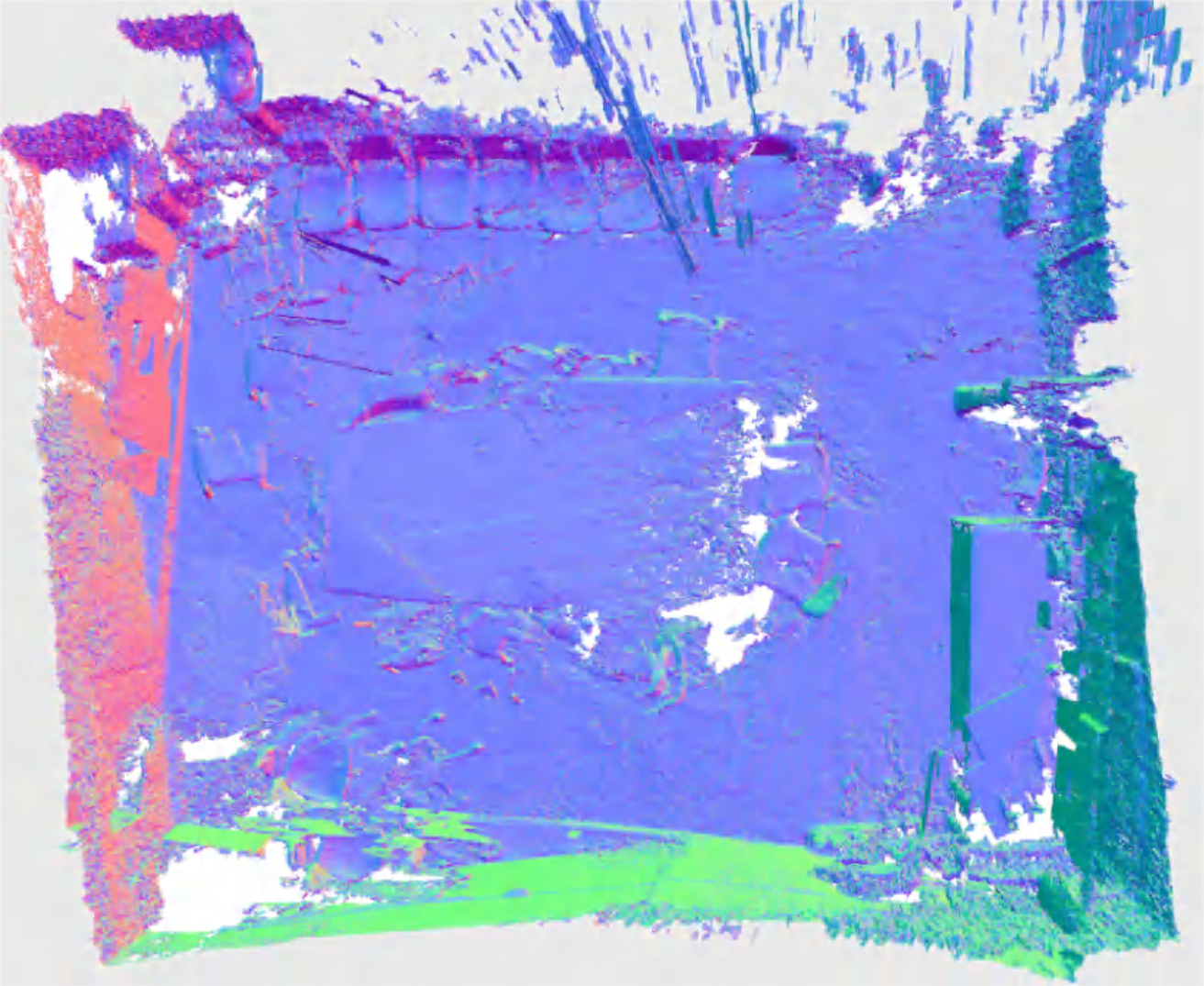}
        \label{169_pointslam}
  \end{subfigure}
  \begin{subfigure}{0.23\linewidth}
        \includegraphics[width=1\textwidth]{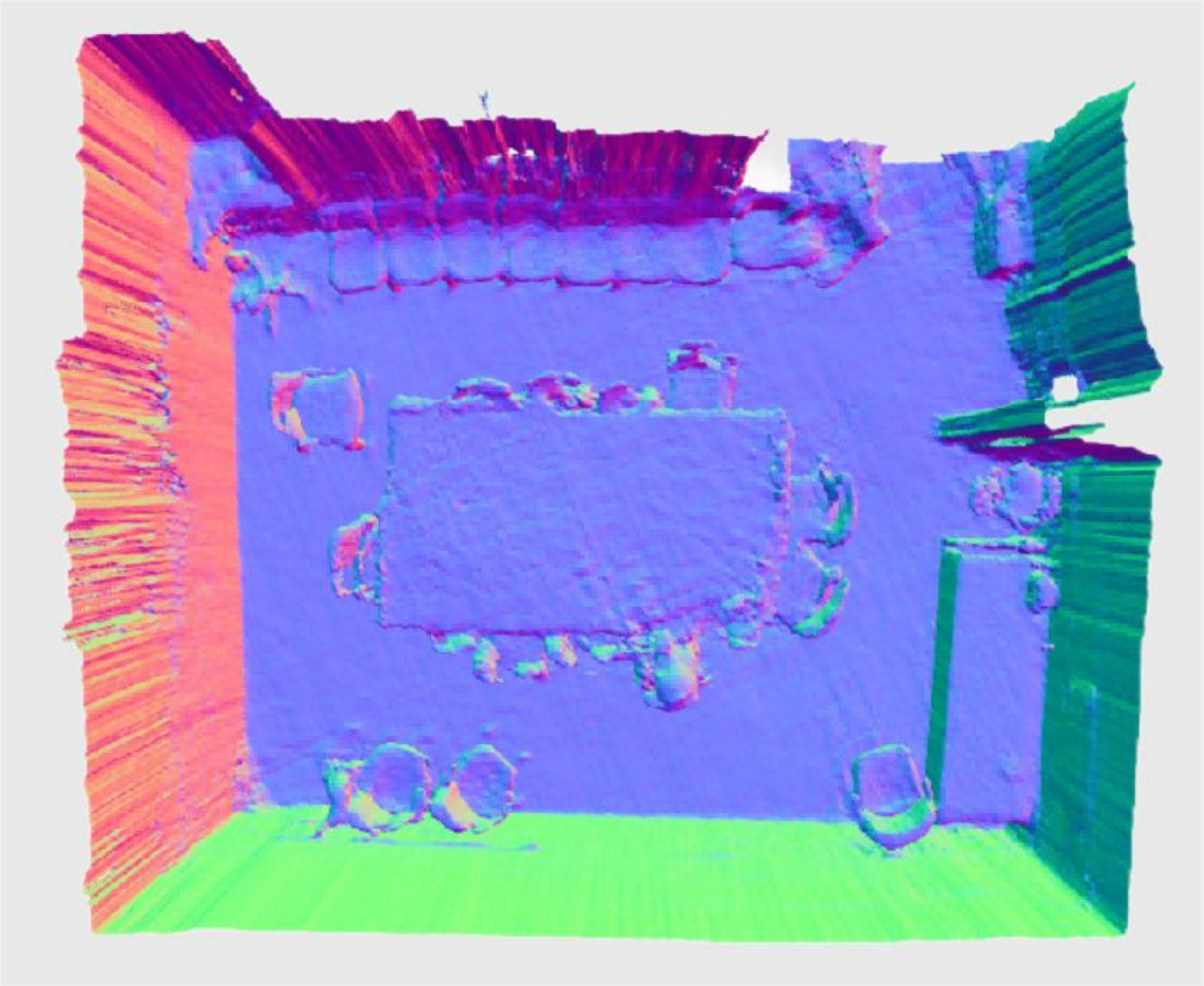}
        \label{169_ours}
  \end{subfigure}
  \begin{subfigure}{0.23\linewidth}
        \includegraphics[width=1\textwidth]{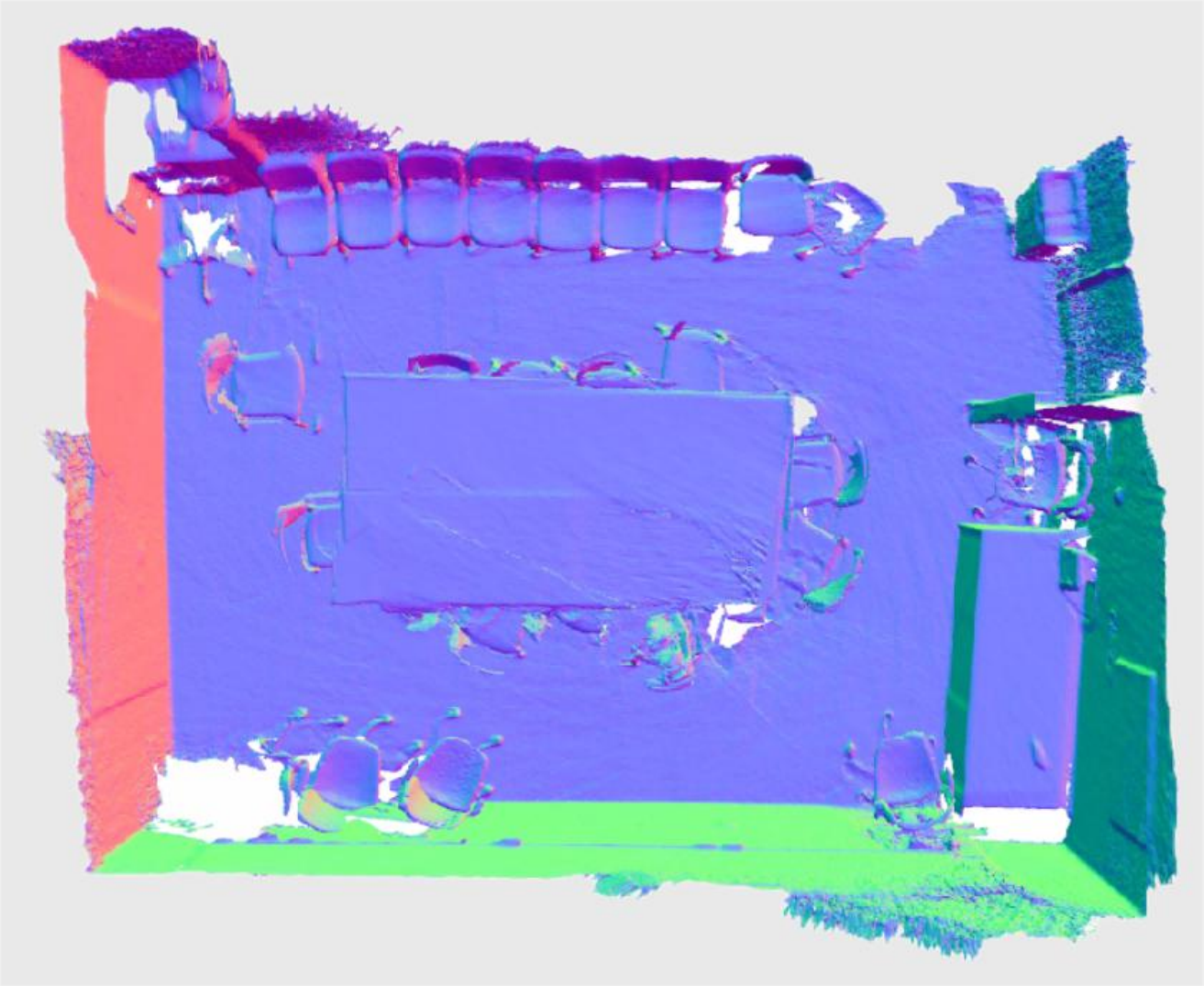}
        \label{169_gt}
  \end{subfigure}

  \vspace{-6mm}
  \caption{Qualitative comparison of geometric results on the ScanNet\cite{dai2017scannet} dataset demonstrates our method can produce complete and coherent geometry. Compared to Co-SLAM\cite{wang2023coslam}, S3-SLAM can reconstruct higher-resolution geometry, especially regarding surface contours. In contrast to Point-SLAM\cite{sandstrom2023pointslam}, our method can generate more accurate and complete geometry.}
  \label{fig:scannet_eval}
  \vspace{-3mm}
\end{figure*}

\begin{figure*}[!t]
  \vspace{-1mm}
  \centering
  \captionsetup[subfigure]{labelformat=empty}

  \rotatebox[origin=lb]{90}{\fontsize{9pt}{1pt} \hspace{4mm} \ttfamily {scene0000}}
  \begin{subfigure}{0.23\linewidth}
        \centering
        \captionsetup{justification=raggedright}
        \caption{\scriptsize{NICE-SLAM\cite{zhu2022nice}}}
        \includegraphics[width=1\textwidth]{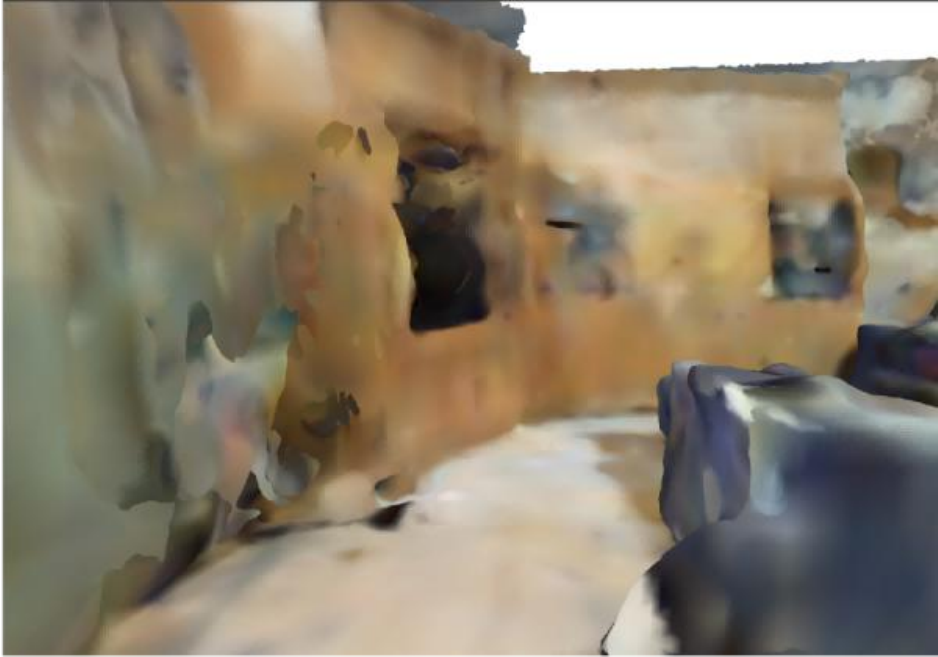}
        \label{000_niceslam}
  \end{subfigure}
  \begin{subfigure}{0.23\linewidth}
        \centering
        \captionsetup{justification=raggedright}
        \caption{\scriptsize{Co-SLAM\cite{wang2023coslam}}}
        \includegraphics[width=1\textwidth]{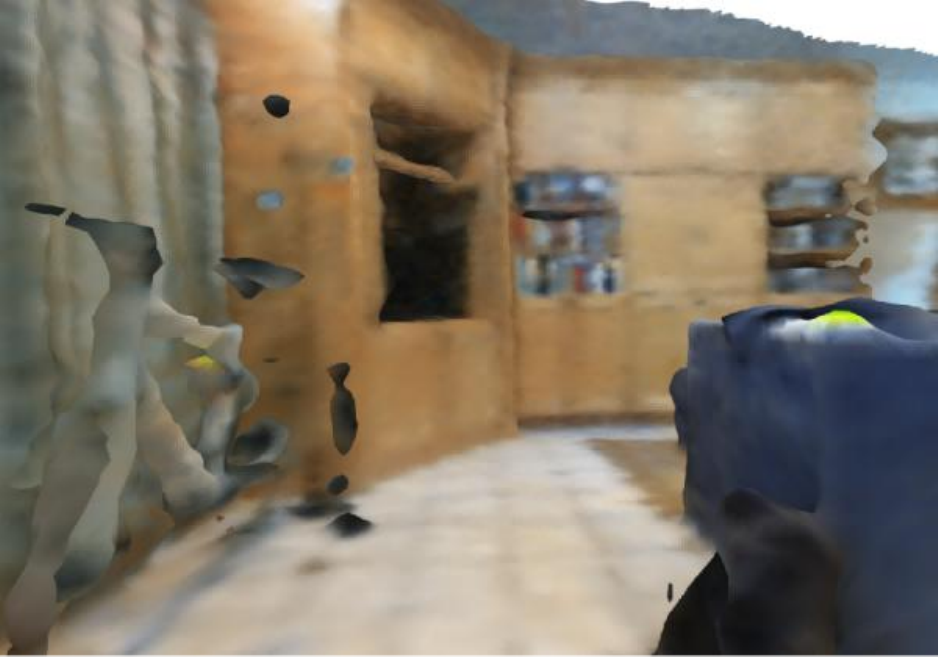}
        \label{000_coslam}
  \end{subfigure}
  \begin{subfigure}{0.23\linewidth}
        \captionsetup{justification=raggedright}
        \caption{\scriptsize{Ours}}
        \includegraphics[width=1\textwidth]{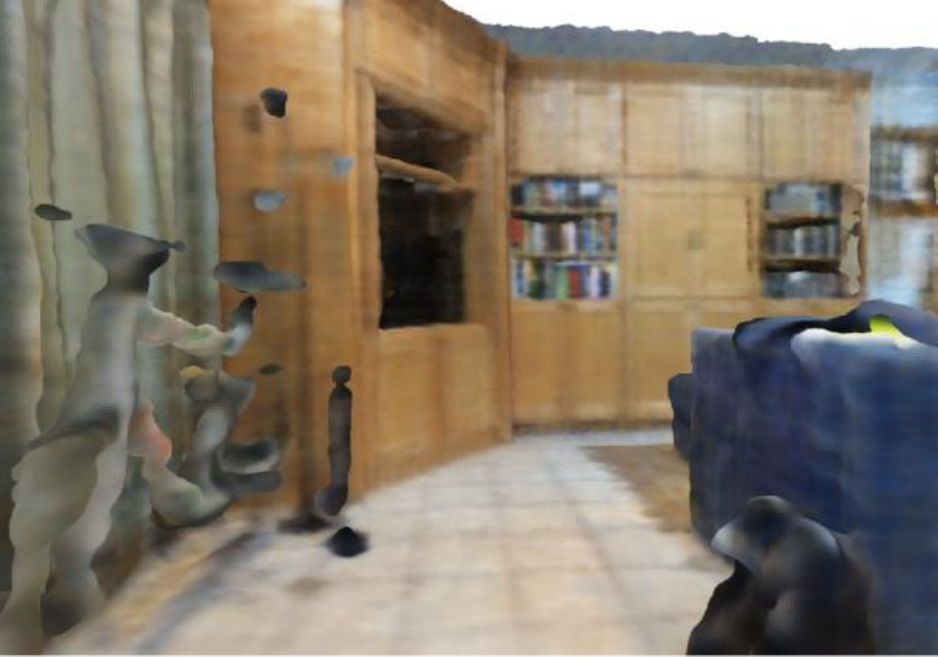}
        \label{000_ours}
  \end{subfigure}
  \begin{subfigure}{0.23\linewidth}
        \captionsetup{justification=raggedright}
        \caption{\scriptsize{GT}}
        \includegraphics[width=1\textwidth]{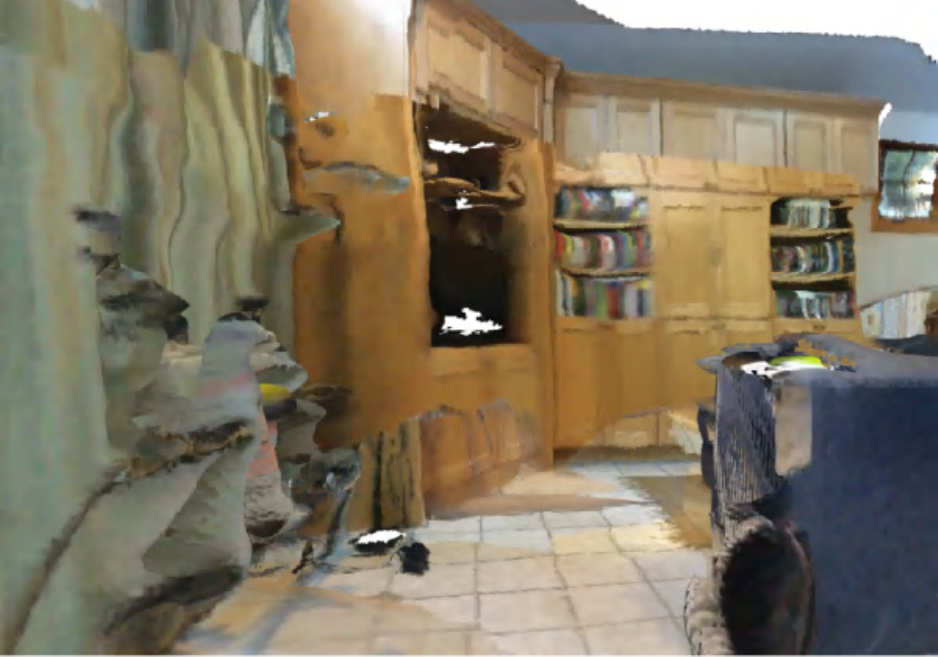}
        \label{000_gt}
  \end{subfigure}
  
  \vspace{-4mm}
  \rotatebox[origin=lb]{90}{\fontsize{9pt}{1pt} \hspace{3mm} \ttfamily {scene0059}}
  \begin{subfigure}{0.23\linewidth}
        \includegraphics[width=1\textwidth]{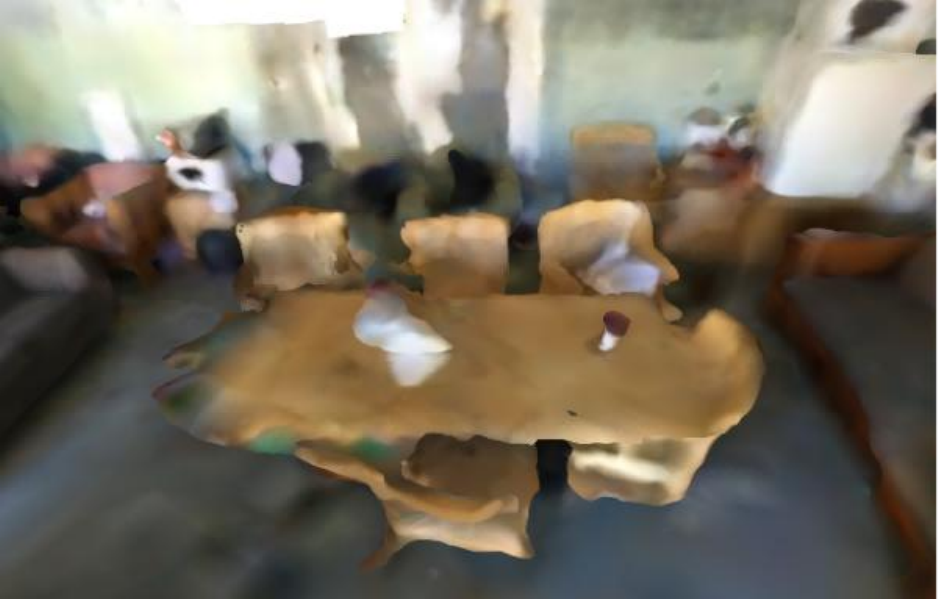}
        \label{59_niceslam}
  \end{subfigure}
  \begin{subfigure}{0.23\linewidth}
        \includegraphics[width=1\textwidth]{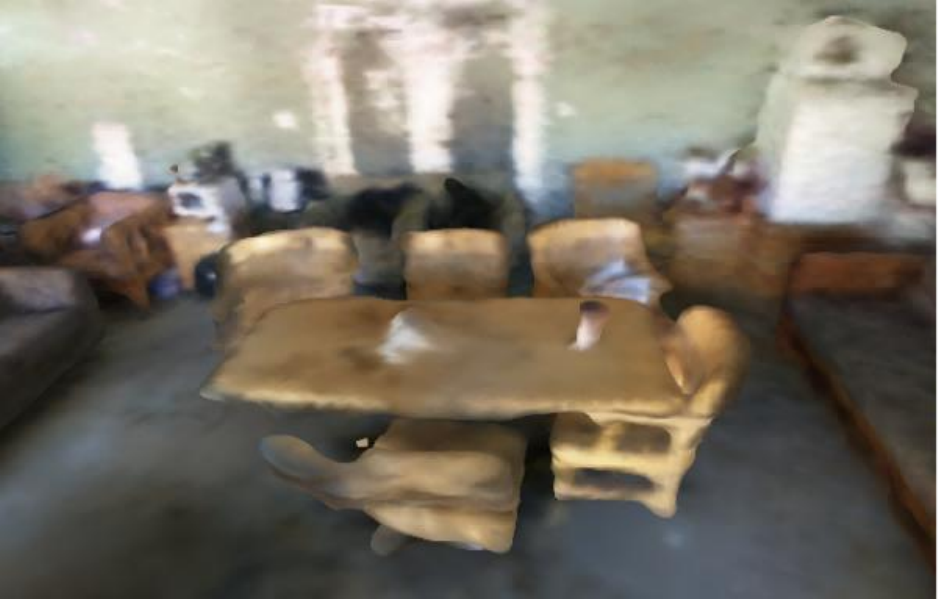}
        \label{59_coslam}
  \end{subfigure}
  \begin{subfigure}{0.23\linewidth}
        \includegraphics[width=1\textwidth]{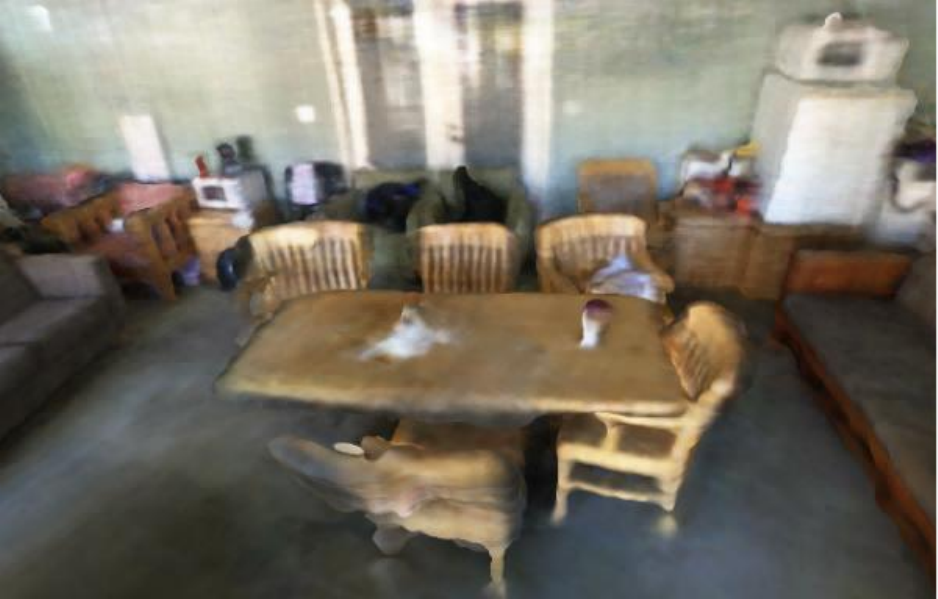}
        \label{59_ours}
  \end{subfigure}
  \begin{subfigure}{0.23\linewidth}
        \includegraphics[width=1\textwidth]{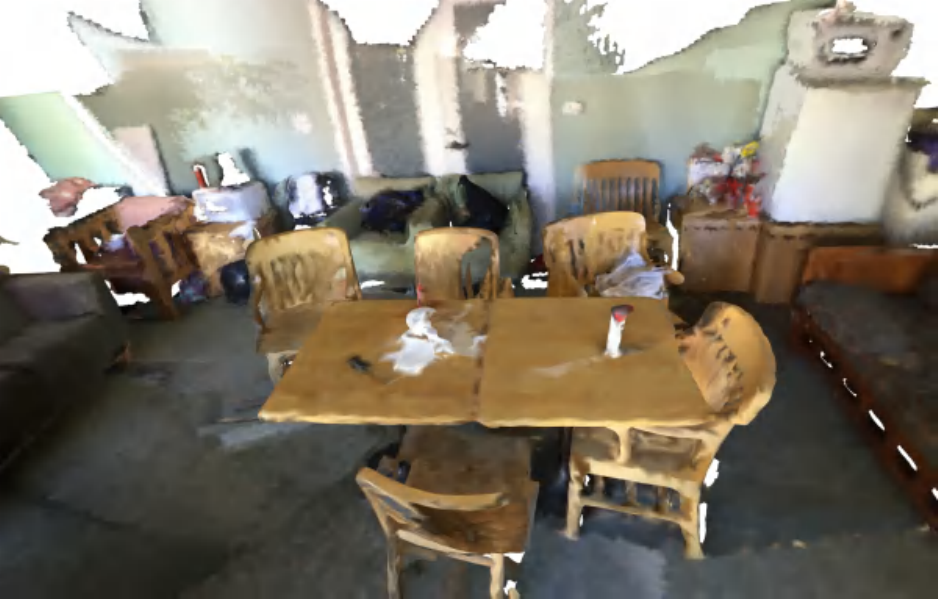}
        \label{59_gt}
  \end{subfigure}

  \vspace{-4mm}
  \rotatebox[origin=lb]{90}{\fontsize{9pt}{1pt} \hspace{2mm} \ttfamily {scene0169}}
  \begin{subfigure}{0.23\linewidth}
        \includegraphics[width=1\textwidth]{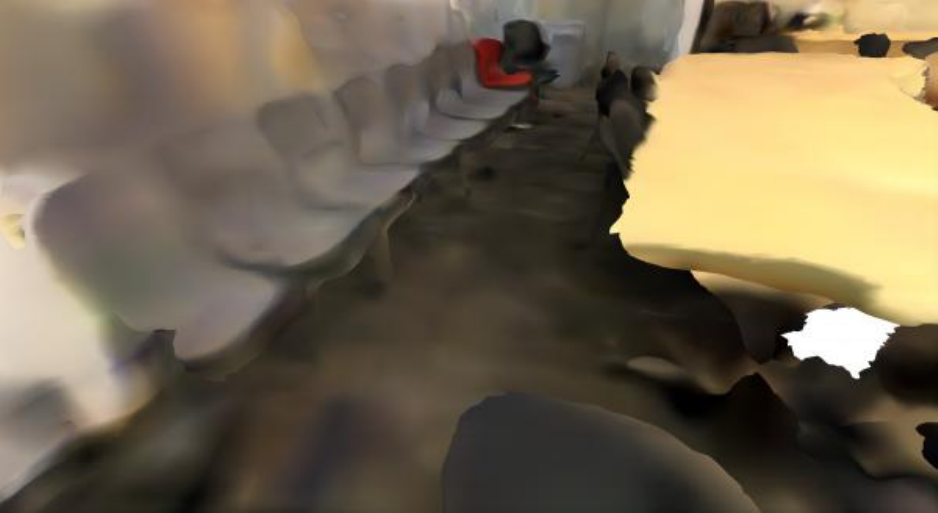}
        \label{169_niceslam}
  \end{subfigure}
  \begin{subfigure}{0.23\linewidth}
        \includegraphics[width=1\textwidth]{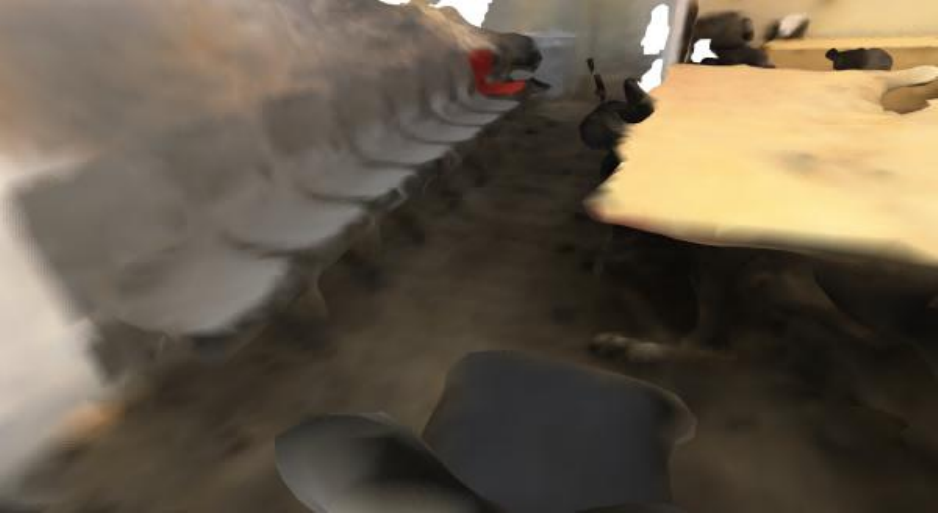}
        \label{169_coslam}
  \end{subfigure}
  \begin{subfigure}{0.23\linewidth}
        \includegraphics[width=1\textwidth]{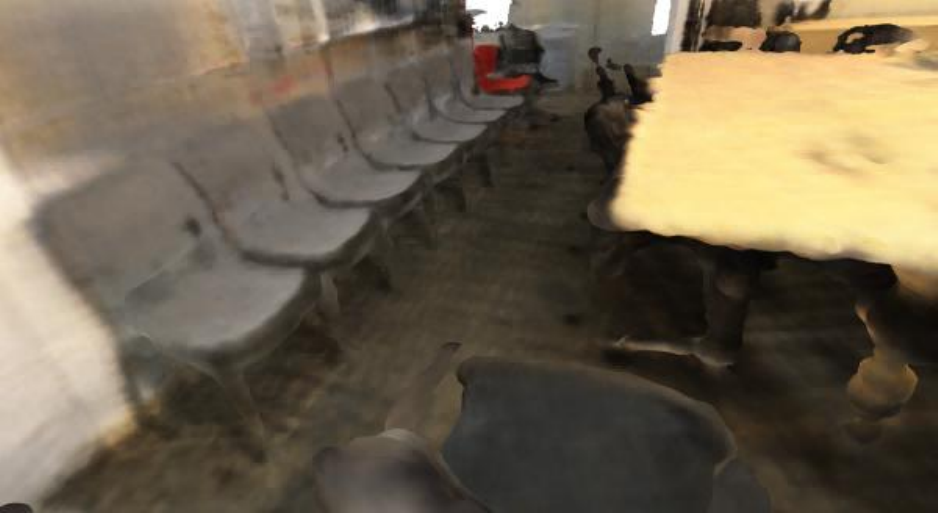}
        \label{169_ours}
  \end{subfigure}
  \begin{subfigure}{0.23\linewidth}
        \includegraphics[width=1\textwidth]{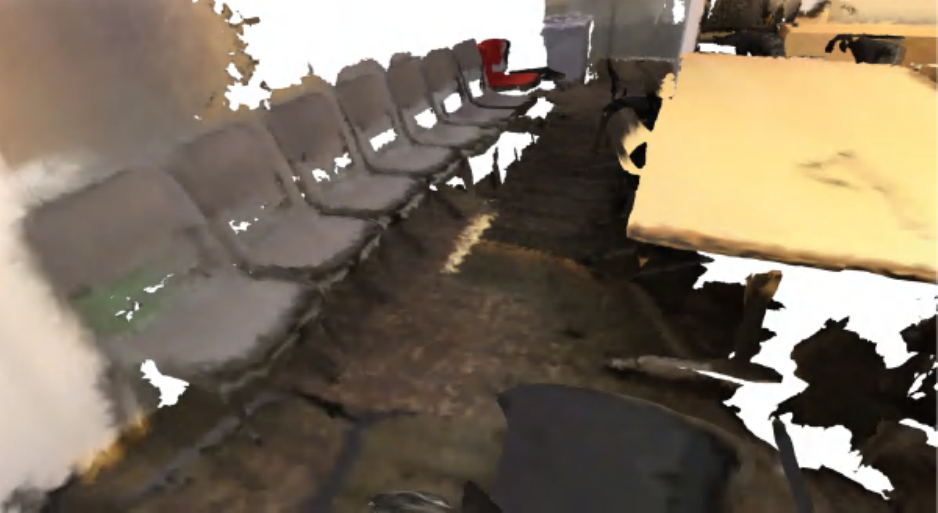}
        \label{169_gt}
  \end{subfigure}

  \vspace{-6mm}
  \caption{Qualitative results on the ScanNet\cite{dai2017scannet} dataset demonstrate that our method can reconstruct fine details. Compared to NICE-SLAM\cite{zhu2022nice} and Co-SLAM\cite{wang2023coslam}, our S3-SLAM can reconstruct more detailed object contours and colors. For example, the details of objects like tables, chairs, and chair backs are reconstructed more accurately.}
  \label{fig:scannet_vis}
  \vspace{-8mm}
\end{figure*}


\begin{table}[!t]
  \vspace{-3mm}
  \scriptsize
  \centering
        \setlength{\tabcolsep}{2.5mm}{
        \begin{tabular}{llccccccc}
        \toprule
        & Method & 0000 & 0059 & 0106 & 0169 & 0181 & 0207 & Avg.\\
        \midrule
        & NICE-SLAM\cite{zhu2022nice} & 8.64 & 12.25 & 8.09 & 10.28 
              & 12.93 & \textbf{5.59} & 9.63\\
        & Co-SLAM \cite{wang2023coslam} 
              & 7.18 & 12.29 & 9.57 
              & 6.62 
              & 13.43 & 7.13 & 9.37\\    
        & Point-SLAM\cite{xu2022point}
        & 10.24 & \textbf{7.81} & 8.65 & 22.16 & 14.77 
        & 9.54 & 12.19\\ 
        & Ours & \textbf{6.31} & 9.84 & \textbf{8.02} & \textbf{5.81} & \textbf{12.45} & 6.41 & \textbf{8.14}\\
        \bottomrule
        \end{tabular}}
  \caption{We calculate the ATE RMSE$\downarrow$ [cm] for camera tracking on ScanNet\cite{dai2017scannet}. Quantitative results on ScanNet demonstrate that our method achieves highly competitive camera tracking.}
  \vspace{-10mm}
  \label{tab:scannet_tracking}
\end{table}

\noindent\noindent
\textbf{Evaluation on ScanNet\cite{dai2017scannet}.}
Our tracking performance outperforms existing neural implicit SLAM methods. As shown in \cref{tab:scannet_tracking}, leveraging sparse tri-plane encoding, our S3-SLAM achieves competitive camera tracking. In most scenes, our S3-SLAM outperforms the baseline, thanks to our compact orthogonal plane features that result in higher accuracy of camera tracking.
Qualitatively, as shown in \cref{fig:scannet_eval} and \cref{fig:scannet_vis}, S3-SLAM utilizes our new sparse tri-plane encoding, rapidly reconstructs geometric details and completes unobserved scene geometry compared to the three baseline methods. NICE-SLAM\cite{zhu2022nice} struggles to reconstruct fine-level geometry and appearance because hierarchical dense feature grid scene representation. Although faster speed of Co-SLAM\cite{wang2023coslam}, Co-SLAM results in geometric deformations. Point-SLAM\cite{sandstrom2023pointslam} uses neural points to reconstruct geometry and it cannot complete unobserved regions, resulting in many holes in the reconstructed mesh.

\begin{table}[!t]
      \scriptsize
      \centering
      \setlength{\tabcolsep}{4.7mm}{
      \begin{tabular}{llcccc}
      \toprule
      & Method & fr1/desk & fr2/xyz & fr3/office & Avg. \\
      \midrule
      & NICE-SLAM\cite{zhu2022nice} & 2.7 & 1.8 & 3.0 & 2.5\\
      & Co-SLAM \cite{wang2023coslam} & 2.7 & 1.9 & 2.6 & 2.4\\     
      & Point-SLAM \cite{sandstrom2023pointslam} & 4.3 & \textbf{1.3} & 3.5 & 3.0\\     
      & Ours & \textbf{2.6} & 1.7 & \textbf{2.4} & \textbf{2.3}\\
      \bottomrule
      \end{tabular}}
      \caption{Quantitative results for camera tracking in the three scenes of TUM RGBD\cite{sturm12tum}. We calculate the ATE RMSE$\downarrow$ [cm] for camera tracking in the three scenes.}
      \label{tab:tum_tracking}
      \vspace{-8mm}
\end{table}

\noindent\noindent
\textbf{Evaluation on TUM RGBD\cite{sturm12tum}.}
Unlike the other two datasets, the three sequences of TUM RGBD\cite{sturm12tum} are desktop scenes containing numerous objects. 
In terms of tracking performance in TUM\cite{sturm12tum}, our method is capable of producing competitive tracking results. 
As shown in \cref{tab:tum_tracking}, our method accurately represents the coordinates in indoor tabletop-level scenes by integrating orthogonal features and parameter sparsification using sparse tri-plane encoding, enabling precise camera tracking. 
NICE-SLAM\cite{zhu2022nice} struggles to generate high-fidelity object colors and contours due to limitations in hierarchical grid resolution and the absence of an efficient feature retrieval strategy. Co-SLAM\cite{wang2023coslam} struggles to represent detailed desktop-level objects, constrained by joint encoding, leading to geometry distortions. Although Point-SLAM\cite{sandstrom2023pointslam} excels in generating intricate textures, it cannot fully complete object contours. 

\begin{figure*}[!t]
      \centering
      \captionsetup[subfigure]{labelformat=empty}
      
      \begin{subfigure}{0.47\linewidth}
            \includegraphics[width=1\textwidth]{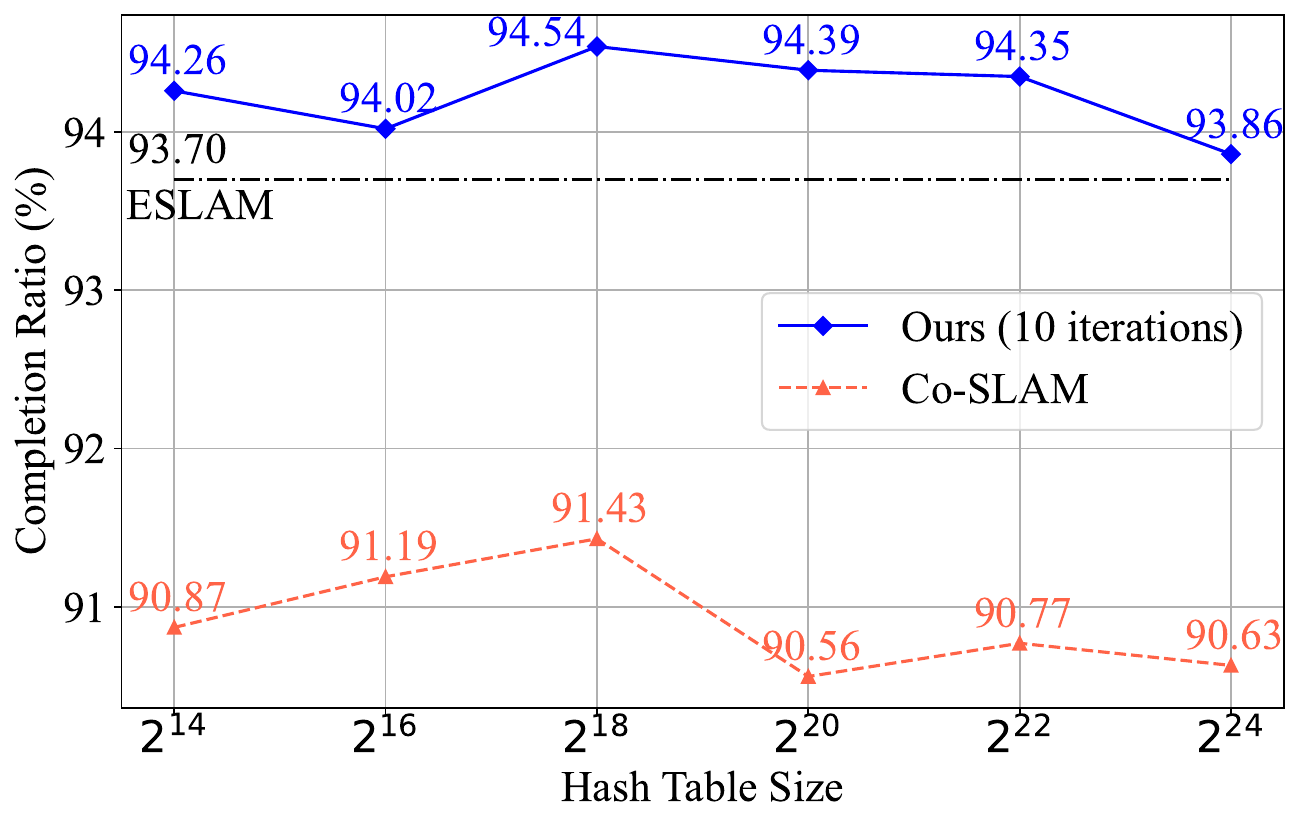}
      \end{subfigure}
      \begin{subfigure}{0.48\linewidth}
            \includegraphics[width=1\textwidth]{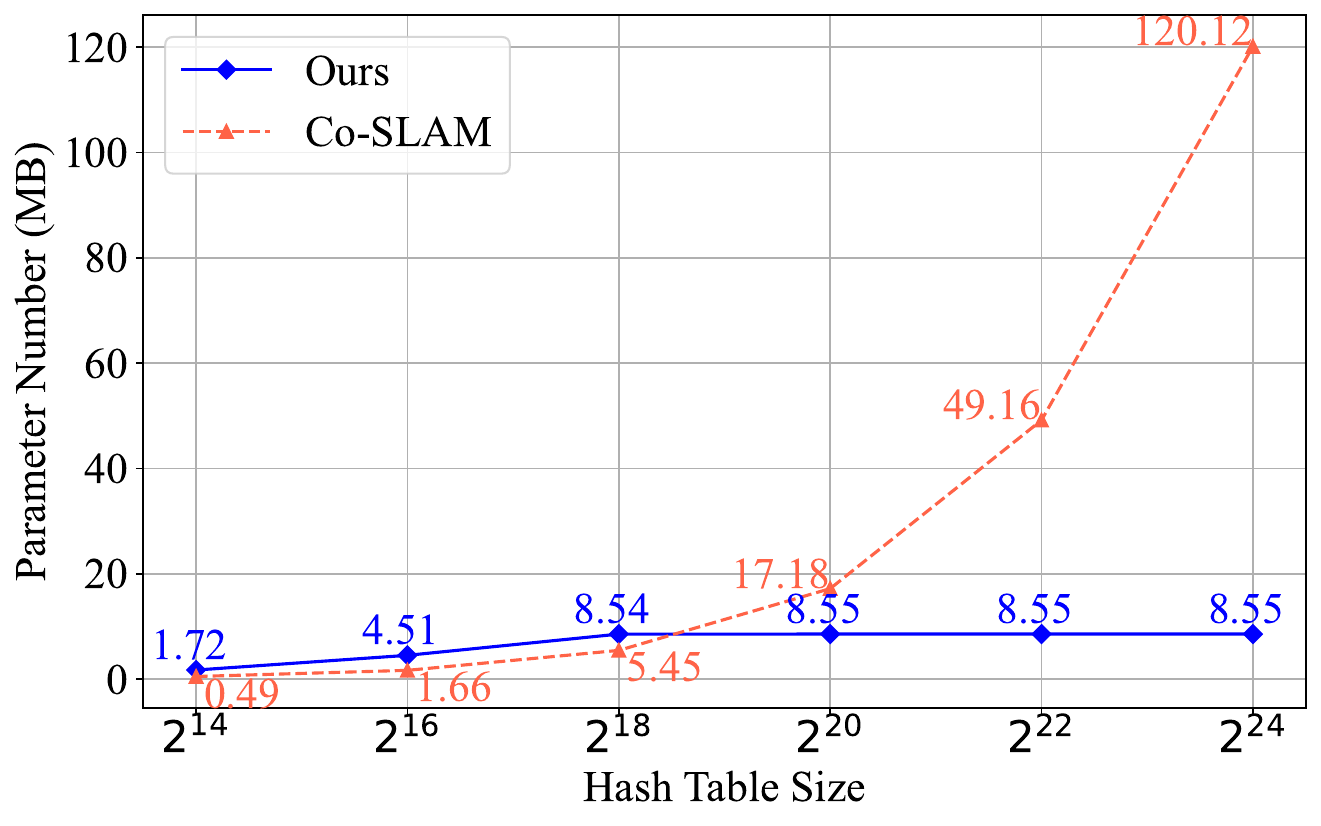}
      \end{subfigure}
      \vspace{-4mm}
      \caption{Taking office4 scene as an example, the performance of our method far surpasses Co-SLAM as the hash table size increases, and the growth of parameters is slow.}
      \label{fig:effect_s3slam_co}
      \vspace{-4mm}
\end{figure*}

\begin{figure*}[!t]
      \centering
      \captionsetup[subfigure]{labelformat=empty}
      \begin{subfigure}{0.48\linewidth}
            \includegraphics[width=1\textwidth]{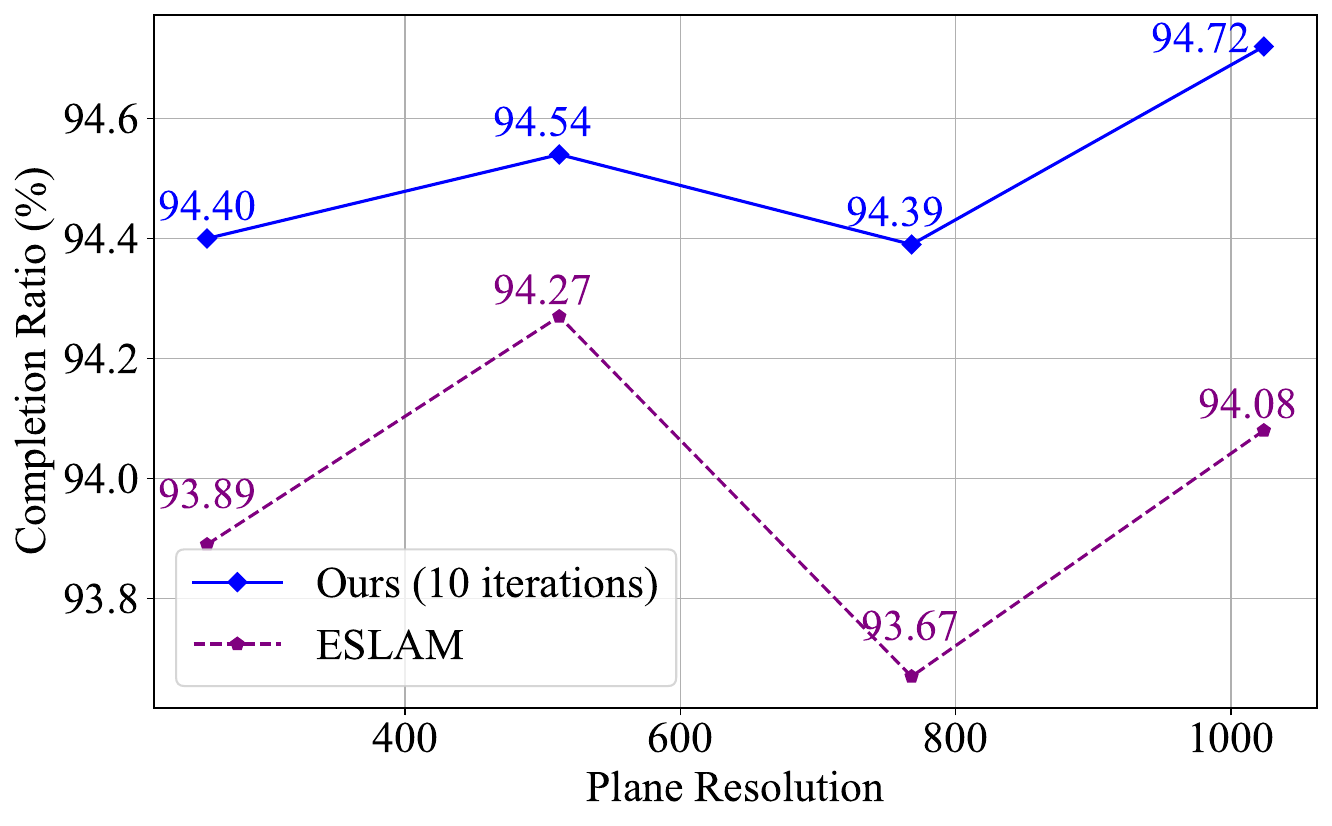}
      \end{subfigure}
      \begin{subfigure}{0.47\linewidth}
            \includegraphics[width=1\textwidth]{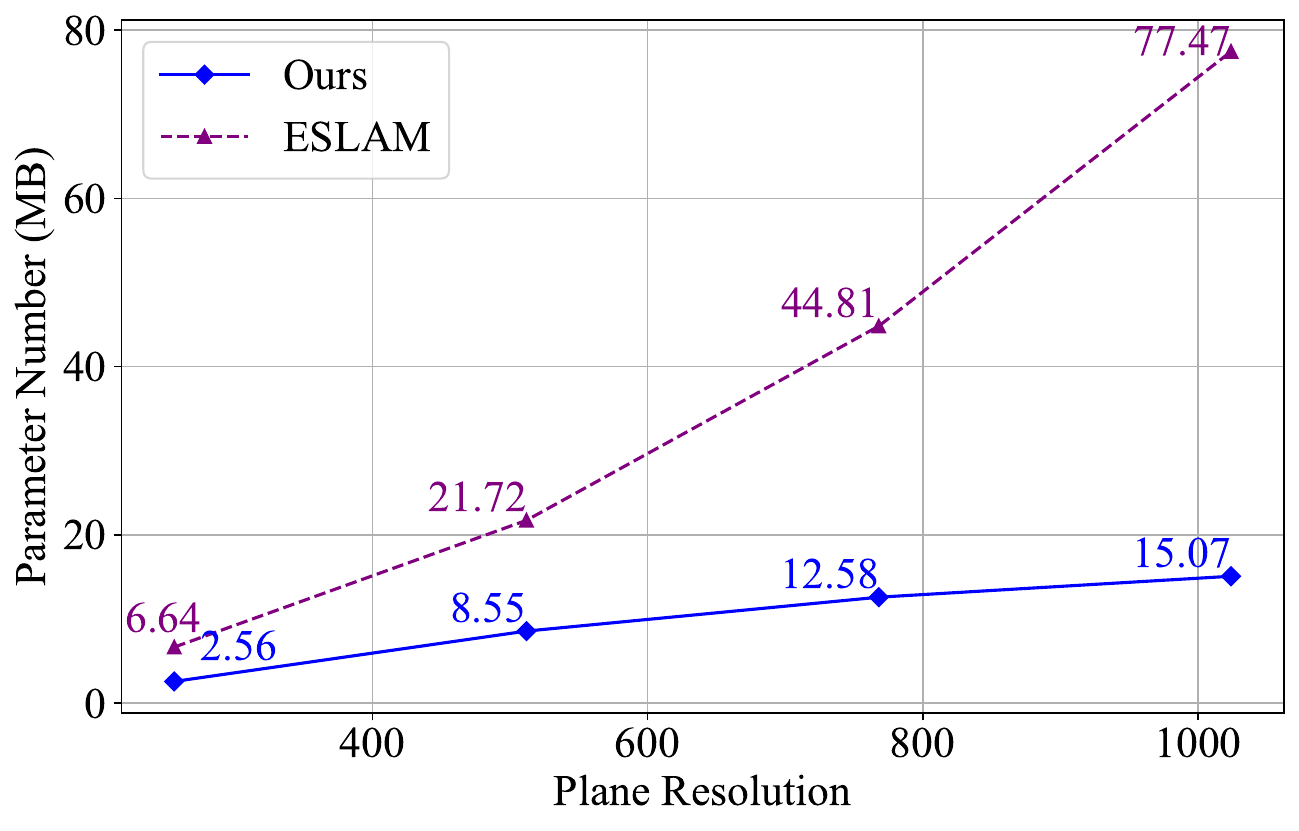}
      \end{subfigure}

      \vspace{-4mm}
      \caption{In office4, as the resolution of 2D planes increases, our method demonstrates superior reconstruction completeness compared to ESLAM using the original tri-plane.}
      \label{fig:effect_s3slam}
      \vspace{-5mm}
\end{figure*}

\subsection{Effectiveness of Sparse Tri-plane}
\textbf{Global Geometry Consistency.} Unlike 3D grids limited to small-range points, sparse tri-plane projects all 3D points onto three orthogonal planes consisting of 2D hash-grids, offering a global representation. The global feature allows direct utilization of learned projection plane features to extend characteristics to unseen 3D points along orthogonal axes. 
As shown in \cref{tab:triplane_eslam}, compared to 3D hash grids only representing points within a local range, our encoding employs projection to represent unseen 3D points extended along a specific dimension, leveraging the features of the corresponding projection plane to obtain more complete and accurate scene geometry.

\noindent\noindent
\textbf{Parametric Efficiency.} 
In \cref{fig:effect_s3slam_co}, Co-SLAM\cite{wang2023coslam} utilizes sparse 3D hash grids and coordinate encoding for achieving a complact and effective scene representation, but it remains insufficient to maintain all 3D grid vertices feathers by hash tables, leading to a continuous significant growth in the parameter count of Co-SLAM. In contrast, our method utilizes the projection property to represent all 3D grids with 2D planes, reducing the parameter count from cubic to square levels. Additionally, the parameters are further sparsified through the utitity of hash tables. It is noteworthy that due to the substantial reduction in parameter count facilitated by our 2D hash grids, even with an enlarged hash table, our parameter count remains unchanged. In \cref{fig:effect_s3slam}, regarding 2D plane resolution, ESLAM\cite{johari2023eslam}, employing tri-plane encoding, encounters a rapid surge in parameters with increasing resolution. In contrast, our method using $2^{18}$-length hash tables effectively reduces parameter count through the utilization of hash tables and attains outstanding performance via multi-resolution 2D hash grids.
      
\begin{table}[t]
      \scriptsize
      \centering
            \setlength{\tabcolsep}{1.6mm}{
            \begin{tabular}{lccccc}
            \toprule
            Method & Depth L1 (cm)$\downarrow$ & Acc. (cm)$\downarrow$ & Comp. (cm)$\downarrow$ & Comp. Ratio (\%)$\uparrow$ \\
            \midrule
            Ours (3D Hash Grid) & 1.47 & 2.16 & 1.98 & 94.37\\  
            Ours & \textbf{0.98} & \textbf{2.11} & \textbf{1.66} & \textbf{96.71}\\
            \bottomrule
            \end{tabular}}
      \caption{
            Quantitative comparison results on the Replica\cite{straub2019replica} dataset, where each metric represents the average value across all scenes in the datasets. Compared to our method using 3D hash grids, our S3-SLAM achieves high-quality reconstruction.
      }
      \vspace{-6mm}
      \label{tab:triplane_eslam}
\end{table}

\begin{table}[t]
        \scriptsize
        \centering
        \setlength{\tabcolsep}{1.9mm}{
        \begin{tabular}{l|l|cc|c}
        \toprule
        & Method & Track. (ms) $\times$ $its$ $\downarrow$ & Map. (ms) $\times$ $its$ $\downarrow$ & Param. (MB)$\downarrow$\\
        \midrule
        \multirow{4}*{\rotatebox[origin=c]{90}{\scriptsize{ScanNet}}} &NICE-SLAM\cite{zhu2022nice} & 15.2$\times$50 & 200.3$\times$60 & 10.1 \\
        & Co-SLAM\cite{wang2023coslam} & 9.1$\times$10 & 28.6$\times$10 & 1.7 \\
        & Point-SLAM\cite{sandstrom2023pointslam} & 27.5$\times$100 & 27.6$\times$300 & 20+ \\
        & Ours (origin tri-plane) & 26.2$\times$15 & 94.1$\times$15 & 106.4\\
        & Ours & 17.5$\times$15 & 40.8$\times$15 & 4.2\\
        \midrule
        \multirow{4}*{\rotatebox[origin=c]{90}{\scriptsize{TUM}}} &NICE-SLAM\cite{zhu2022nice} & 61.4$\times$200 & 264.9$\times$60 & 103.1  \\
        & Co-SLAM\cite{wang2023coslam} & 8.9$\times$10 & 26.8$\times$10 & 1.7\\
        & Point-SLAM\cite{sandstrom2023pointslam}  & 25.2$\times$200 & 27.4$\times$150 & 20+ \\
        & Ours (origin tri-plane) & 21.6$\times$10 & 34.2$\times$10 & 99.8\\
        & Ours & 15.7$\times$10 & 30.7$\times$10 & 2.9\\
        \bottomrule
        \end{tabular}}
        \caption{Our S3-SLAM utilizes sparse tri-plane and achieves high-quality mapping with a few iterations ($its$). In terms of runtime, S3-SLAM outperforms most baselines.}
        \vspace{-6mm}
        \label{tab:performance}
      \end{table}

\subsection{Performance Analysis}
In our experimental design, to demonstrate the fitting speed of our method, the tracking and mapping iterations of S3-SLAM with a few iterations. As shown in \cref{tab:performance}, NICE-SLAM\cite{zhu2022nice} utilizes dense hierarchical grids to represent the scene, with the parameter count growing cubically. Point-SLAM\cite{sandstrom2023pointslam} employs an explicit structure of surface point clouds to build structural features based on the scene surface. However, this approach restricts scene representation strictly to observed regions, leading to local overfitting and an inability to render unobserved regions. 
Co-SLAM\cite{wang2023coslam} significantly reduces storage consumption by leveraging sparse parametric encoding, but Co-SLAM lacks fine-level appearance and geometry reconstruction. 
Origin tri-plane\cite{gao2022get3d} drastically reduces the parameter count by representing the three-dimensional scene with three orthogonal planes. However, when the reconstruction resolution reaches 512, the tri-plane parameter count will reach around 100MB, as shown in \cref{tab:performance}.
Surprisingly, thanks to the hashing properties introduced by the hash grids, our sparse tri-plane encoding only requires around 2$\sim$4MB parameters to represent a 512-resolution scene.
Due to sparsity of spatial hash function, the parameter count of our sparse tri-plane encoding is only 2$\sim$4\% of the tri-plane encoding\cite{gao2022get3d} in the 512-resolution scene reconstruction.

\subsection{Ablations}
\begin{figure}[!t]
  \centering
   \includegraphics[width=1\linewidth]{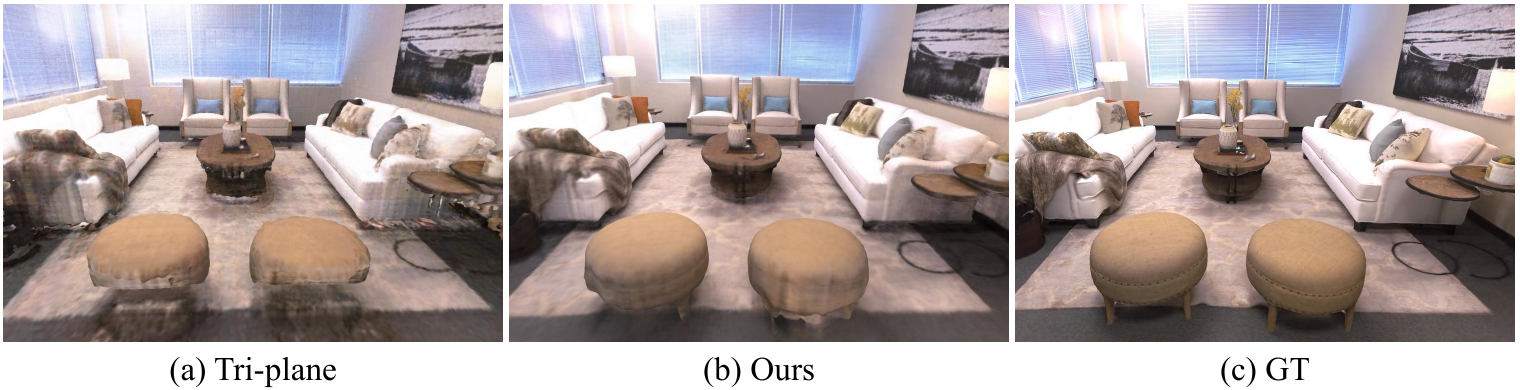}
   \vspace{-7mm}
   \caption{A comparison of the reconstruction quality between our sparse tri-plane encoding and tri-plane encoding\cite{gao2022get3d} demonstrates that our sparse tri-plane encoding can reconstruct smoother surface geometry and high-quality appearance.}
   \vspace{-6mm}
   \label{fig:ab}
\end{figure}

\noindent\noindent
\textbf{Sparse Tri-plane Encoding.}
Compared with the conventional tri-plane encoding\cite{gao2022get3d}, our method utilizes hierarchical representation and spatial hash function, resulting in a more compact scene representation, efficiently encoding the geometry and appearance. Regarding memory consumption, as shown in \cref{tab:performance}, with a 512-resolution reconstruction, the parameter count of our sparse tri-plane encoding decreases to 2$\sim$4\% of that in tri-plane encoding\cite{gao2022get3d}. Regarding reconstruction quality, our sparse tri-plane encoding can reconstruct higher-resolution geometry and appearance. As shown in \cref{fig:ab}, our sparse tri-plane encoding can predict complete geometry of subtle regions in corners, whereas tri-plane encoding results in artifacts.

\begin{table}[!t]
      \scriptsize
      \centering
            \setlength{\tabcolsep}{2.8mm}{
            \begin{tabular}{llccccccc}
            \toprule
            & Method & 0000 & 0059 & 0106 & 0169 & 0181 & 0207 & Avg.\\
            \midrule
            & w/o HBA 
            & 17.66 & 14.72 & 21.81 
            & 27.96 
            & 23.22 & 21.26 & 21.11\\    
            & w HBA & \textbf{6.31} & \textbf{9.84} & \textbf{8.02} & \textbf{5.81} & \textbf{12.45} & \textbf{6.41} & \textbf{8.14}\\
            \bottomrule
            \end{tabular}}
      \caption{We conduct comparative experiments on the ScanNet dataset between S3-SLAM without (w/o) and with (w) HBA. For tracking evaluation metric, we calculate the ATE RMSE$\downarrow$ [cm] for camera tracking in the six scenes. Due to the incorporation of HBA, S3-SLAM can predict more accurate camera poses.}
      \vspace{-6mm}
      \label{tab:scannet_ab}
    \end{table}

\noindent\noindent
\textbf{Hierarchical Bundle Adjustment.}
We conduct quantitative experiments on the ScanNet\cite{dai2017scannet} dataset, and the global-to-local structural optimization introduced by HBA results in more accurate pose estimation. In the experiments, we randomly sample 2048 rays from all keyframes. We set our re-projected keyframe selection threshold as 0.8, and the local sliding window optimizes the twenty newest keyframes. As shown in \cref{tab:scannet_ab}, the quantitative tracking result demonstrates that our HBA enhances the camera tracking accuracy of S3-SLAM. 
We also conduct supplementary qualitative experiments on S3-SLAM by adding HBA and other excellent works, as shown in \cref{fig:cover}. After incorporating HBA, S3-SLAM generates appearance closer to the ground truth.

\section{Conclusion}
We propose an efficient and compact sparse tri-plane encoding and develop a SLAM for high-quality reconstruction of fine-level appearance and geometry. To encode complex scenes more effectively, we integrate multi-level sparse tri-planes consisting of 2D hash-grids. Experimental results demonstrate that our S3-SLAM converges rapidly and can reconstruct high-fidelity scenes. Additionally, to achieve fine-grained local appearance, we design HBA to refine the reconstruction of local appearances while maintaining the quality of global mapping.
\textbf{Future work.} 
Although our approach achieves a high level of geometric completeness through relatively compact global features, implementing genuine local updates is crucial to mitigate the forgetting issue. In the next phase of our work, we will implement local updates to further address the forgetting problem.

\par\vfill\par

\bibliographystyle{splncs04}
\bibliography{main}

\setcounter{section}{0}
\renewcommand\thesection{\Alph{section}}
\section{Implementation Details}
\label{sec:implementation}
\textit{\textbf{Experimental Details.}}
We design our loss function as follows:
\begin{equation}
    \begin{split}
        \mathcal{L} = \lambda_{c} \mathcal{L}_{c} + \lambda_{d} \mathcal{L}_{d} + 
        \lambda_{sdf} \mathcal{L}_{sdf} +
        \lambda_{fs} \mathcal{L}_{fs}
        \label{render}
    \end{split}
\end{equation}
where $\lambda_{c}=5.0$, $\lambda_{d}=1.0$, $\lambda_{sdf}=10000.0$ and $\lambda_{fs}=10.0$. 
Regarding the number of sampled points for every dataset, we adopt our depth-guided sampling strategy to sample 64 points for every ray on Replica\cite{straub2019replica}, and TUM\cite{sturm12tum}. Additionally, because ScanNet\cite{dai2017scannet} converges more challengingly, we sample 128 points on each ray. We use the Adam optimizer, setting the learning rate for pose $\lambda_{pose}=0.005$ and the learning rate for MLP decoders parameters $\lambda_{decoder}=0.010$ and sparse tri-plane encoding $\lambda_{encoding}=0.010$.

\vspace{+4mm}
\noindent\noindent
\textit{\textbf{Tracking and Mapping.}}
Our tracking methodology resembles other neural implicit SLAM approaches, utilizing the loss function during Bundle Adjustment (BA) to optimize the camera pose. We chose not to adopt parallel programming, unlike other excellent works. In pursuing higher accuracy, we adopt serial programming, conducting tracking before mapping. Initially, we initialize the pose of the first frame and perform 400-600 mapping module iterations to ensure a robust initialization for our sparse tri-plane encoding. Within the tracking module, for each input RGB-D observation, we provide an initial estimated pose using a constant-speed camera mortion. Subsequently, we focus solely on optimizing the frame pose, minimizing the rendering losses to update the frame pose. In the mapping module, following the tracking finishes, we adopt our keyframe selection strategy to determine keyframe sequence. We subsequently perform our HBA to optimize frame poses and encoding and decoder parameters jointly.

\begin{figure*}[!t]
    \vspace{-3mm}
    \centering
    \captionsetup[subfigure]{labelformat=empty}

    \begin{subfigure}{0.32\linewidth}
          \centering
          \captionsetup{justification=raggedright}
          \caption{\fontsize{9pt}{1pt}
          \ttfamily w/o HBA}
          \includegraphics[width=1\textwidth]{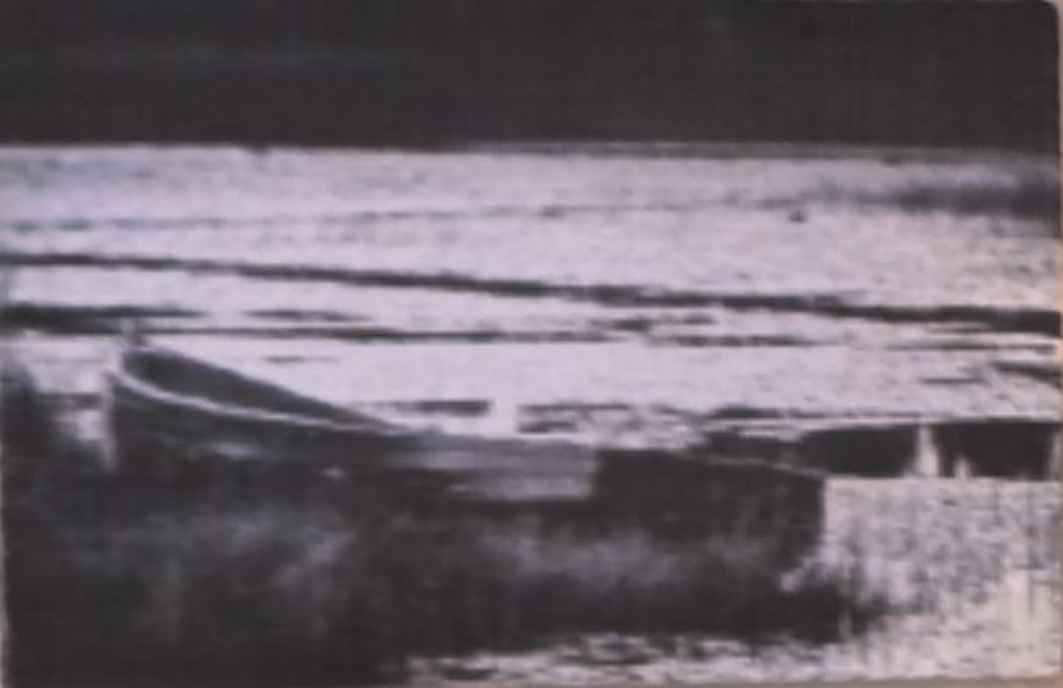}
          \label{room_0_coslam}
    \end{subfigure}
    \begin{subfigure}{0.32\linewidth}
          \centering
          \captionsetup{justification=raggedright}
          \caption{\fontsize{9pt}{3pt}\ttfamily w HBA}
          \includegraphics[width=1\textwidth]{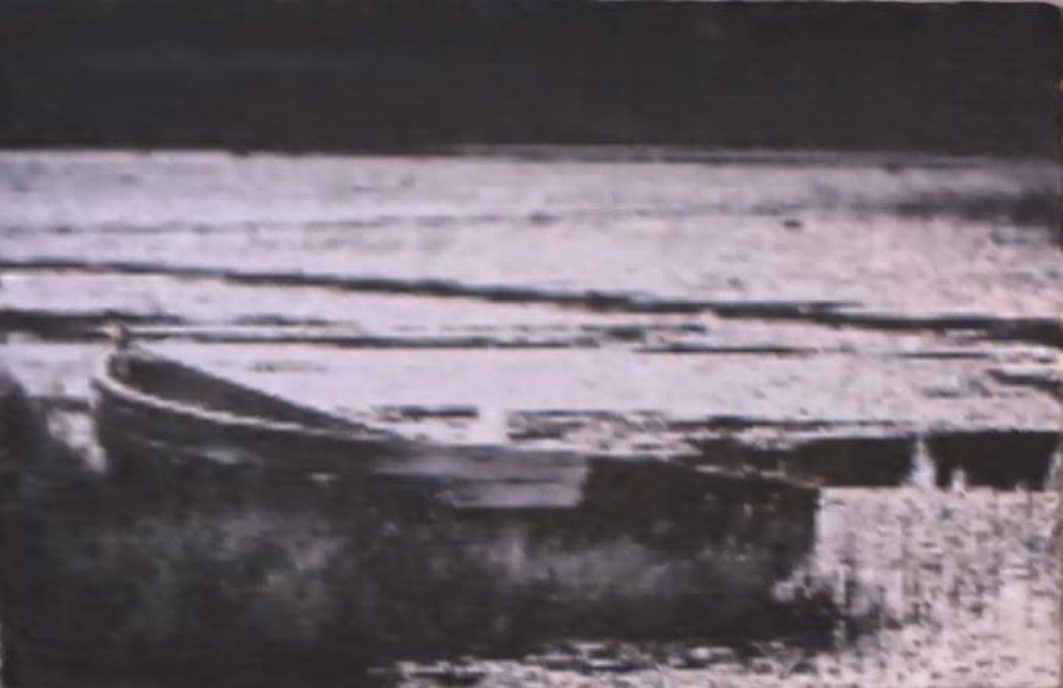}
          \label{room_2_coslam}
    \end{subfigure}
    \begin{subfigure}{0.32\linewidth}
          \captionsetup{justification=raggedright}
          \caption{\fontsize{9pt}{3pt}\ttfamily GT}
          \includegraphics[width=1\textwidth]{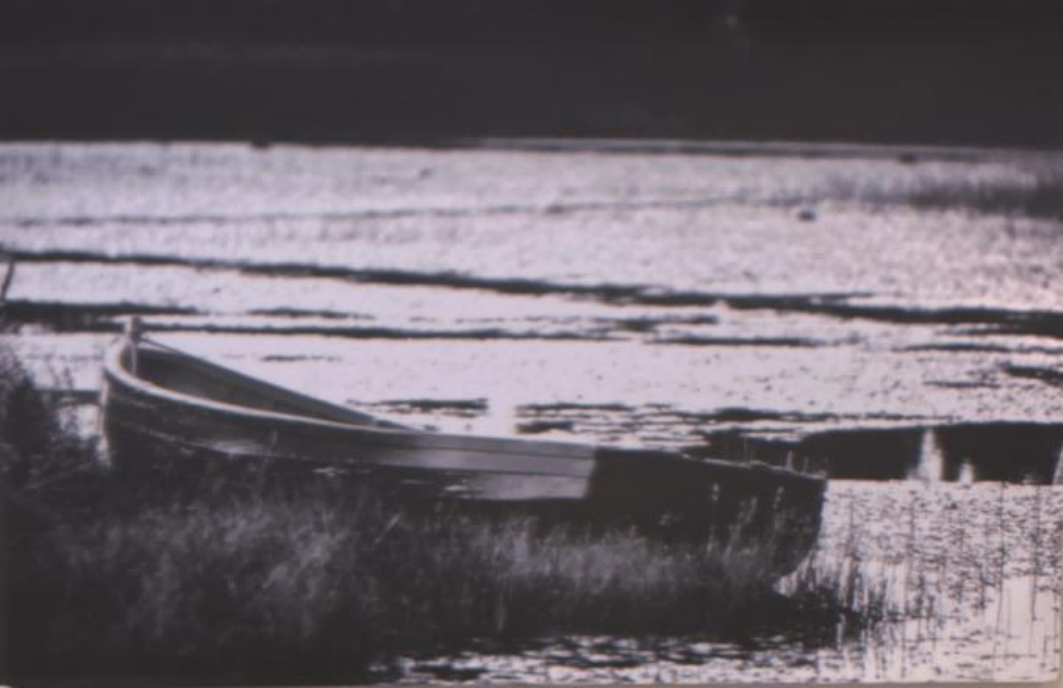}
          \label{office_3_coslam}
    \end{subfigure}
    
    \vspace{-4mm}
    \begin{subfigure}{0.32\linewidth}
          \includegraphics[width=1\textwidth]{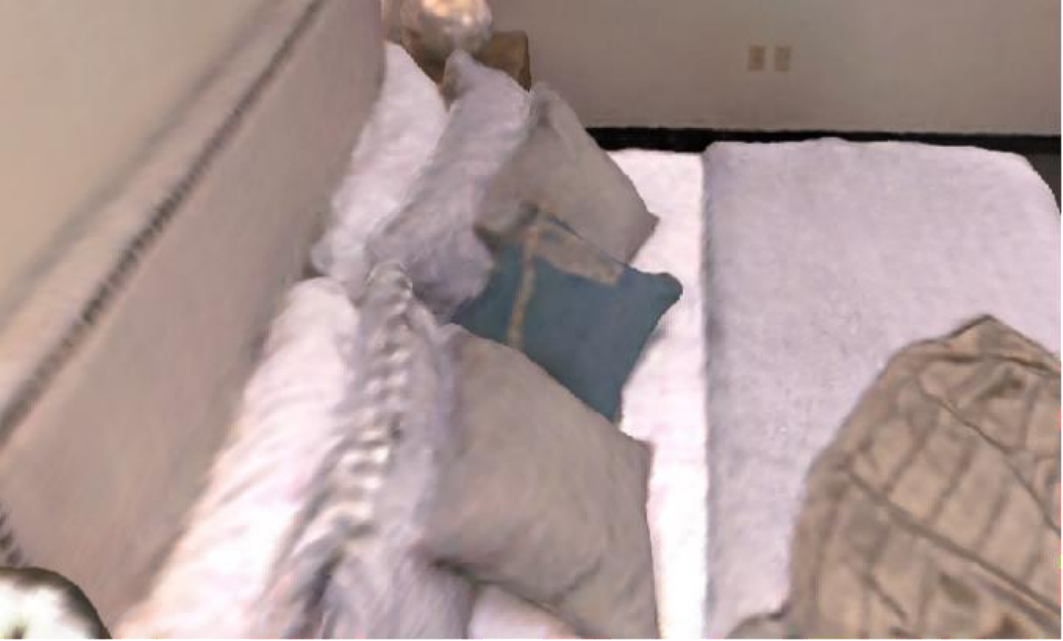}
          \label{room0_pointslam}
    \end{subfigure}
    \begin{subfigure}{0.32\linewidth}
          \includegraphics[width=1\textwidth]{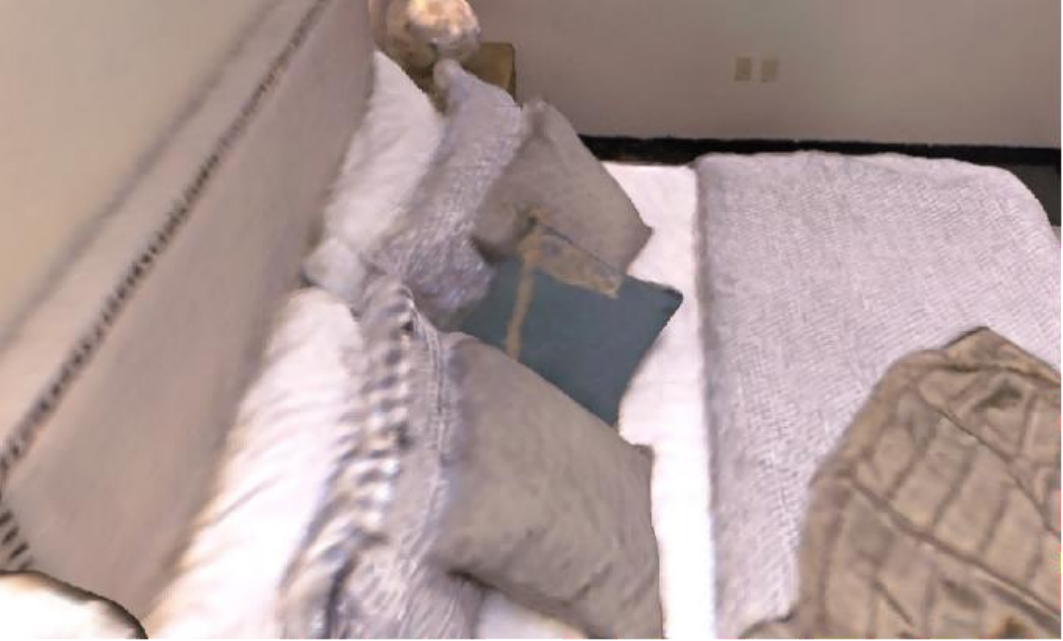}
          \label{room2_pointslam}
    \end{subfigure}
    \begin{subfigure}{0.32\linewidth}
          \includegraphics[width=1\textwidth]{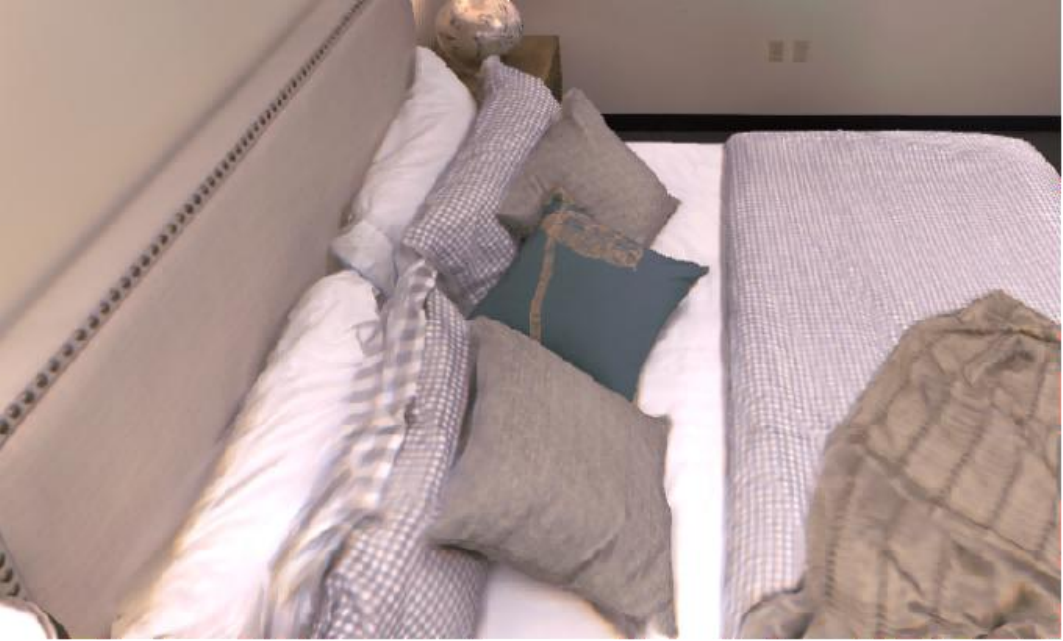}
          \label{office3_pointslam}
    \end{subfigure}

    \vspace{-4mm}
    \begin{subfigure}{0.32\linewidth}
          \includegraphics[width=1\textwidth]{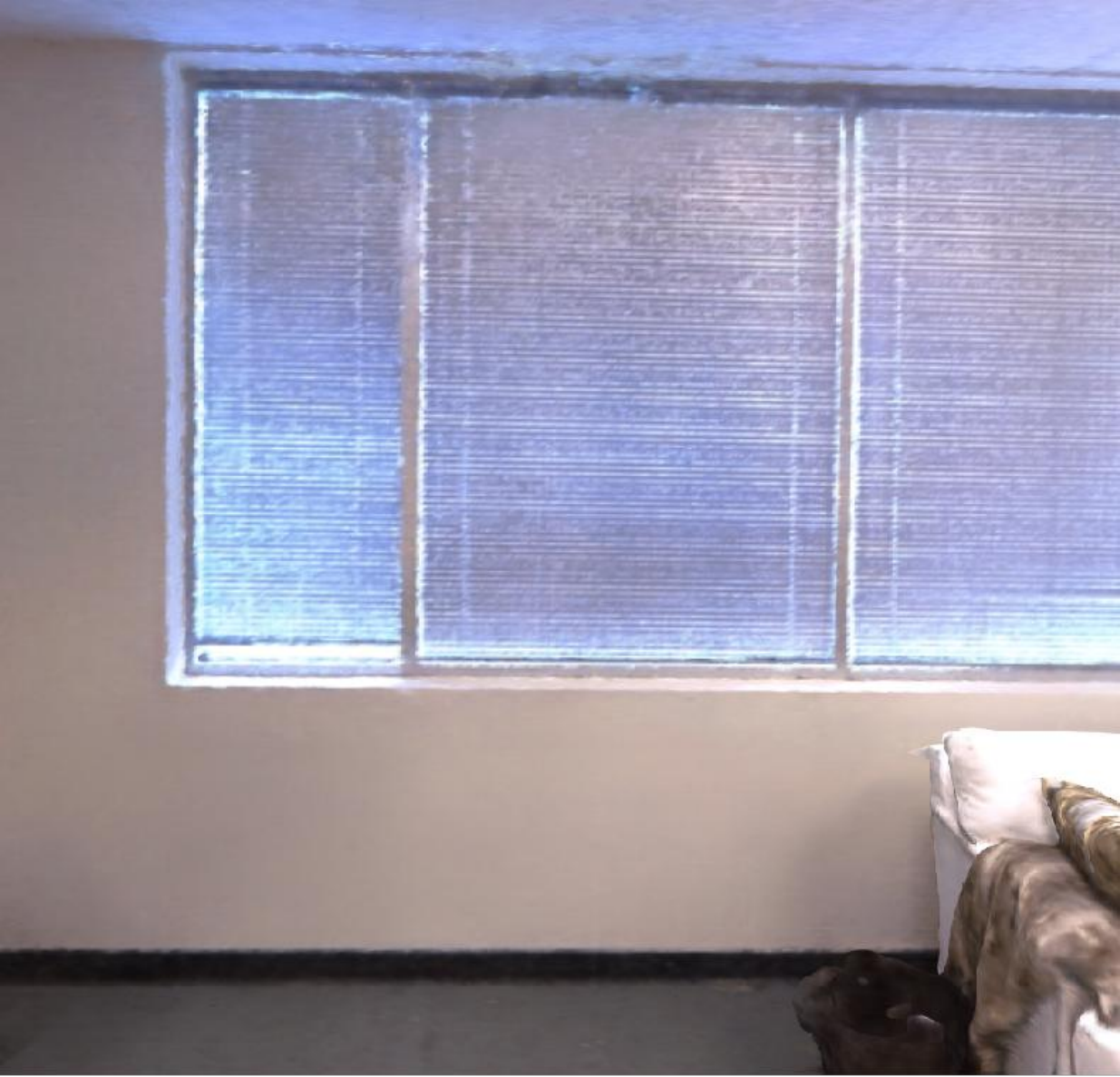}
          \label{room0_ours}
    \end{subfigure}
    \begin{subfigure}{0.32\linewidth}
          \includegraphics[width=1\textwidth]{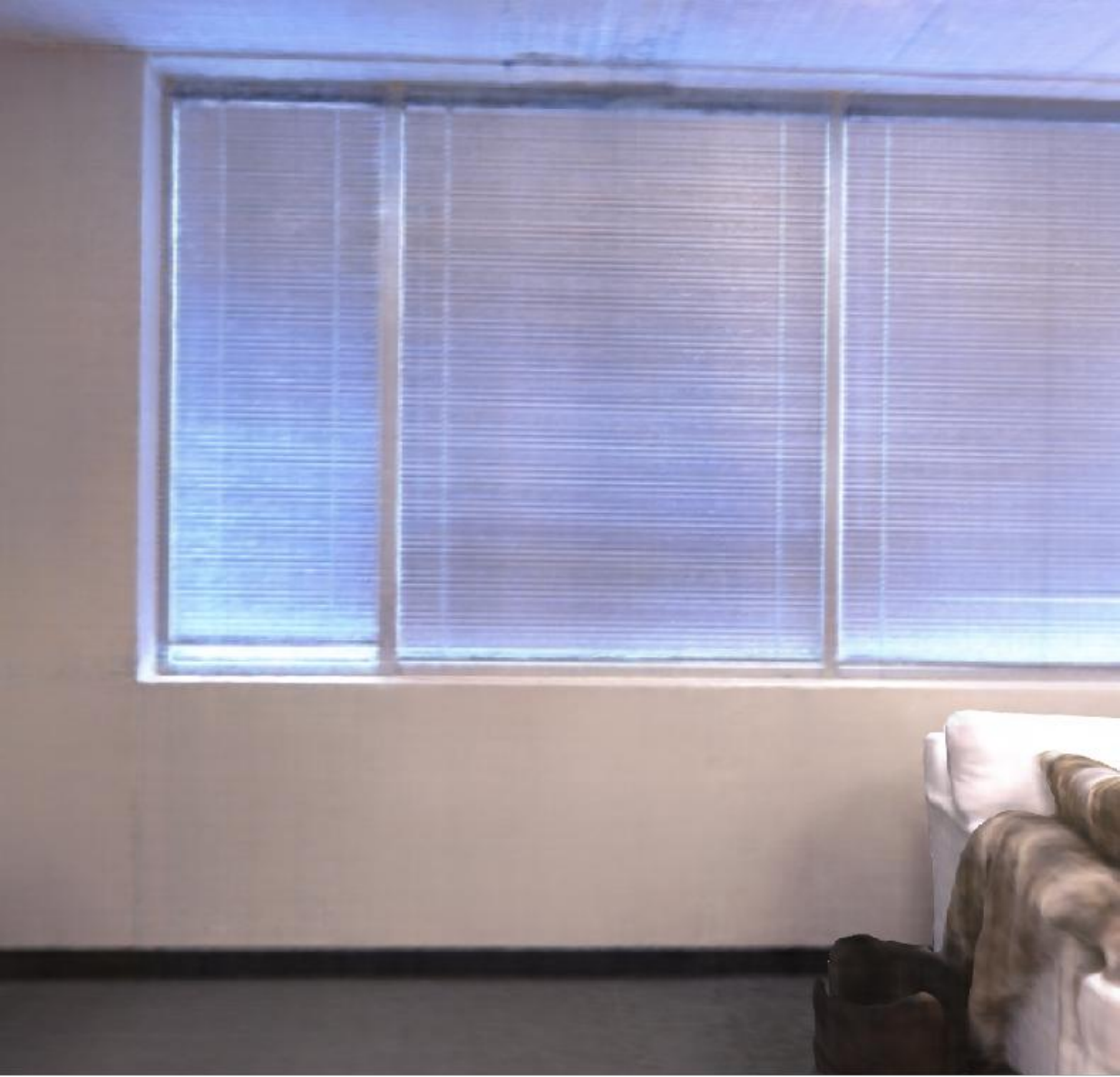}
          \label{room2_ours}
    \end{subfigure}
    \begin{subfigure}{0.32\linewidth}
          \includegraphics[width=1\textwidth]{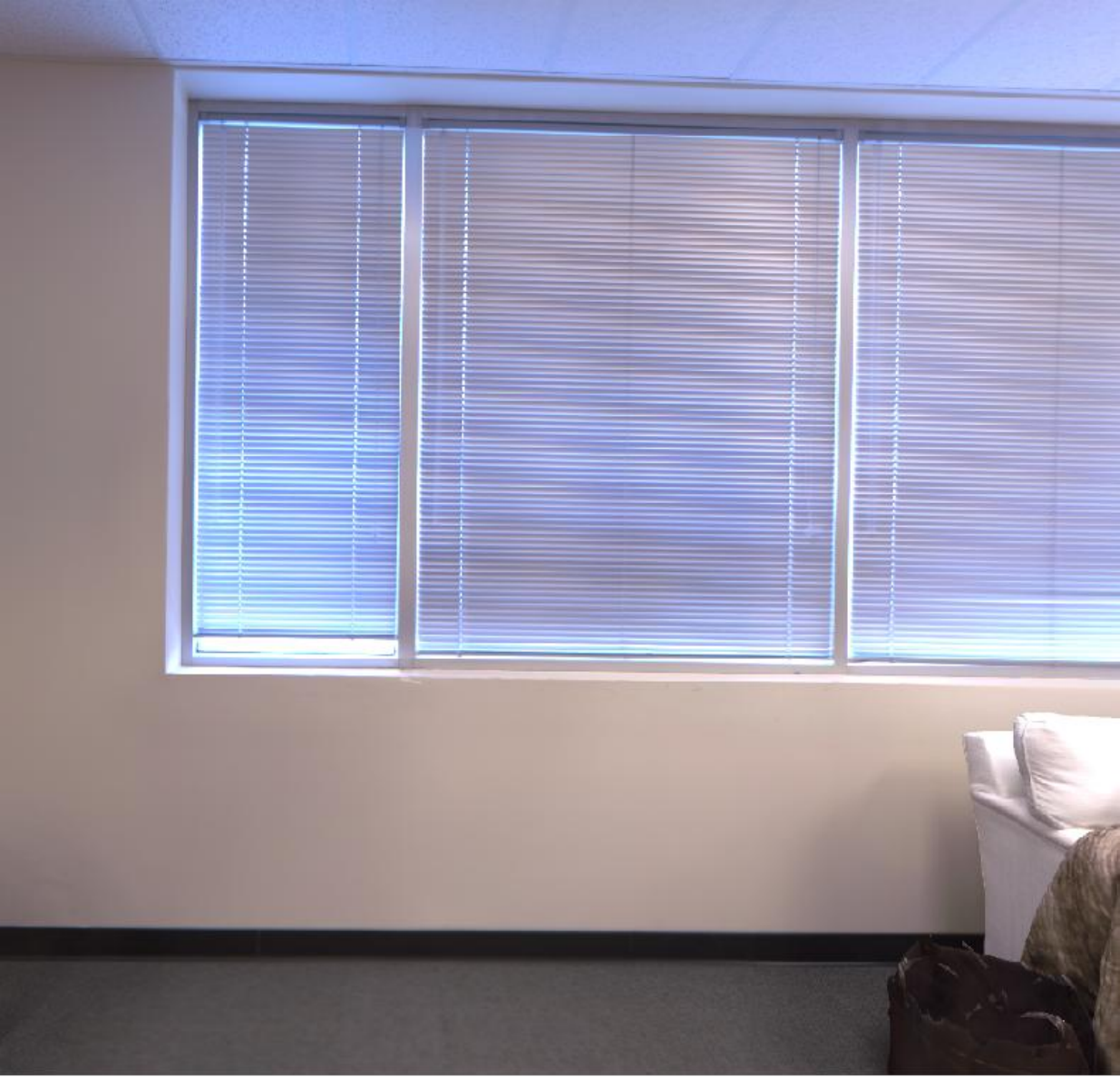}
          \label{office3_ours}
    \end{subfigure}

    \vspace{-6mm}
    \caption{The qualitative results indicate that the HBA implemented in our S3-SLAM enhances the reconstruction of fine-grained local appearance details. For example, HBA achieves improved details on murals, bed decorations, and blanket patterns.}
     \label{fig:hba_ab}
     \vspace{-2mm}
\end{figure*}

\section{2D Hash Grid}
\indent\indent
Sparse parametric encodings mainly include two basic approaches: multi-resolution hash encoding\cite{muller2022instantngp} and permutohedral lattice encoding\cite{rosu2023permutosdf}. We extensively use multi-resolution hash encoding\cite{muller2022instantngp} because of its ability to compactly represent scenes and significantly reduce the training time of neural radiance fields (NeRF) by hash grids. Permutohedral lattice encoding demonstrates perfect performance in temporal dimensions, owing to its linear time complexity growth with the utilization of permutohedral lattice hash-grid. In our construction of the plane representation, as shown in \cref{tab:scannet_ab_sup}, we test both grid types. Experimental results indicate that the hash grid is more suitable than the lattice hash-grid for our SLAM design. We also attempt to enhance permutohedral lattice grids by high-frequency positional encoding and other methods, but hash grids still outperform these methods.

\begin{table}[!t]
      \scriptsize
      \centering
      \setlength{\tabcolsep}{1.1mm}{
      \begin{tabular}{llccccccc}
      \toprule
      & Method & 0000 & 0059 & 0106 & 0169 & 0181 & 0207 & Avg.\\
      \midrule
      & perm. enc. \cite{rosu2023permutosdf} 
            & 9.42 & 12.95 & 13.51 
            & 8.41 
            & 12.56 & 6.84 & 10.62 \\
      & hash. enc. \cite{muller2022instantngp} 
      & \textbf{6.31} & \textbf{9.84} & \textbf{8.02} & \textbf{5.81} & \textbf{12.45} & \textbf{6.41} & \textbf{8.14}\\
      \bottomrule
      \end{tabular}}
\caption{We quantitatively compare the tracking performance of permutohedral lattice-grid\cite{rosu2023permutosdf} and hash-grid\cite{muller2022instantngp} for our design on ScanNet\cite{dai2017scannet}. Experimental results demonstrate that the hash-grid\cite{muller2022instantngp} is better in the camera tracking for our design.}
\vspace{-8mm}
\label{tab:scannet_ab_sup}
\end{table}

\begin{table*}[t]
      \renewcommand{\arraystretch}{1.5}
      \scriptsize
      \centering
      \setlength{\tabcolsep}{0.9mm}{
      \begin{tabular}{llccccccccc}
      \toprule
      & Metric & \ttfamily room0 & \ttfamily room1 & \ttfamily room2 &\ttfamily office0 & \ttfamily office1 & \ttfamily office2 & \ttfamily office3 & \ttfamily office4 &  Avg.\\
      \midrule
      \multirow{5}*{\rotatebox[origin=c]{90}{\scriptsize{NICE-SLAM\cite{zhu2022nice}}}}
      & Depth L1 [cm]$\downarrow$  
      & 1.79 & 1.33 & 2.20 & 1.43 & 1.58 & 2.70 & 2.10 & 2.06 & 1.90\\
      & Acc. [cm]$\downarrow $ 
      & 2.44 & 2.10 & 2.17 & 1.85 & 1.56 & 3.28 & 3.01 & 2.54 & 2.37\\
      & Comp. [cm]$\downarrow $ 
      & 2.60 & 2.19 & 2.73 & 1.84 & 1.82 & 
      3.11 & 3.16 & 3.61 & 2.63\\
      & Comp. Ratio [\%]$\uparrow $ 
      & 91.81 & 93.56 & 91.48 & 94.93 & 94.11 & 88.27 & 87.68 & 87.23 & 91.13\\
      & Param. [MB]$\downarrow $ & 12.20 & 8.79 & 30.20 & 22.90 & 21.60 & 11.10 & 15.20 & 17.20 & 17.40 \\
      \midrule
      \multirow{5}*{\rotatebox[origin=c]{90}{\scriptsize{Co-SLAM\cite{wang2023coslam}}}}
      & Depth L1 [cm]$\downarrow$  
      & 1.05 & 0.85 & 2.37 & 1.24 & 1.48 & 1.86 & 1.66 & 1.54 & 1.51\\
      & Acc. [cm]$\downarrow $ 
      & 2.11 & 1.68 & 1.99 & 1.57 & 1.31 & 2.84 & 3.06 & 2.23 & 2.10\\
      & Comp. [cm]$\downarrow $ 
      & 2.02 & 1.81 & 1.96 & 1.56 & 1.59 & 2.43 & 2.72 & 2.52 & 2.08\\
      & Comp. Ratio [\%]$\uparrow $ 
      & 95.26 & 95.19 & 93.58 & 96.09 & 94.65 & 91.63 & 90.72 & 90.44 & 93.44\\
      & Param. [MB]$\downarrow $ 
      & \textbf{1.72} & \textbf{1.72} & \textbf{1.72} & \textbf{1.72} & \textbf{1.72} & \textbf{1.72} & \textbf{1.72} & \textbf{1.72} & \textbf{1.72} \\
      \midrule
      \multirow{5}*{\rotatebox[origin=c]{90}{\scriptsize{Point-SLAM\cite{sandstrom2023pointslam}}}}
      & Depth L1 [cm]$\downarrow$  
      & 3.21 & 4.93 & 5.00 & 2.14 & 3.42 & 5.13 & 2.92 & 4.63 & 3.92\\
      & Acc. [cm]$\downarrow $ 
      & \textbf{1.45} & \textbf{1.15} & \textbf{1.20} & \textbf{1.05} & \textbf{0.88} & \textbf{1.32} & \textbf{1.55} & \textbf{1.52} & \textbf{1.26}\\
      & Comp. [cm]$\downarrow $ 
      & 3.42 & 3.01 & 2.65 & 1.65 & 2.18 & 3.63 & 3.18 & 3.89 & 2.95\\
      & Comp. Ratio [\%]$\uparrow $ 
      & 88.64 & 89.57 & 90.13 & 93.41 & 90.93 & 86.34 & 87.72 & 86.06 & 89.10\\
      & Param. [MB]$\downarrow $ 
      & 50.71 & 35.40 & 36.80 & 28.12 & 19.87 & 41.91 & 59.99 & 49.41 & 40.28\\
      \midrule
      \multirow{5}*{\rotatebox[origin=c]{90}{\scriptsize{Ours}}}
      & Depth L1 [cm]$\downarrow$  
      & \textbf{0.50} & \textbf{0.72} & \textbf{1.56} & \textbf{0.88} & \textbf{0.97} & \textbf{1.18} & \textbf{1.03} & \textbf{0.99} & \textbf{0.98}\\
      & Acc. [cm]$\downarrow $ 
      & 2.32 & 2.17 & 1.81 & 1.63 & 1.49 & 2.83 & 2.61 & 2.03 & 2.11 \\
      & Comp. [cm]$\downarrow $ 
      & \textbf{1.73} & \textbf{1.63} & \textbf{1.62} & \textbf{1.29} & \textbf{1.16} & \textbf{1.83} & \textbf{2.04} & \textbf{1.97} & \textbf{1.66} \\
      & Comp. Ratio [\%]$\uparrow $ 
      & \textbf{97.75} & \textbf{96.02} & \textbf{95.47} & \textbf{98.40} & \textbf{98.03} & \textbf{96.10} & \textbf{95.98} & \textbf{95.96} & \textbf{96.71}\\
      & Param. [MB]$\downarrow $ 
      & 4.51 & 4.51 & 4.51 & 4.51 & 4.51 & 4.51 & 4.51 & 4.51 & 4.51\\
      \bottomrule
      \end{tabular}
      }
      \caption{Quantitative results for reconstruction in the eight scenes of Replica\cite{straub2019replica}. We report the Depth L1 [cm]$\downarrow$, Acc. [cm]$\downarrow$ and Comp. [cm]$\downarrow$, Comp. Ratio [\%]$\uparrow$ and Param. [MB]$\downarrow$ for reconstruction in the eight scenes of Replica\cite{straub2019replica}. 
      Additionally, we calculate the average values of five metrics.
      Our method accurately reconstructs depth and geometry while completing a large amount of unseen regions.
      }
      \vspace{-8mm}
      \label{replica_recon}
\end{table*}

\section{More Experiments}
\textit{\textbf{Quantitative Experimentation Supplement.}}
We supply the quantitative comparison results for each scene in the Replica\cite{straub2019replica}. As shown in \cref{replica_recon} demonstrates that, due to our sparse tri-plane encoding, our method achieves a high completion ratio without sacrificing the accuracy of the reconstruction.

To validate the reconstruction effectiveness of our sparse tri-plane encoding, we show comparative results with ESLAM\cite{johari2023eslam}. ESLAM achieves high-quality reconstruction using a tri-plane encoding. As shown in \cref{replica_recon_eslam}, we conduct quantitative comparisons on Replica. The results show that our method produces higher reconstruction accuracy and completion. More importantly, we achieve high-quality reconstruction with only 48.92\% of the parameter count used by ESLAM.

To further demonstrate that our sparse tri-plane encoding can significantly reduce parameter count while ensuring high-quality reconstruction results, as shown in \cref{triplane_recon_eslam}, we conduct detailed quantitative experiments in the office4 scene of Replica\cite{straub2019replica}. In order to showcase the superiority of our method, we conduct a comparison with ESLAM, which also utilizes a tri-plane encoding. We also test the performance of two high-resolution feature plane methods to emphasize the performance of our sparse tri-plane encoding at high-resolution reconstruction. 
Firstly, we increase the plane resolution of ESLAM to the 1000+ level. Specifically, we triple the fine-level plane resolution of ESLAM, achieving a 1024 $\times$ 1024 resolution tri-plane encoding. We utilize conventional 1024 $\times$ 1024 resolution tri-plane encoding. As shown in \cref{triplane_recon_eslam}, the experimental results demonstrate that our S3-SLAM achieves high-quality reconstruction with fewer parameter by adopting sparse tri-plane encoding. The hash table size of our sparse tri-plane encoding solely determines the parameter count. The spatial hash mapping allows us to achieve high-quality reconstruction with minimal parameters in Replica\cite{straub2019replica}. Furthermore, we can reconstruct higher-quality geometric results due to the multi-resolution 2D grid design of the sparse tri-plane encoding.

\begin{table*}[t]
      \renewcommand{\arraystretch}{1.2}
      \scriptsize
      \centering
            \setlength{\tabcolsep}{0.9mm}{
            \begin{tabular}{llccccccccc}
            \toprule
            & Metric & \ttfamily room0 & \ttfamily room1 & \ttfamily room2 &\ttfamily office0 & \ttfamily office1 & \ttfamily office2 & \ttfamily office3 & \ttfamily office4 &  Avg.\\
\midrule
\multirow{5}*{\rotatebox[origin=c]{90}{\scriptsize{ESLAM\cite{johari2023eslam}}}}
& Depth L1 [cm]$\downarrow$  
& 0.65 & \textbf{0.59} & \textbf{1.27} & \textbf{0.55} & 1.03 & \textbf{0.88} & \textbf{0.79} & \textbf{0.93} & \textbf{0.87}\\
& Acc. [cm]$\downarrow $ 
& 2.32 & \textbf{1.65} & 1.99 & 1.65 & 1.82 & \textbf{2.80} & \textbf{2.46} & 2.05 & \textbf{2.09}\\
& Comp. [cm]$\downarrow $ 
& 1.79 & \textbf{1.56} & 1.71 & 1.30 & 1.23 & 1.85 & 2.08 & 2.17 & 1.71\\
& Comp. Ratio [\%]$\uparrow $ 
& 97.12 & \textbf{96.83} & \textbf{96.39} & \textbf{98.55} & 97.85 & 95.68 & 95.08 & 93.70 & 96.40\\
& Param. [MB] $\downarrow $ 
& 6.79 & 5.28 & 14.20 & 11.92 & 10.93 & 6.44 & 8.75 & 9.45 & 9.22\\
\midrule
\multirow{5}*{{\rotatebox[origin=c]{90}{\scriptsize{Ours}}}} 
& Depth L1 [cm]$\downarrow$  
& \textbf{0.50} & 0.72 & 1.56 & 0.88 & \textbf{0.97} & 1.18 & 1.03 & 0.99 & 0.98\\
& Acc. [cm]$\downarrow $ 
& \textbf{2.31} & 2.17 & \textbf{1.81} & 1.63 & \textbf{1.49} & 2.83 & 2.61 & \textbf{2.03} & 2.11 \\
& Comp. [cm]$\downarrow $ 
& \textbf{1.73} & 1.63 & \textbf{1.62} & \textbf{1.29} & \textbf{1.16} & \textbf{1.83} & \textbf{2.04} & \textbf{1.97} & \textbf{1.66} \\
& Comp. Ratio [\%]$\uparrow $ 
& \textbf{97.75} & 96.02 & 95.47 & 98.40 & \textbf{98.03} & \textbf{96.10} & \textbf{95.98} & \textbf{95.96} & \textbf{96.71}\\
& Param. [MB] $\downarrow $ 
& \textbf{4.51} & \textbf{4.51} & \textbf{4.51} & \textbf{4.51} & \textbf{4.51} & \textbf{4.51} & \textbf{4.51} & \textbf{4.51} & \textbf{4.51}\\
\bottomrule
\end{tabular}
}
\caption{The qualitative results on Replica\cite{straub2019replica} demonstrate that our method, utilizing the proposed sparse tri-plane encoding, achieves a compact scene representation, resulting in high-quality geometric reconstruction. Compared to ESLAM\cite{johari2023eslam}, our method can complete more unseen regions.
}
\vspace{-4mm}
\label{replica_recon_eslam}
\end{table*}

\begin{table*}[t]
      \scriptsize
      \centering
            \setlength{\tabcolsep}{0.8mm}{
            \begin{tabular}{lccccccccc}
            \toprule
            Method & Depth L1 [cm]$\downarrow$ & Acc. [cm]$\downarrow$ &
            Comp. [cm]$\downarrow$ & 
            Comp. Ratio[\%] $\uparrow$ &
            Param. [MB]$\downarrow$\\
            \midrule
            ESLAM\cite{johari2023eslam}
            & 0.93 & 2.05 & 2.17 & 93.70 & 9.45\\
            $\text{ESLAM}^{\dagger}$\cite{johari2023eslam}
            & \textbf{0.84} & 2.07 & 2.02 & 94.08 & 77.47\\         
            Ours (Tri-plane) & 1.27 & 2.08 & 2.15 & 93.86 & 94.62\\
            Ours (30 iterations) & 0.99 & \textbf{2.03} & \textbf{1.97} & \textbf{95.96} & \textbf{4.51}\\
            \bottomrule
            \end{tabular}
            }
\caption{In the office4 scene of Replica\cite{straub2019replica}, we compare our method with SLAM implemented using tri-plane encoding, ESLAM\cite{johari2023eslam}, and high-resolution ESLAM ($\text{ESLAM}^{\dagger}$). Our method ensures reconstruction quality by employing sparse tri-plane encoding without adding to the parameter count.
}
\vspace{-8mm}
\label{triplane_recon_eslam}
\end{table*}

\begin{figure*}[!t]
    \vspace{-3mm}
    \centering
    \captionsetup[subfigure]{labelformat=empty}

      \begin{subfigure}{0.32\linewidth}
            \centering
            \captionsetup{justification=raggedright}
            \caption{\fontsize{9pt}{3pt}
             GT}
            \includegraphics[width=1\textwidth]{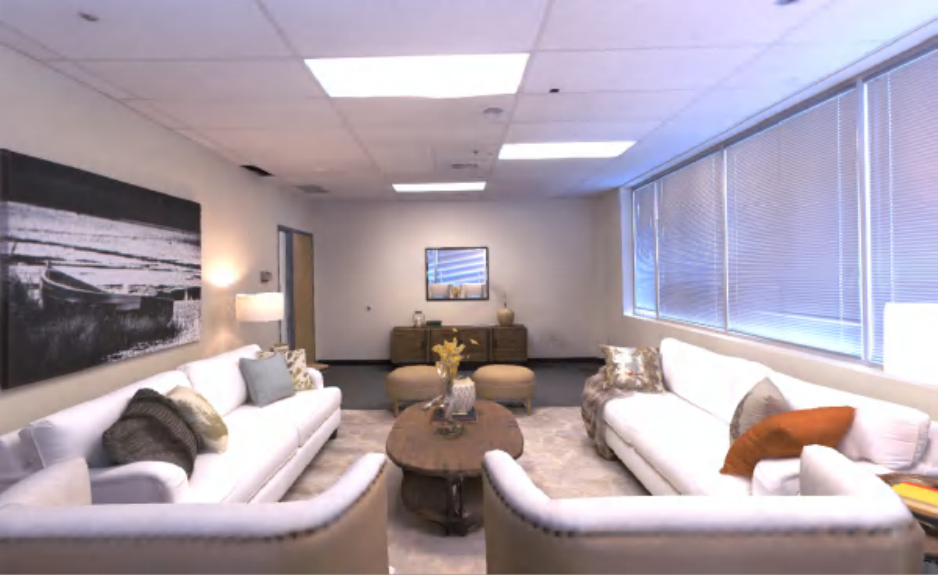}
            \label{room_0_coslam}
      \end{subfigure}
      \begin{subfigure}{0.32\linewidth}
            \centering
            \captionsetup{justification=raggedright}
            \caption{\fontsize{9pt}{1pt} NICE-SLAM\cite{zhu2022nice}}
            \includegraphics[width=1\textwidth]{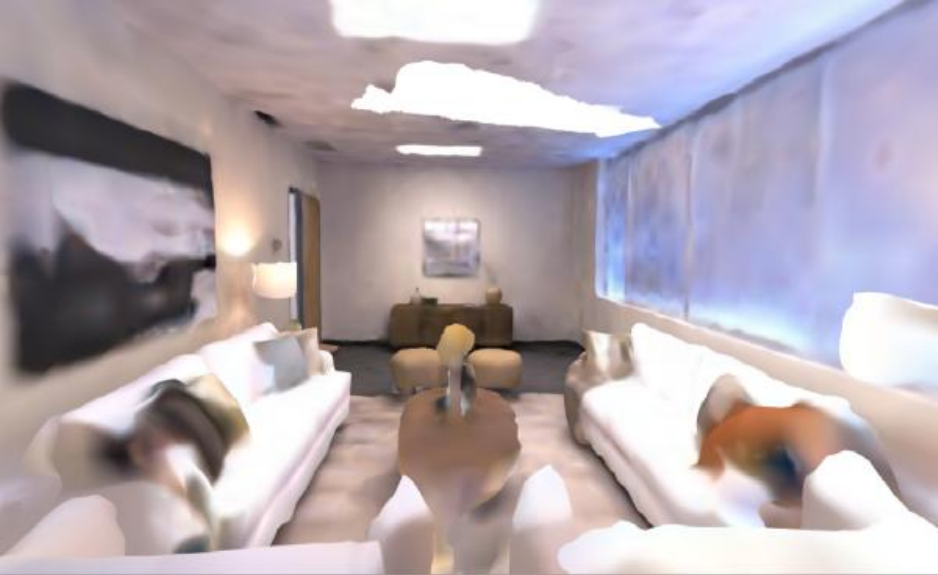}
            \label{room_2_coslam}
      \end{subfigure}
      \begin{subfigure}{0.32\linewidth}
            \centering
            \captionsetup{justification=raggedright}
            \caption{\fontsize{9pt}{1pt} Co-SLAM\cite{wang2023coslam}}
            \includegraphics[width=1\textwidth]{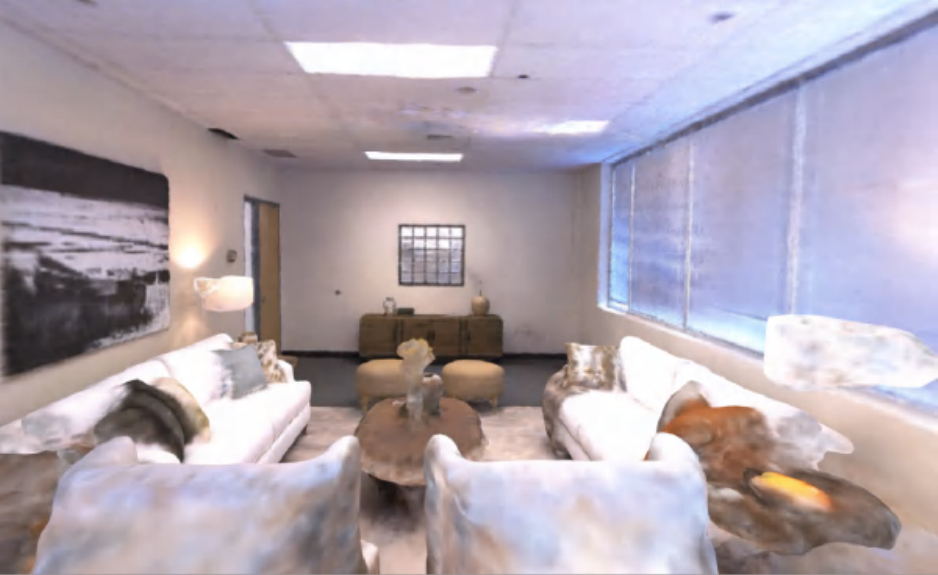}
            \label{office_3_coslam}
      \end{subfigure}
      
      \vspace{-2mm}
        \begin{subfigure}{0.32\linewidth}
              \centering
              \captionsetup{justification=raggedright}
              \caption{\fontsize{9pt}{4pt}
               ESLAM\cite{johari2023eslam}}
              \includegraphics[width=1\textwidth]{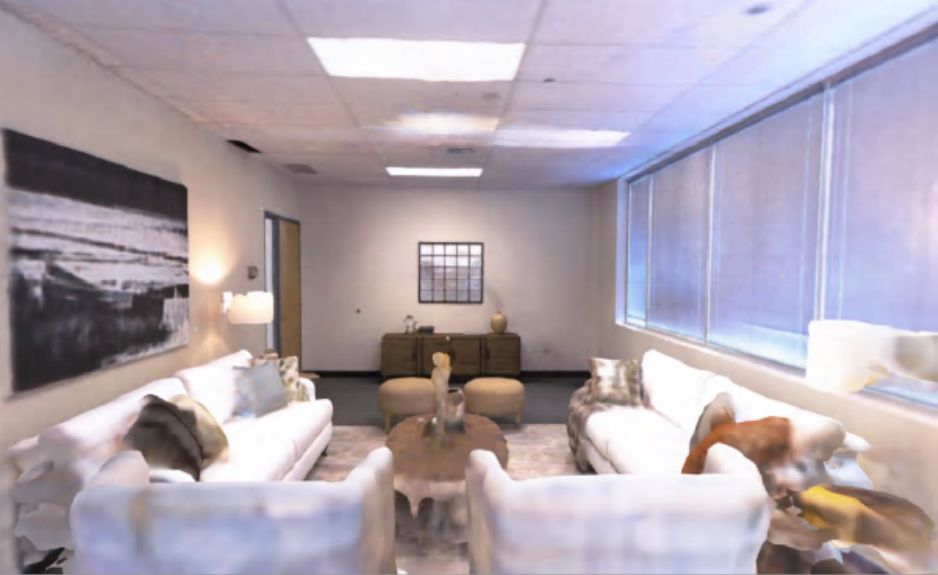}
              \label{room_0_coslam}
        \end{subfigure}
        \begin{subfigure}{0.32\linewidth}
              \centering
              \captionsetup{justification=raggedright}
              \caption{\fontsize{9pt}{1pt} Point-SLAM\cite{sandstrom2023pointslam}}
              \includegraphics[width=1\textwidth]{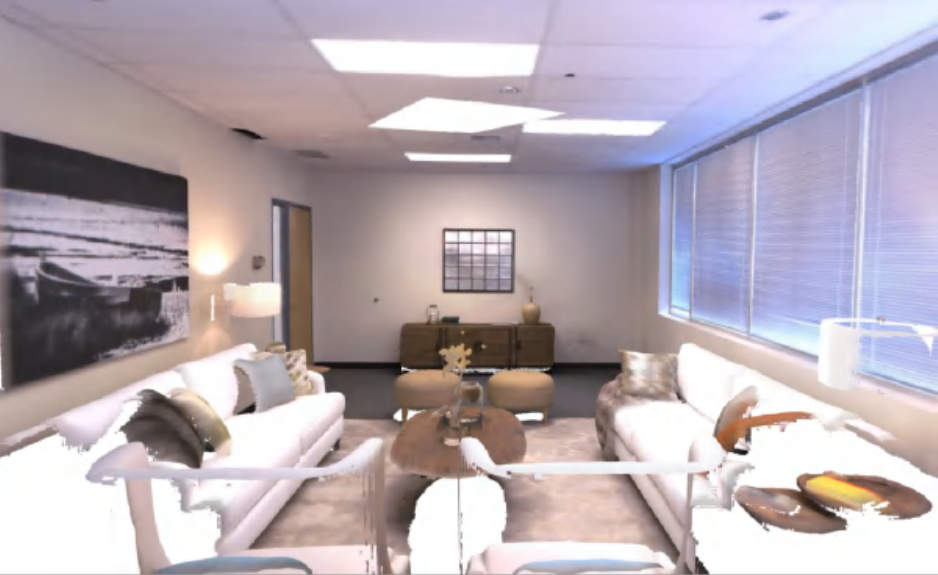}
              \label{room_2_coslam}
        \end{subfigure}
        \begin{subfigure}{0.32\linewidth}
              \centering
              \captionsetup{justification=raggedright}
              \caption{\fontsize{9pt}{4pt} Ours}
              \includegraphics[width=1\textwidth]{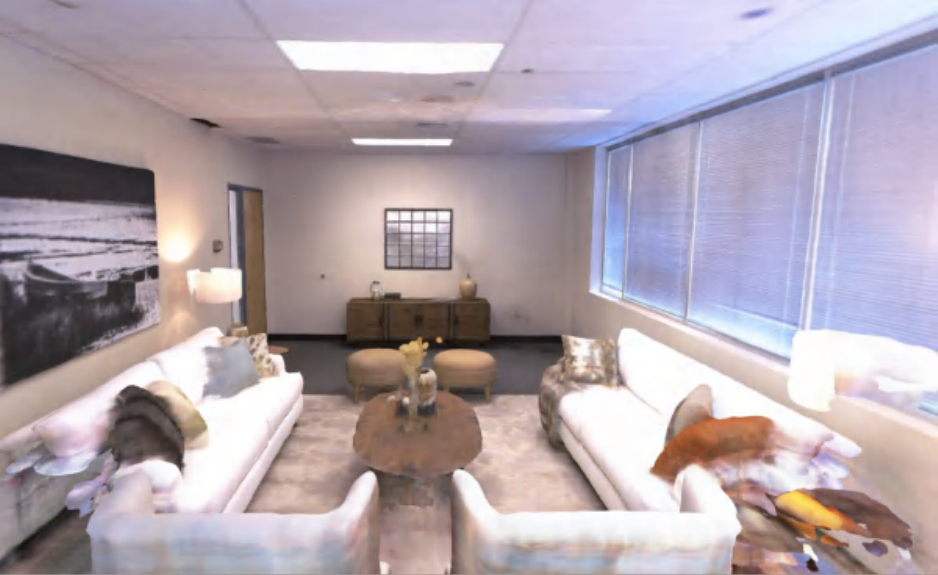}
              \label{office_3_coslam}
        \end{subfigure}



      \begin{subfigure}{0.32\linewidth}
            \centering
            \captionsetup{justification=raggedright}
            \caption{\fontsize{9pt}{3pt}
            GT}
            \includegraphics[width=1\textwidth]{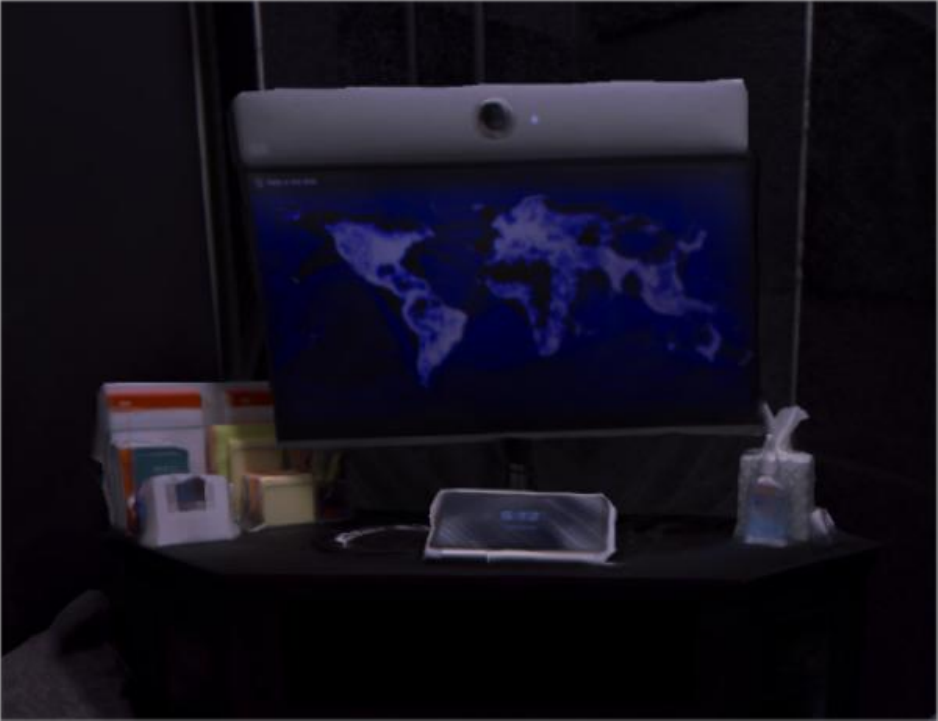}
            \label{room_0_coslam}
      \end{subfigure}
      \begin{subfigure}{0.32\linewidth}
            \centering
            \captionsetup{justification=raggedright}
            \caption{\fontsize{9pt}{1pt} NICE-SLAM\cite{zhu2022nice}}
            \includegraphics[width=1\textwidth]{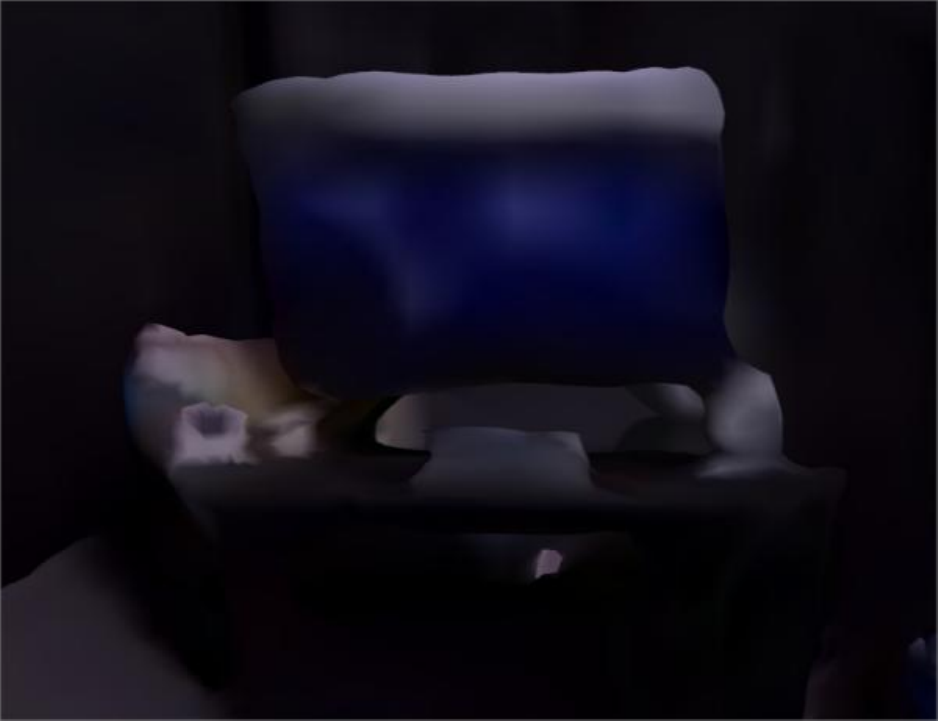}
            \label{room_2_coslam}
      \end{subfigure}
      \begin{subfigure}{0.32\linewidth}
            \centering
            \captionsetup{justification=raggedright}
            \caption{\fontsize{9pt}{1pt} Co-SLAM\cite{wang2023coslam}}
            \includegraphics[width=1\textwidth]{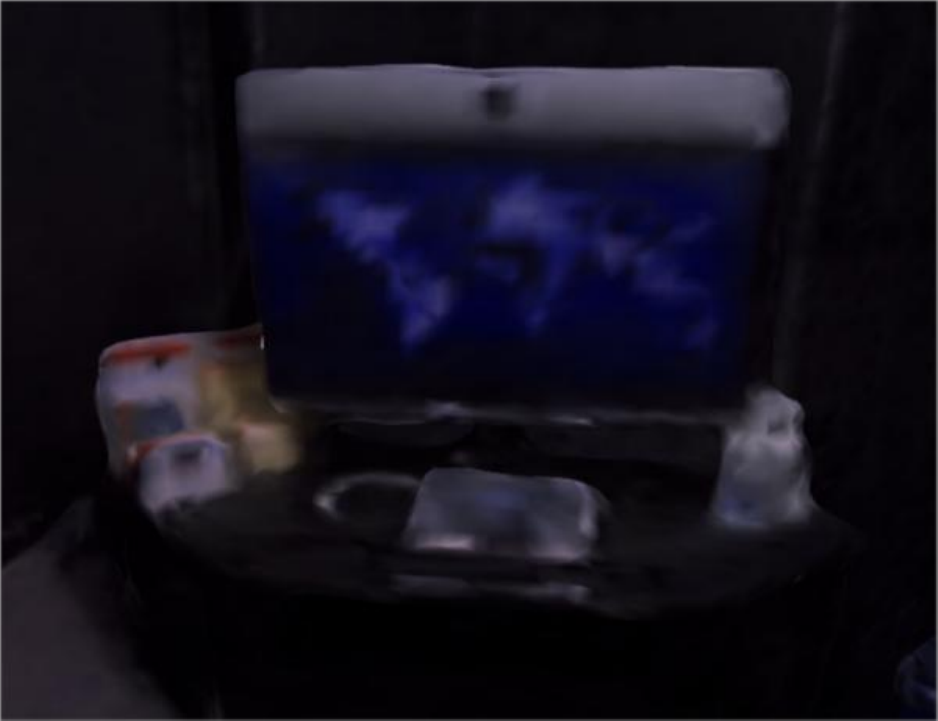}
            \label{office_3_coslam}
      \end{subfigure}
      
      \vspace{-2mm}
        \begin{subfigure}{0.32\linewidth}
              \centering
              \captionsetup{justification=raggedright}
              \caption{\fontsize{9pt}{4pt}
              ESLAM\cite{johari2023eslam}}
              \includegraphics[width=1\textwidth]{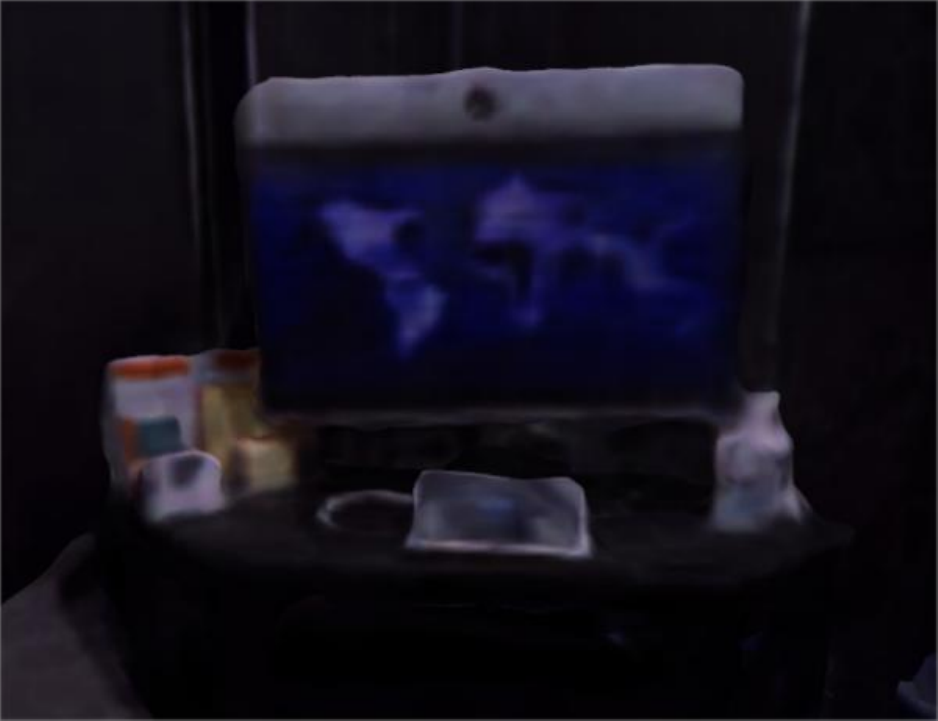}
              \label{room_0_coslam}
        \end{subfigure}
        \begin{subfigure}{0.32\linewidth}
              \centering
              \captionsetup{justification=raggedright}
              \caption{\fontsize{9pt}{1pt}Point-SLAM\cite{sandstrom2023pointslam}}
              \includegraphics[width=1\textwidth]{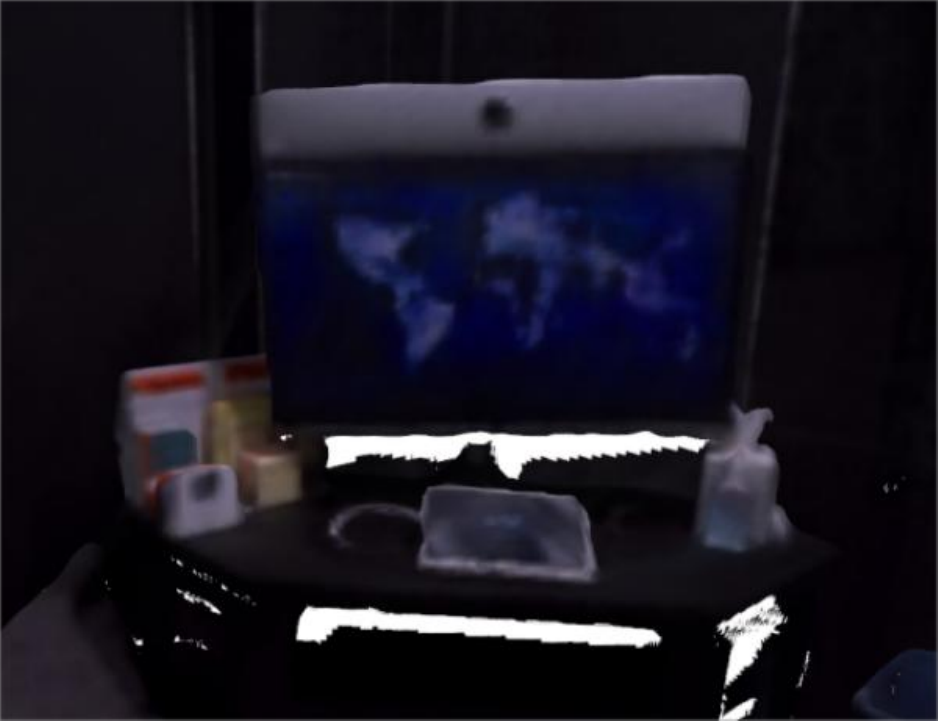}
              \label{room_2_coslam}
        \end{subfigure}
        \begin{subfigure}{0.32\linewidth}
              \centering
              \captionsetup{justification=raggedright}
              \caption{\fontsize{9pt}{4pt}Ours}
              \includegraphics[width=1\textwidth]{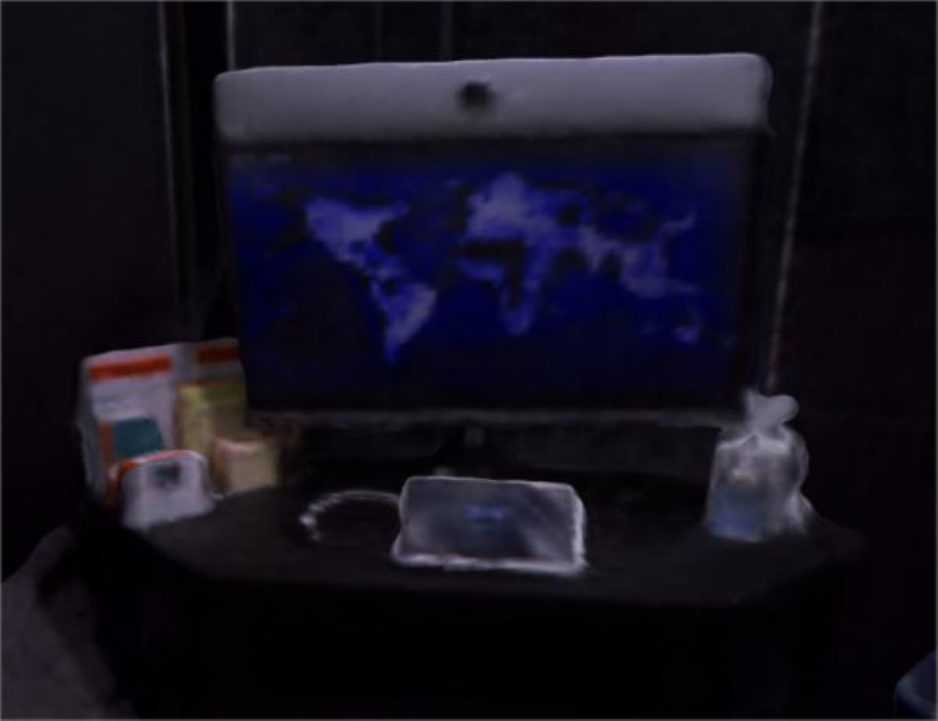}
              \label{office_3_coslam}
        \end{subfigure}

      \vspace{-6mm}
      \caption{The qualitative results on Replica\cite{straub2019replica} indicate that our method can reconstruct more fine-level appearances and geometry, producing more competitive results in appearance reconstruction.}
       \label{fig:replica_1}
       \vspace{-2mm}
  \end{figure*}

\begin{figure*}[!t]
      \centering
      \captionsetup[subfigure]{labelformat=empty}

      \begin{subfigure}{0.32\linewidth}
            \centering
            \captionsetup{justification=raggedright}
            \caption{\fontsize{9pt}{3pt}
            GT}
            \includegraphics[width=1\textwidth]{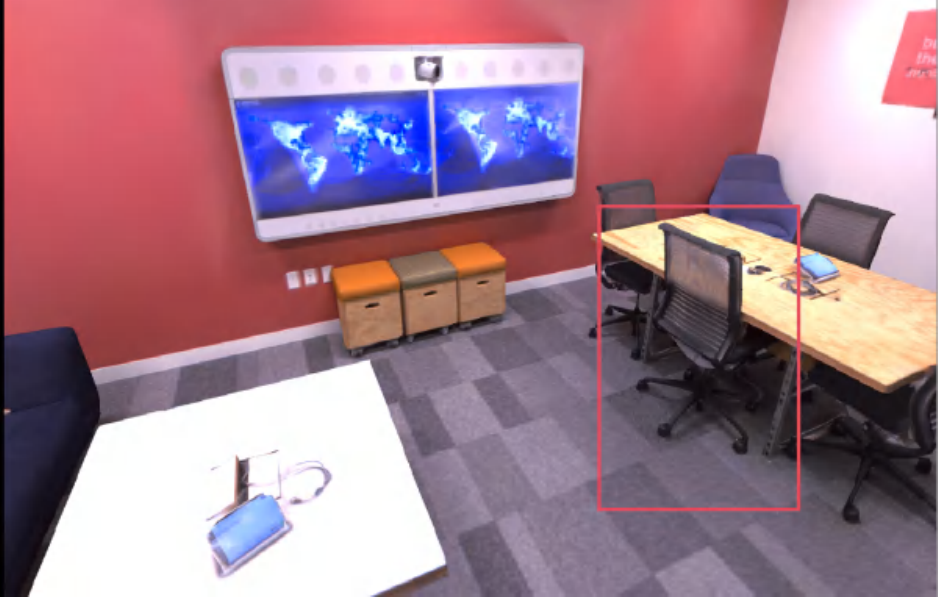}
            \label{room_0_coslam}
      \end{subfigure}
      \begin{subfigure}{0.32\linewidth}
            \centering
            \captionsetup{justification=raggedright}
            \caption{\fontsize{9pt}{1pt} NICE-SLAM\cite{zhu2022nice}}
            \includegraphics[width=1\textwidth]{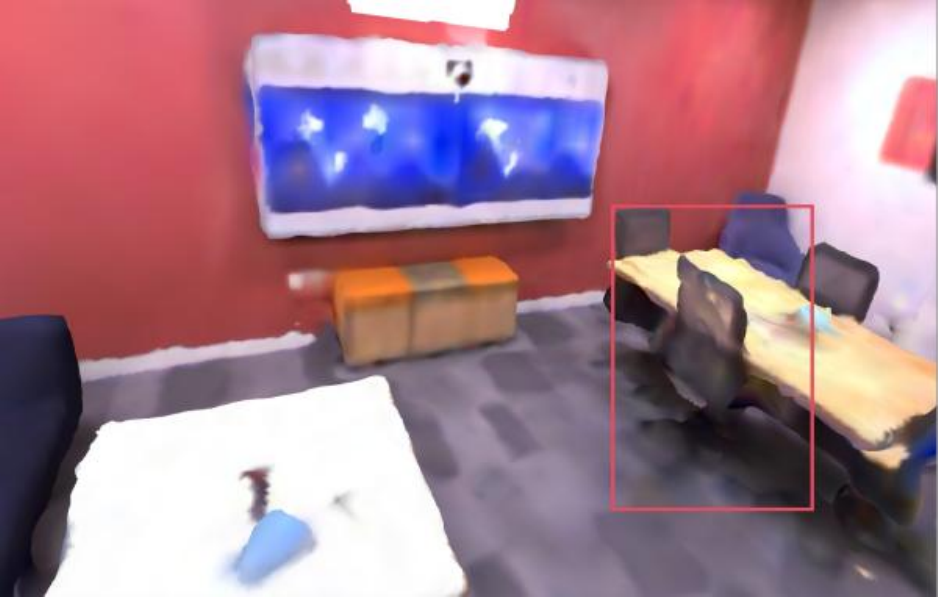}
            \label{room_2_coslam}
      \end{subfigure}
      \begin{subfigure}{0.32\linewidth}
            \centering
            \captionsetup{justification=raggedright}
            \caption{\fontsize{9pt}{1pt} Co-SLAM\cite{wang2023coslam}}
            \includegraphics[width=1\textwidth]{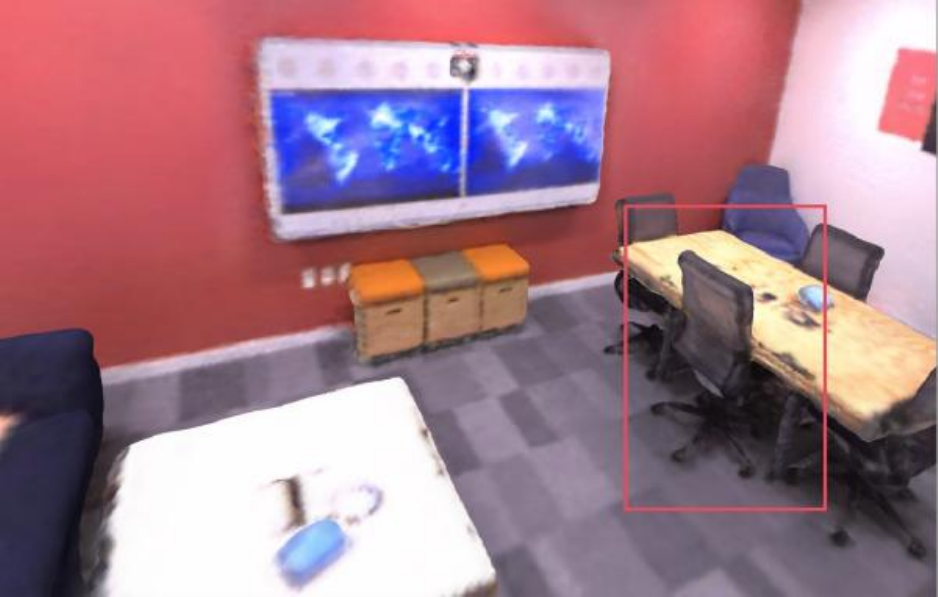}
            \label{office_3_coslam}
      \end{subfigure}
      
      \vspace{-2mm}
        \begin{subfigure}{0.32\linewidth}
              \centering
              \captionsetup{justification=raggedright}
              \caption{\fontsize{9pt}{4pt}
              ESLAM\cite{johari2023eslam}}
              \includegraphics[width=1\textwidth]{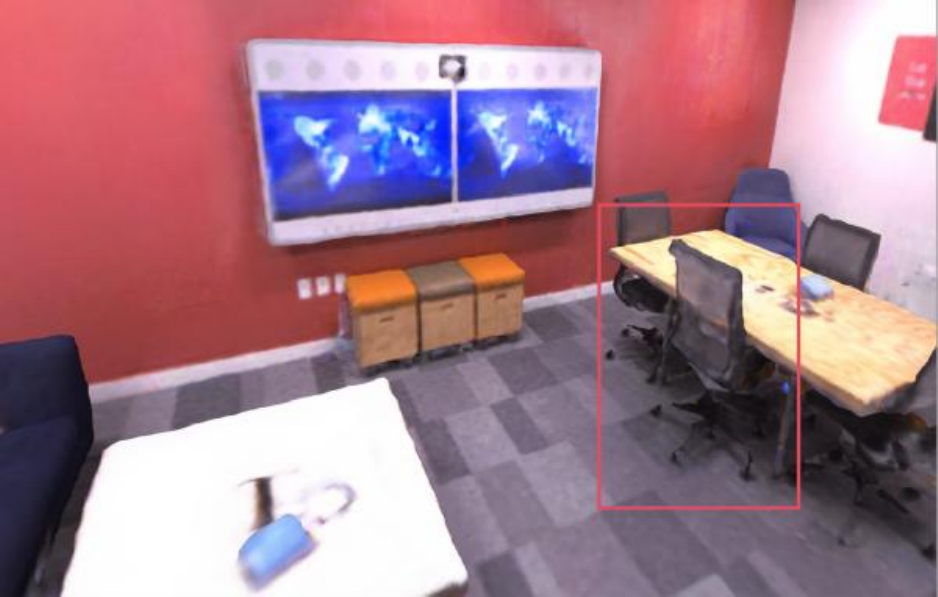}
              \label{room_0_coslam}
        \end{subfigure}
        \begin{subfigure}{0.32\linewidth}
              \centering
              \captionsetup{justification=raggedright}
              \caption{\fontsize{9pt}{1pt}Point-SLAM\cite{sandstrom2023pointslam}}
              \includegraphics[width=1\textwidth]{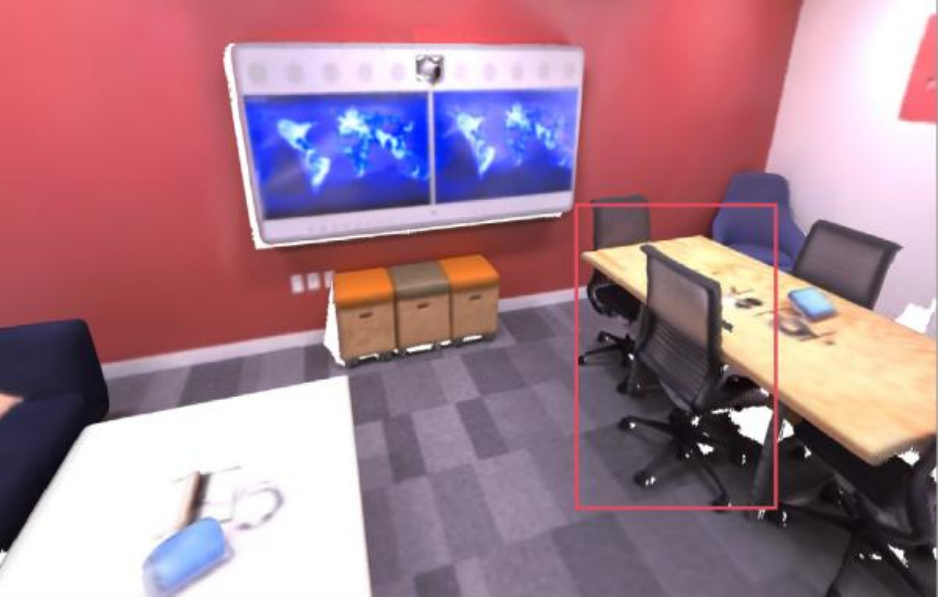}
              \label{room_2_coslam}
        \end{subfigure}
        \begin{subfigure}{0.32\linewidth}
              \centering
              \captionsetup{justification=raggedright}
              \caption{\fontsize{9pt}{4pt}Ours}
              \includegraphics[width=1\textwidth]{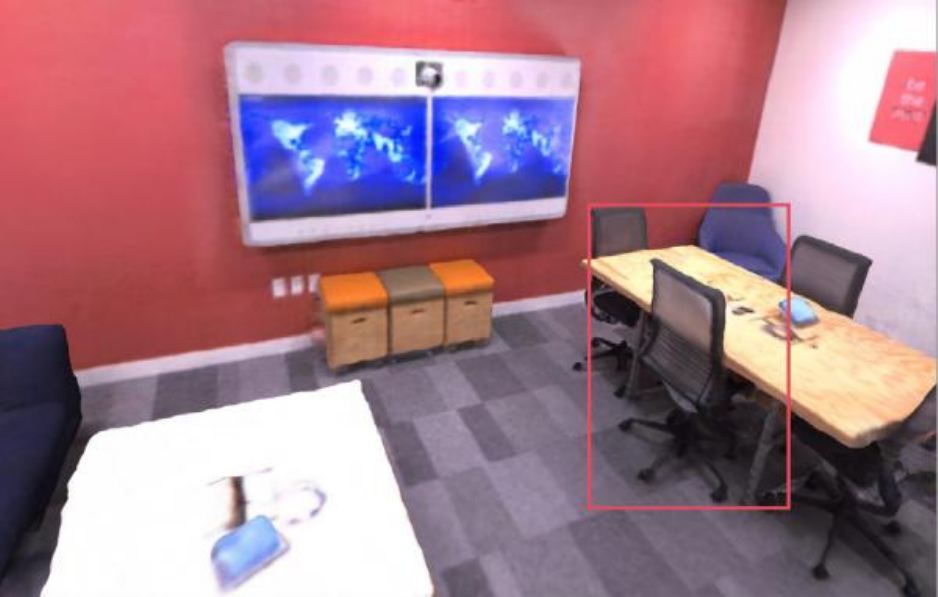}
              \label{office_3_coslam}
        \end{subfigure}



      \begin{subfigure}{0.32\linewidth}
            \centering
            \captionsetup{justification=raggedright}
            \caption{\fontsize{9pt}{3pt}
             GT}
            \includegraphics[width=1\textwidth]{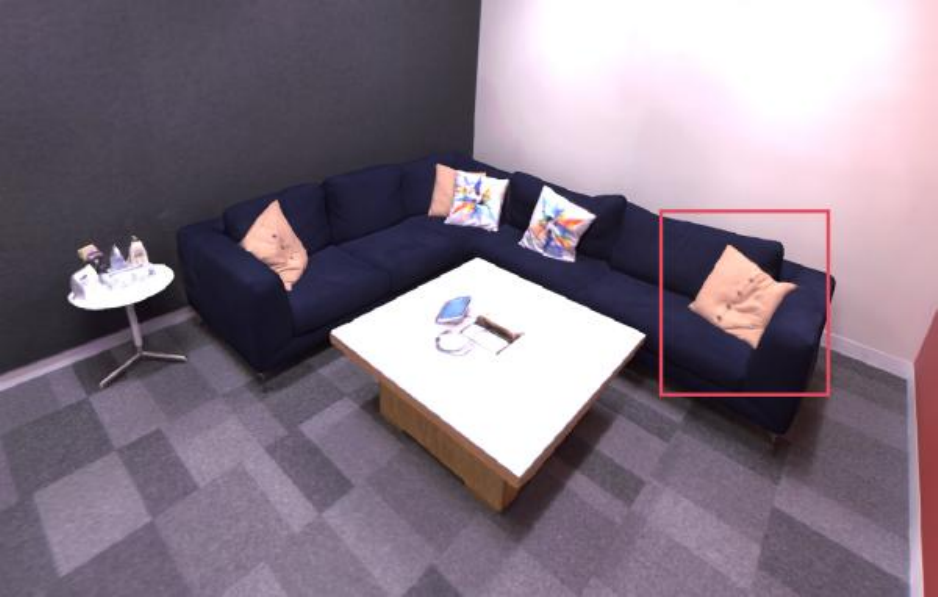}
            \label{room_0_coslam}
      \end{subfigure}
      \begin{subfigure}{0.32\linewidth}
            \centering
            \captionsetup{justification=raggedright}
            \caption{\fontsize{9pt}{1pt} NICE-SLAM\cite{zhu2022nice}}
            \includegraphics[width=1\textwidth]{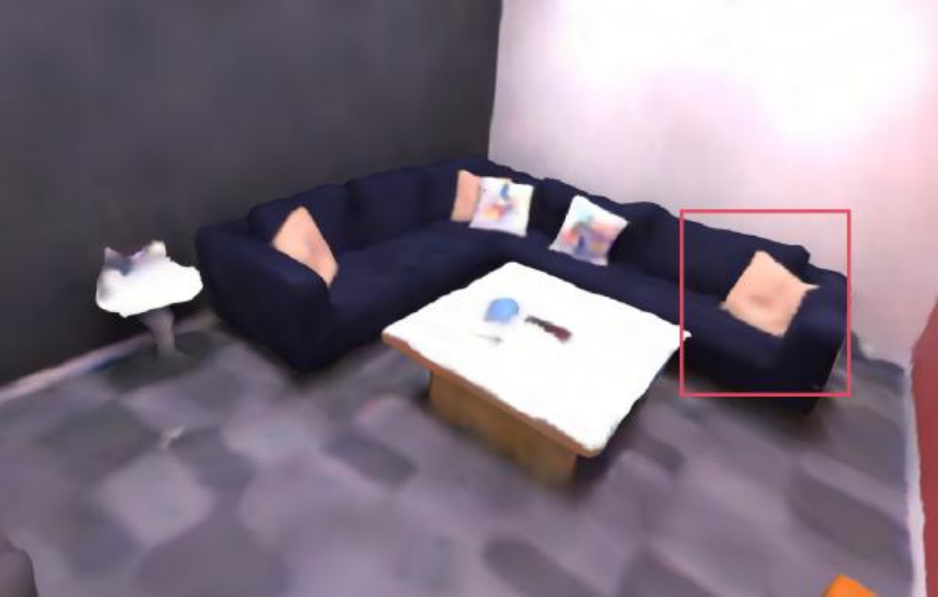}
            \label{room_2_coslam}
      \end{subfigure}
      \begin{subfigure}{0.32\linewidth}
            \centering
            \captionsetup{justification=raggedright}
            \caption{\fontsize{9pt}{1pt} Co-SLAM\cite{wang2023coslam}}
            \includegraphics[width=1\textwidth]{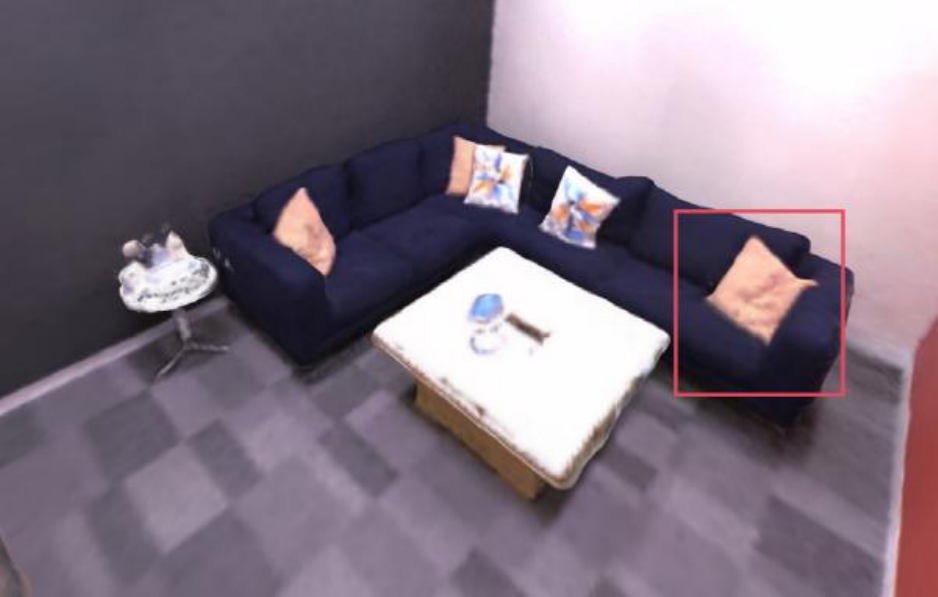}
            \label{office_3_coslam}
      \end{subfigure}
      
      \vspace{-2mm}
        \begin{subfigure}{0.32\linewidth}
              \centering
              \captionsetup{justification=raggedright}
              \caption{\fontsize{9pt}{4pt}
               ESLAM\cite{johari2023eslam}}
              \includegraphics[width=1\textwidth]{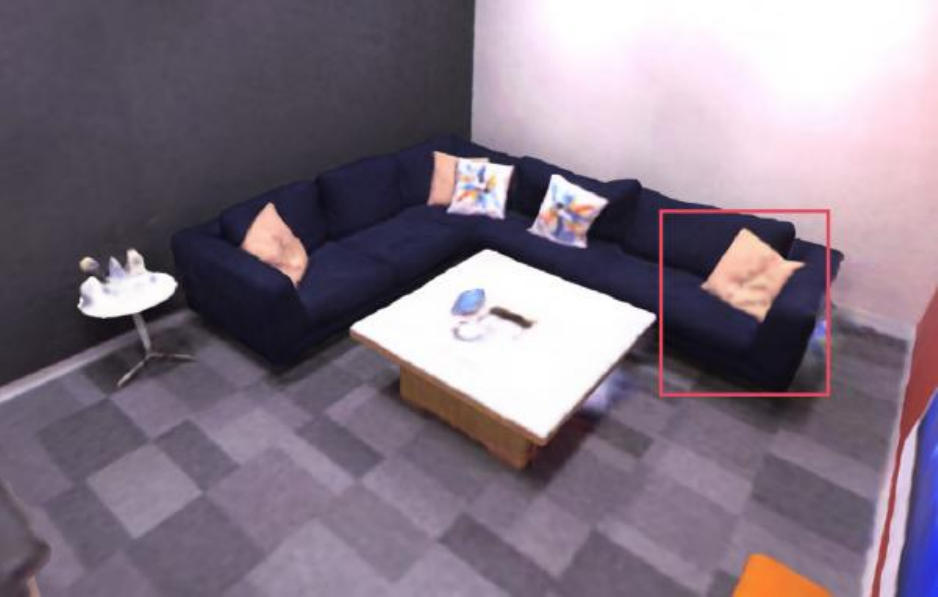}
              \label{room_0_coslam}
        \end{subfigure}
        \begin{subfigure}{0.32\linewidth}
              \centering
              \captionsetup{justification=raggedright}
              \caption{\fontsize{9pt}{1pt} Point-SLAM\cite{sandstrom2023pointslam}}
              \includegraphics[width=1\textwidth]{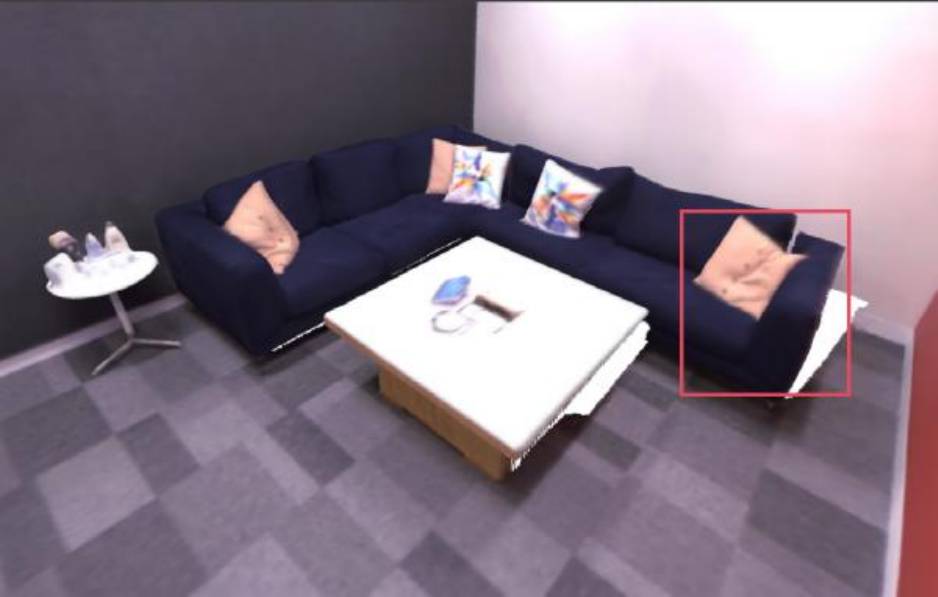}
              \label{room_2_coslam}
        \end{subfigure}
        \begin{subfigure}{0.32\linewidth}
              \centering
              \captionsetup{justification=raggedright}
              \caption{\fontsize{9pt}{4pt} Ours}
              \includegraphics[width=1\textwidth]{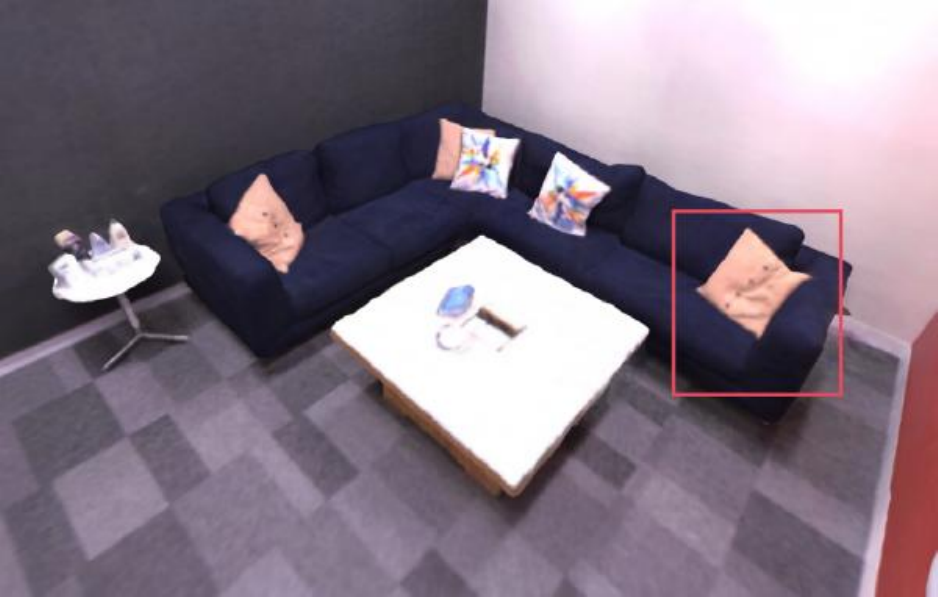}
              \label{office_3_coslam}
        \end{subfigure}

      \vspace{-6mm}
      \caption{The qualitative results on Replica\cite{straub2019replica} demonstrate that our method can reconstruct tiny patterns on sofa cushions and achieve more accurate geometry reconstruction for small items like tables.}
       \label{fig:replica_2}
       \vspace{-2mm}
  \end{figure*}

\begin{figure*}[!t]
      \centering
      \captionsetup[subfigure]{labelformat=empty}

      \begin{subfigure}{0.32\linewidth}
            \centering
            \captionsetup{justification=raggedright}
            \caption{\fontsize{9pt}{3pt}
             GT}
            \includegraphics[width=1\textwidth]{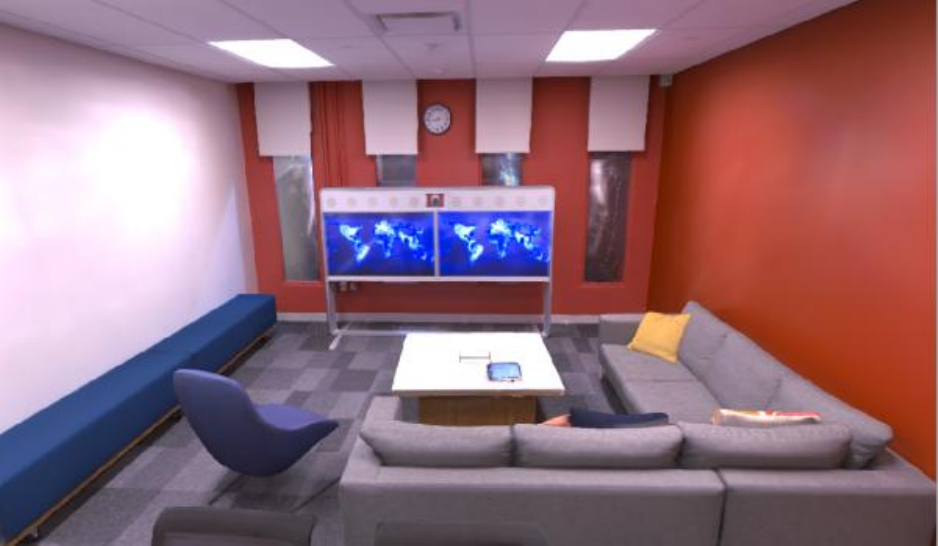}
            \label{room_0_coslam}
      \end{subfigure}
      \begin{subfigure}{0.32\linewidth}
            \centering
            \captionsetup{justification=raggedright}
            \caption{\fontsize{9pt}{1pt} NICE-SLAM\cite{zhu2022nice}}
            \includegraphics[width=1\textwidth]{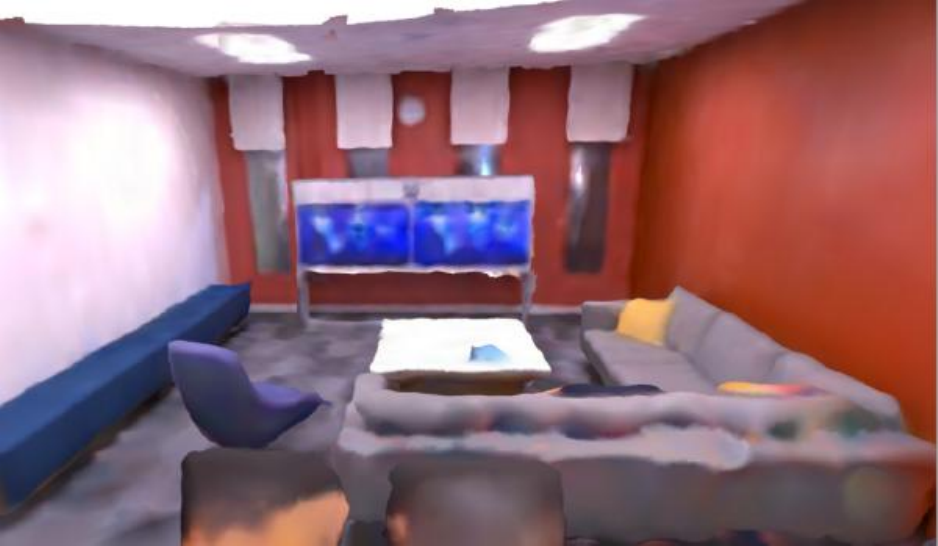}
            \label{room_2_coslam}
      \end{subfigure}
      \begin{subfigure}{0.32\linewidth}
            \centering
            \captionsetup{justification=raggedright}
            \caption{\fontsize{9pt}{1pt} Co-SLAM\cite{wang2023coslam}}
            \includegraphics[width=1\textwidth]{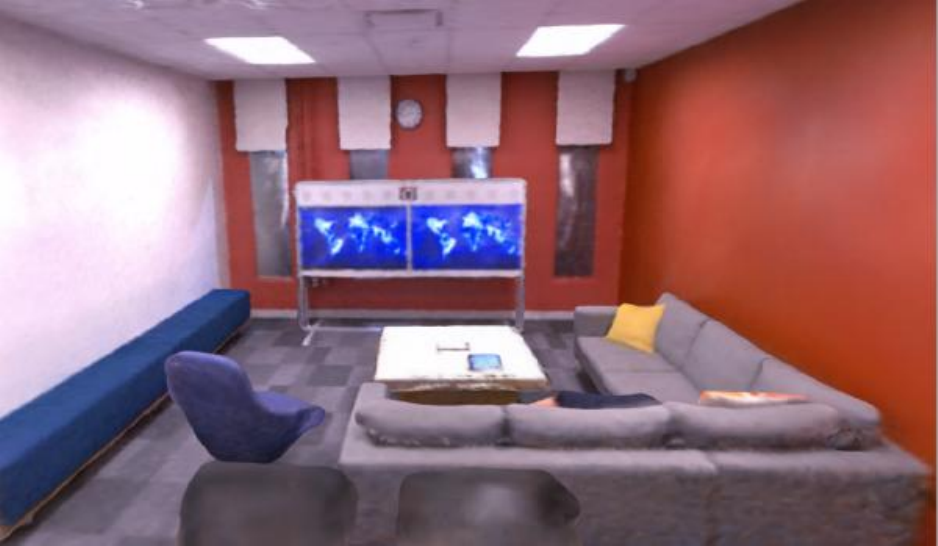}
            \label{office_3_coslam}
      \end{subfigure}
      
      \vspace{-2mm}
        \begin{subfigure}{0.32\linewidth}
              \centering
              \captionsetup{justification=raggedright}
              \caption{\fontsize{9pt}{4pt}
               ESLAM\cite{johari2023eslam}}
              \includegraphics[width=1\textwidth]{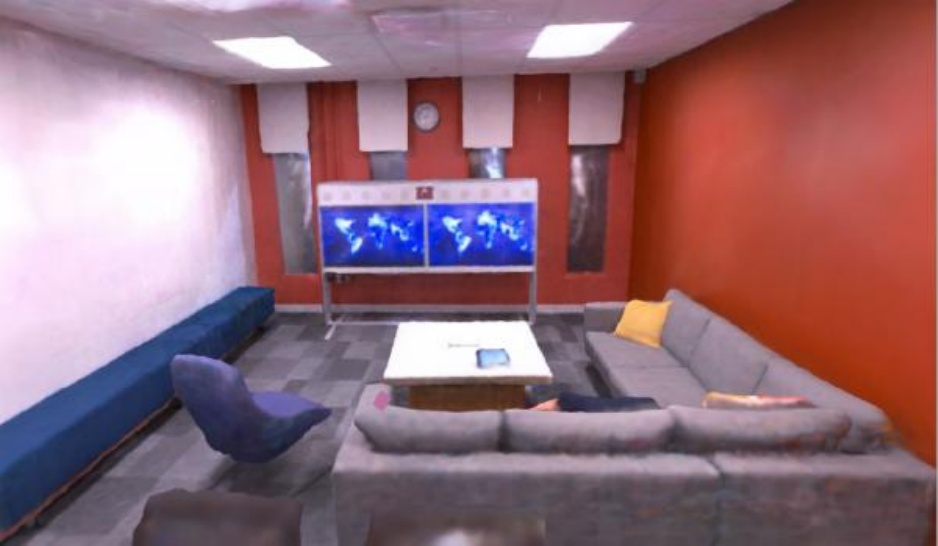}
              \label{room_0_coslam}
        \end{subfigure}
        \begin{subfigure}{0.32\linewidth}
              \centering
              \captionsetup{justification=raggedright}
              \caption{\fontsize{9pt}{1pt} Point-SLAM\cite{sandstrom2023pointslam}}
              \includegraphics[width=1\textwidth]{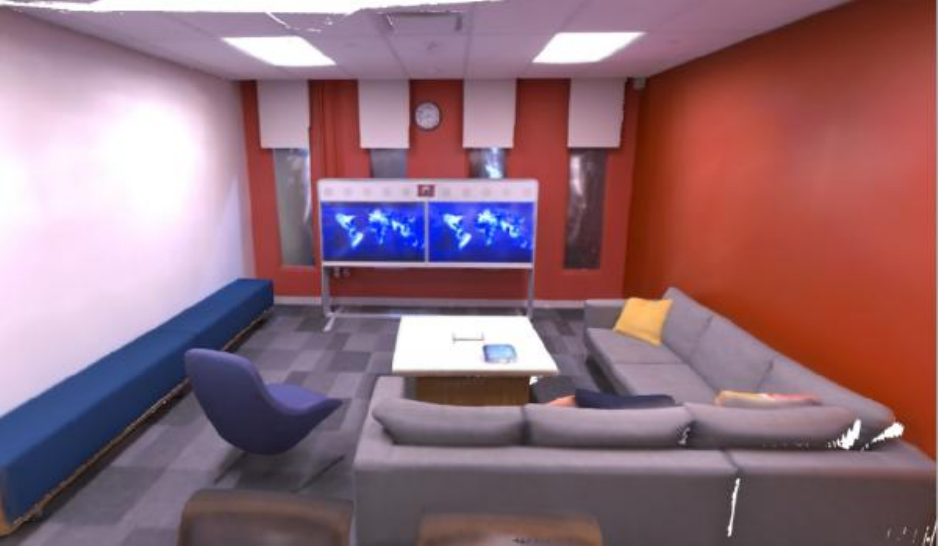}
              \label{room_2_coslam}
        \end{subfigure}
        \begin{subfigure}{0.32\linewidth}
              \centering
              \captionsetup{justification=raggedright}
              \caption{\fontsize{9pt}{4pt} Ours}
              \includegraphics[width=1\textwidth]{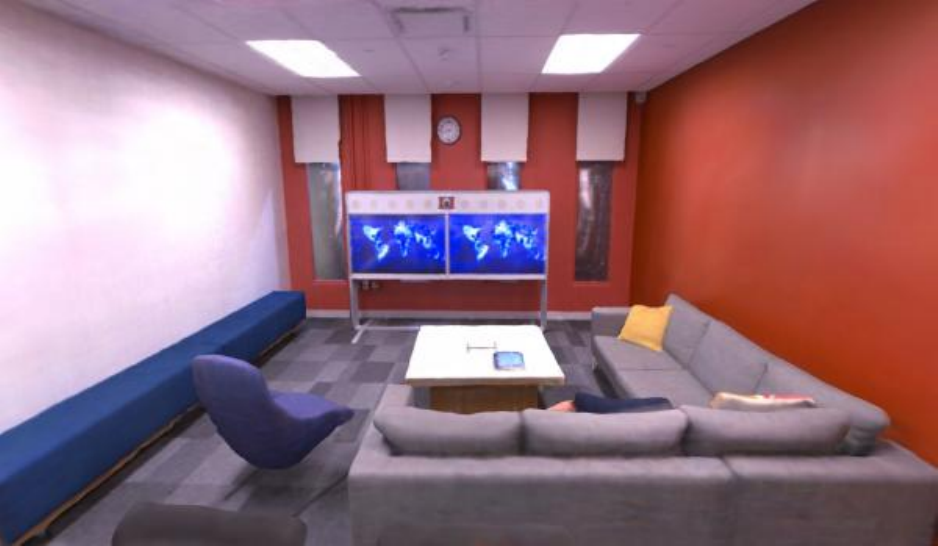}
              \label{office_3_coslam}
        \end{subfigure}



      \begin{subfigure}{0.32\linewidth}
            \centering
            \captionsetup{justification=raggedright}
            \caption{\fontsize{9pt}{3pt}
             GT}
            \includegraphics[width=1\textwidth]{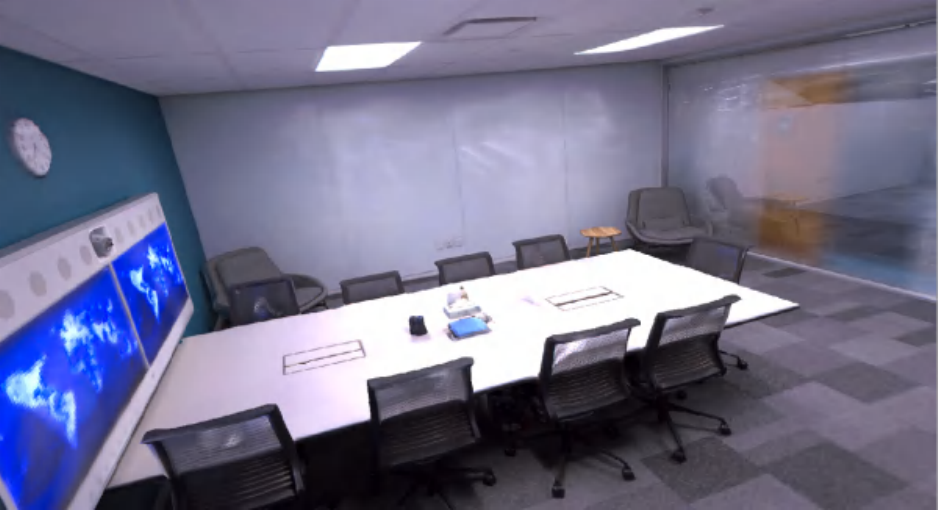}
            \label{room_0_coslam}
      \end{subfigure}
      \begin{subfigure}{0.32\linewidth}
            \centering
            \captionsetup{justification=raggedright}
            \caption{\fontsize{9pt}{1pt} NICE-SLAM\cite{zhu2022nice}}
            \includegraphics[width=1\textwidth]{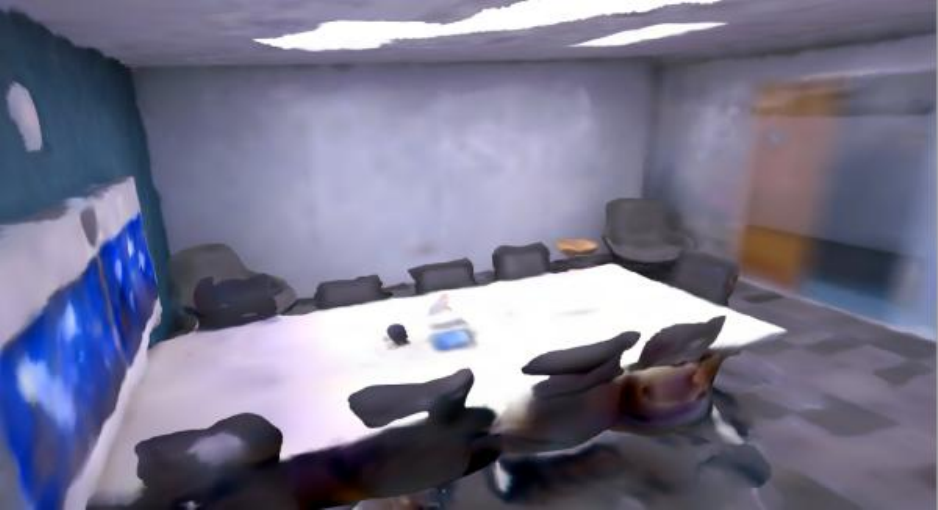}
            \label{room_2_coslam}
      \end{subfigure}
      \begin{subfigure}{0.32\linewidth}
            \centering
            \captionsetup{justification=raggedright}
            \caption{\fontsize{9pt}{1pt} Co-SLAM\cite{wang2023coslam}}
            \includegraphics[width=1\textwidth]{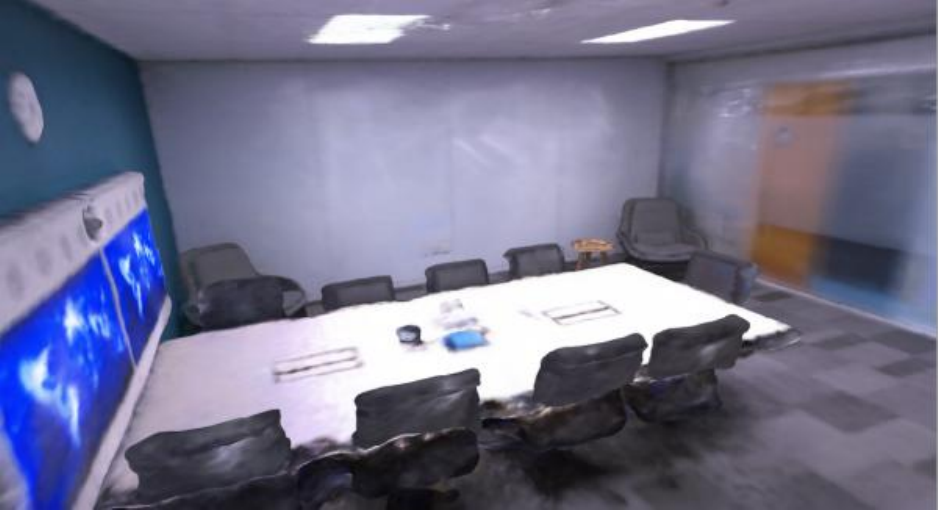}
            \label{office_3_coslam}
      \end{subfigure}
      
      \vspace{-2mm}
        \begin{subfigure}{0.32\linewidth}
              \centering
              \captionsetup{justification=raggedright}
              \caption{\fontsize{9pt}{4pt}
               ESLAM\cite{johari2023eslam}}
              \includegraphics[width=1\textwidth]{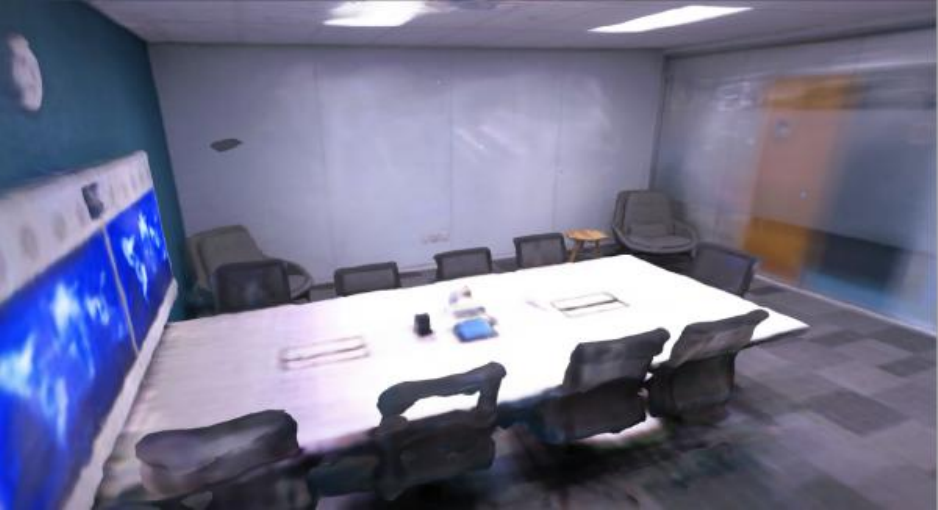}
              \label{room_0_coslam}
        \end{subfigure}
        \begin{subfigure}{0.32\linewidth}
              \centering
              \captionsetup{justification=raggedright}
              \caption{\fontsize{9pt}{1pt} Point-SLAM\cite{sandstrom2023pointslam}}
              \includegraphics[width=1\textwidth]{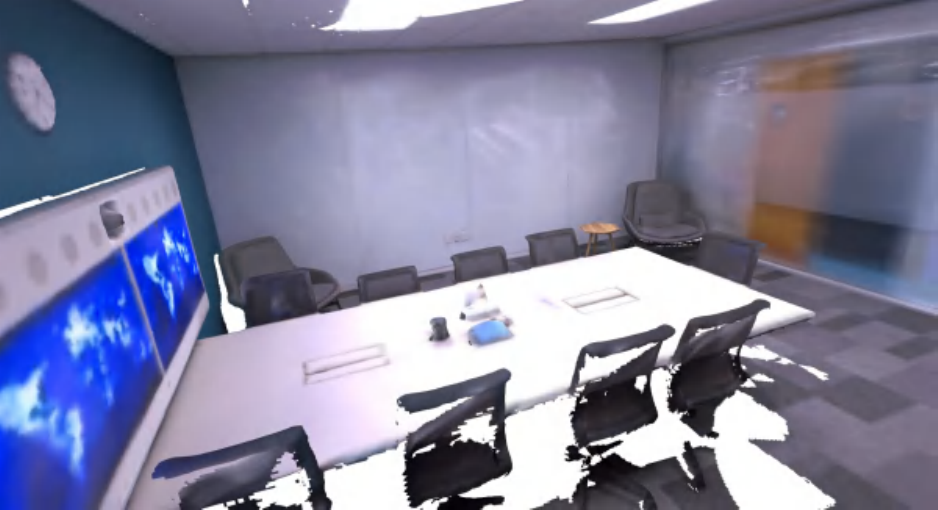}
              \label{room_2_coslam}
        \end{subfigure}
        \begin{subfigure}{0.32\linewidth}
              \centering
              \captionsetup{justification=raggedright}
              \caption{\fontsize{9pt}{4pt} Ours}
              \includegraphics[width=1\textwidth]{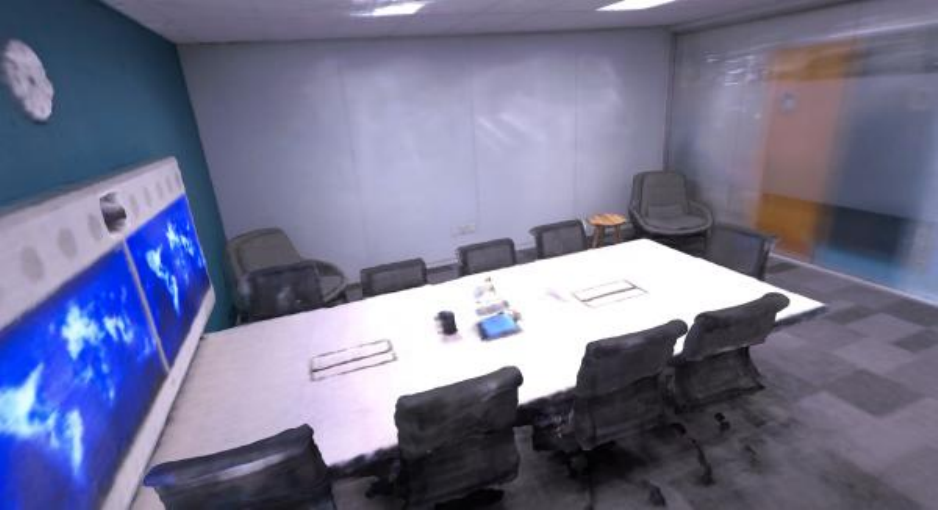}
              \label{office_3_coslam}
        \end{subfigure}

      \vspace{-6mm}
      \caption{The qualitative results on Replica\cite{straub2019replica} indicate that our method accurately completes a significant amount of unseen areas and can perform fine-grained reconstruction for tabletop-level scenes, including capturing small details on windows.}
       \label{fig:replica_3}
       \vspace{-2mm}
  \end{figure*}

\begin{figure*}[!t]
      \centering
      \captionsetup[subfigure]{labelformat=empty}
 
      \rotatebox[origin=lb]{90}{\scriptsize{NICE-SLAM\cite{zhu2022nice}}}
      \begin{subfigure}{0.31\linewidth}
            \centering
            \captionsetup{justification=raggedright}
            \includegraphics[width=1\textwidth]{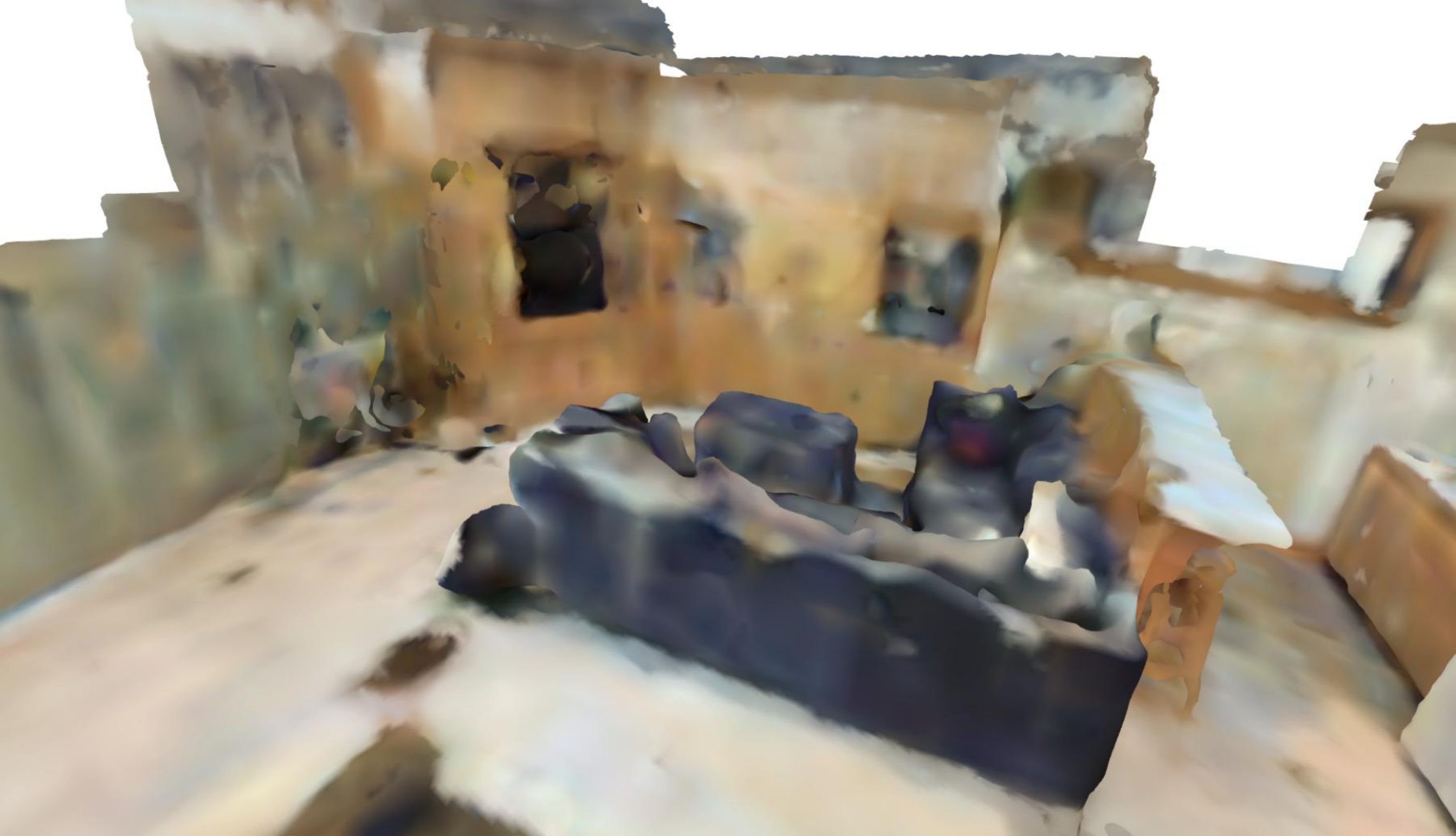}
      \end{subfigure}
      \begin{subfigure}{0.31\linewidth}
            \centering
            \captionsetup{justification=raggedright}
            \includegraphics[width=1\textwidth]{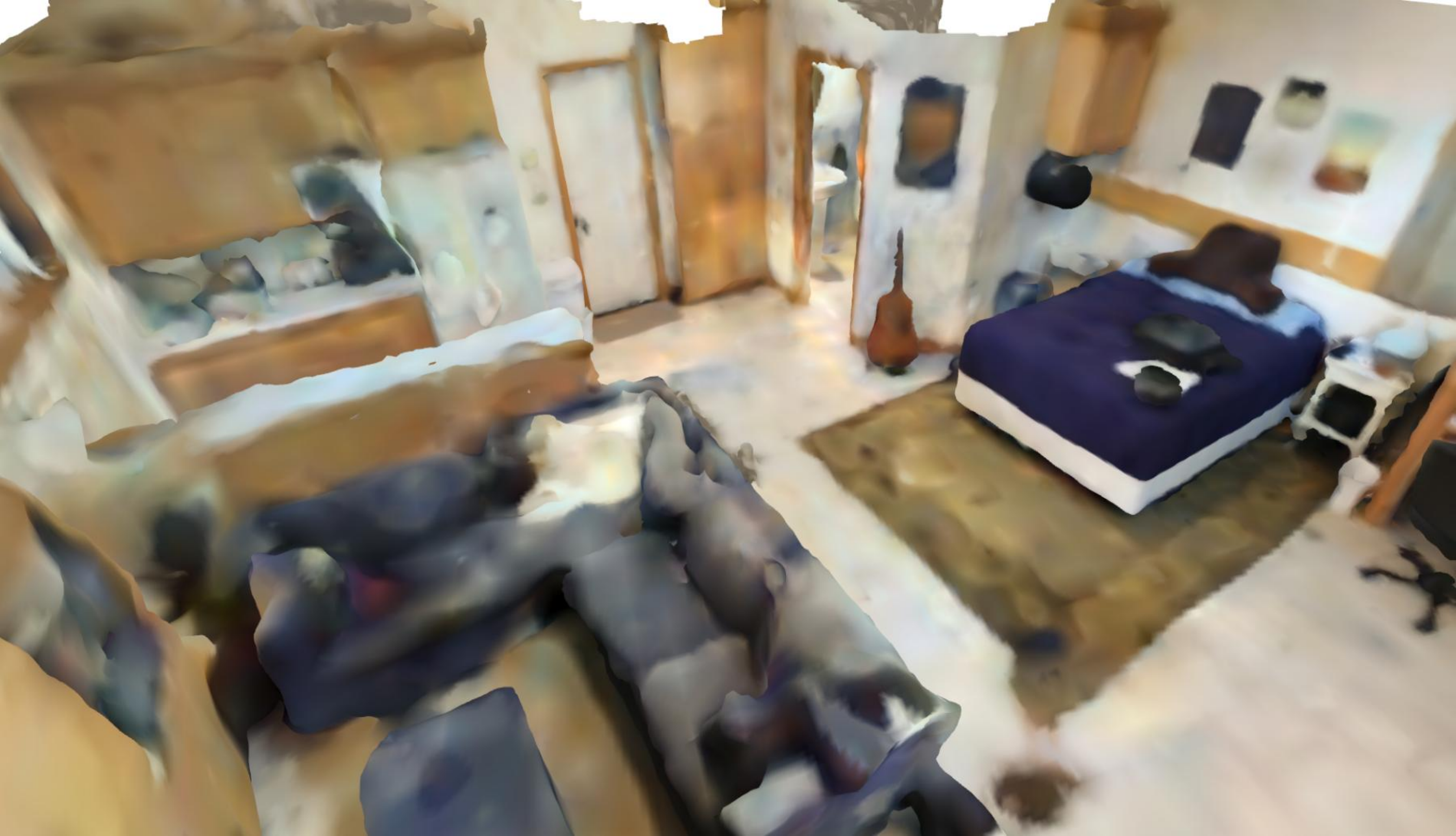}
      \end{subfigure}
      \begin{subfigure}{0.31\linewidth}
            \captionsetup{justification=raggedright}
            \includegraphics[width=1\textwidth]{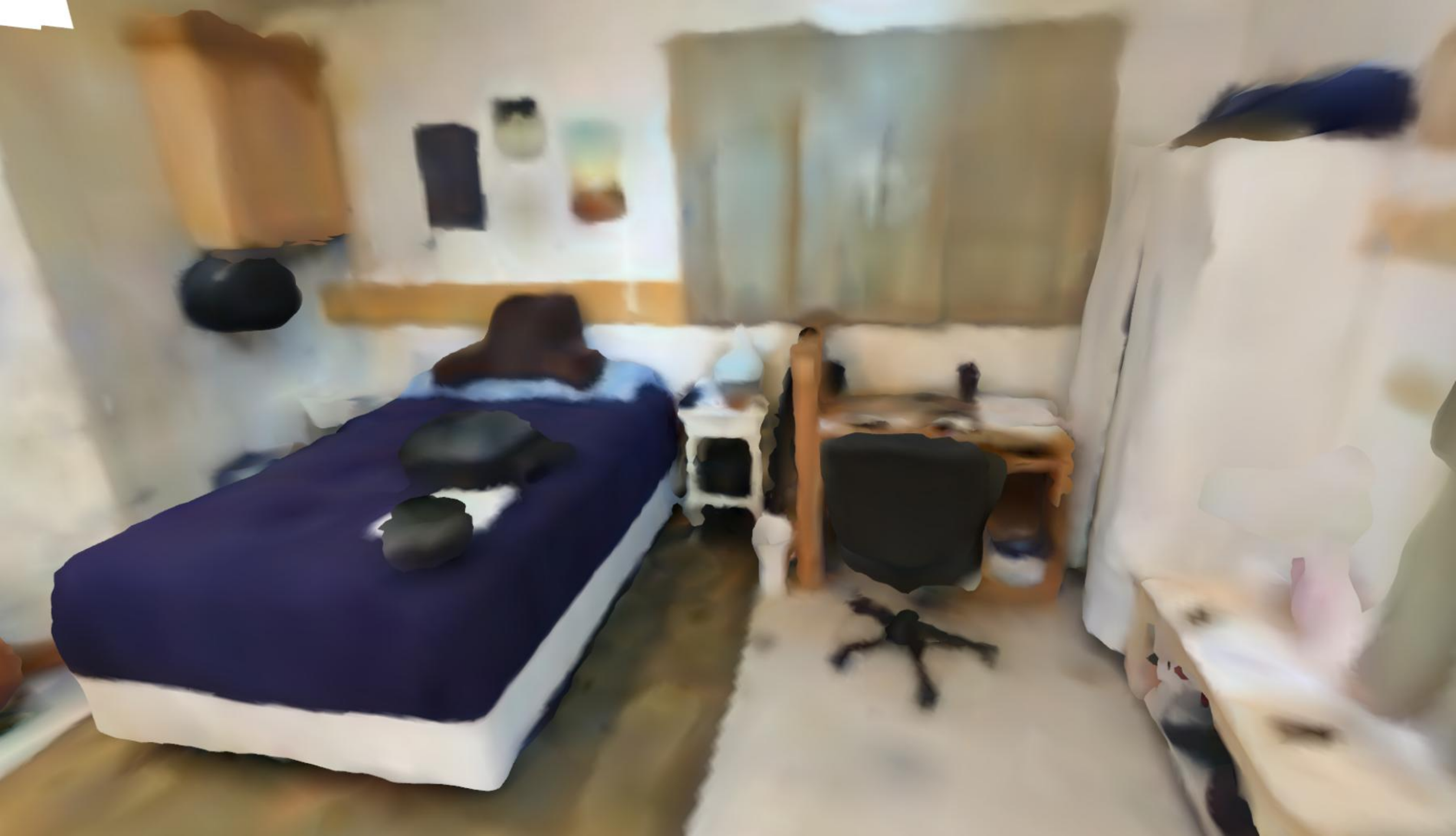}
      \end{subfigure}

      \rotatebox[origin=lb]{90}{\hspace{2mm} \scriptsize{ESLAM\cite{johari2023eslam}}}
      \begin{subfigure}{0.31\linewidth}
            \includegraphics[width=1\textwidth]{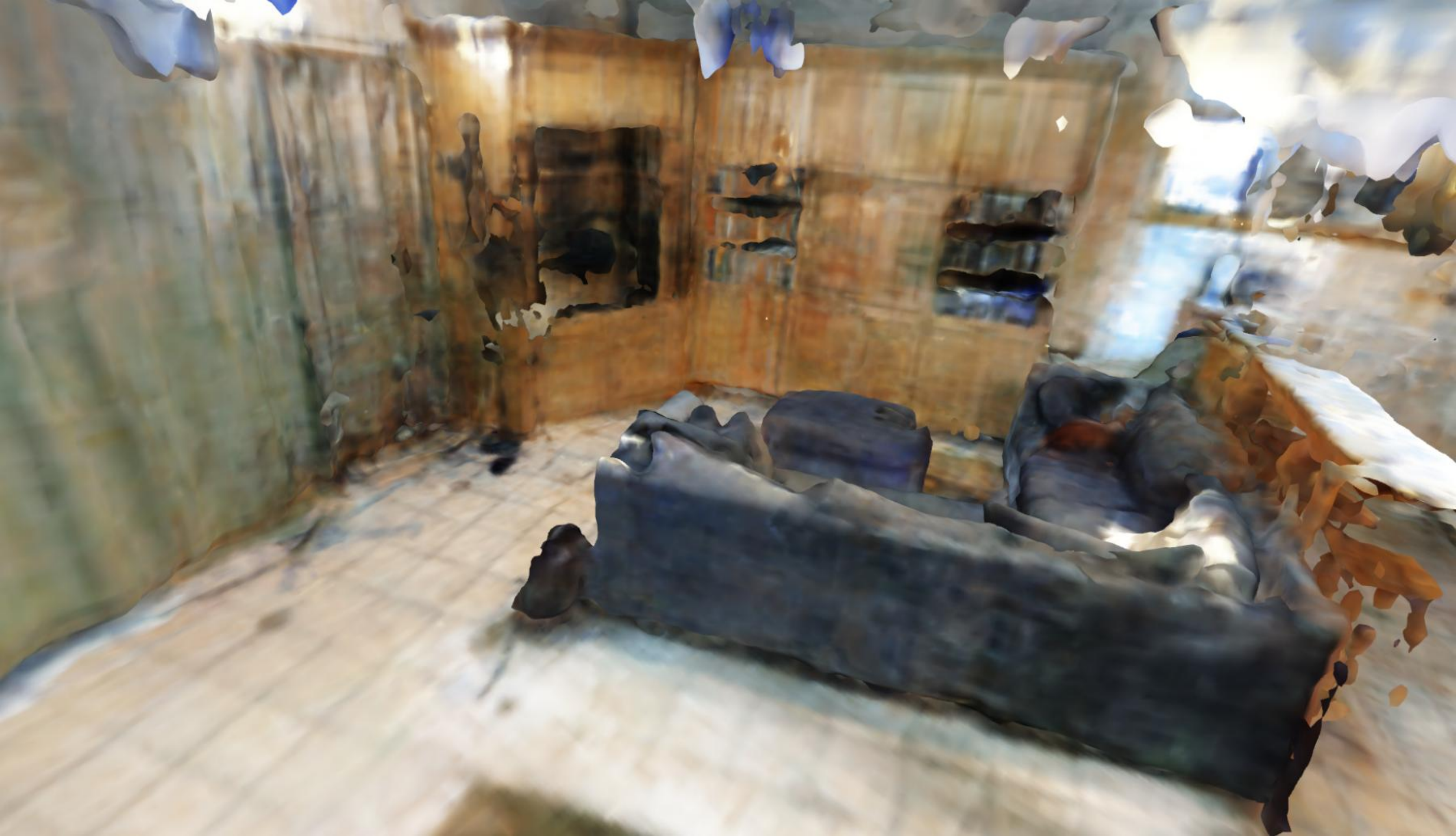}
      \end{subfigure}
      \begin{subfigure}{0.31\linewidth}
            \includegraphics[width=1\textwidth]{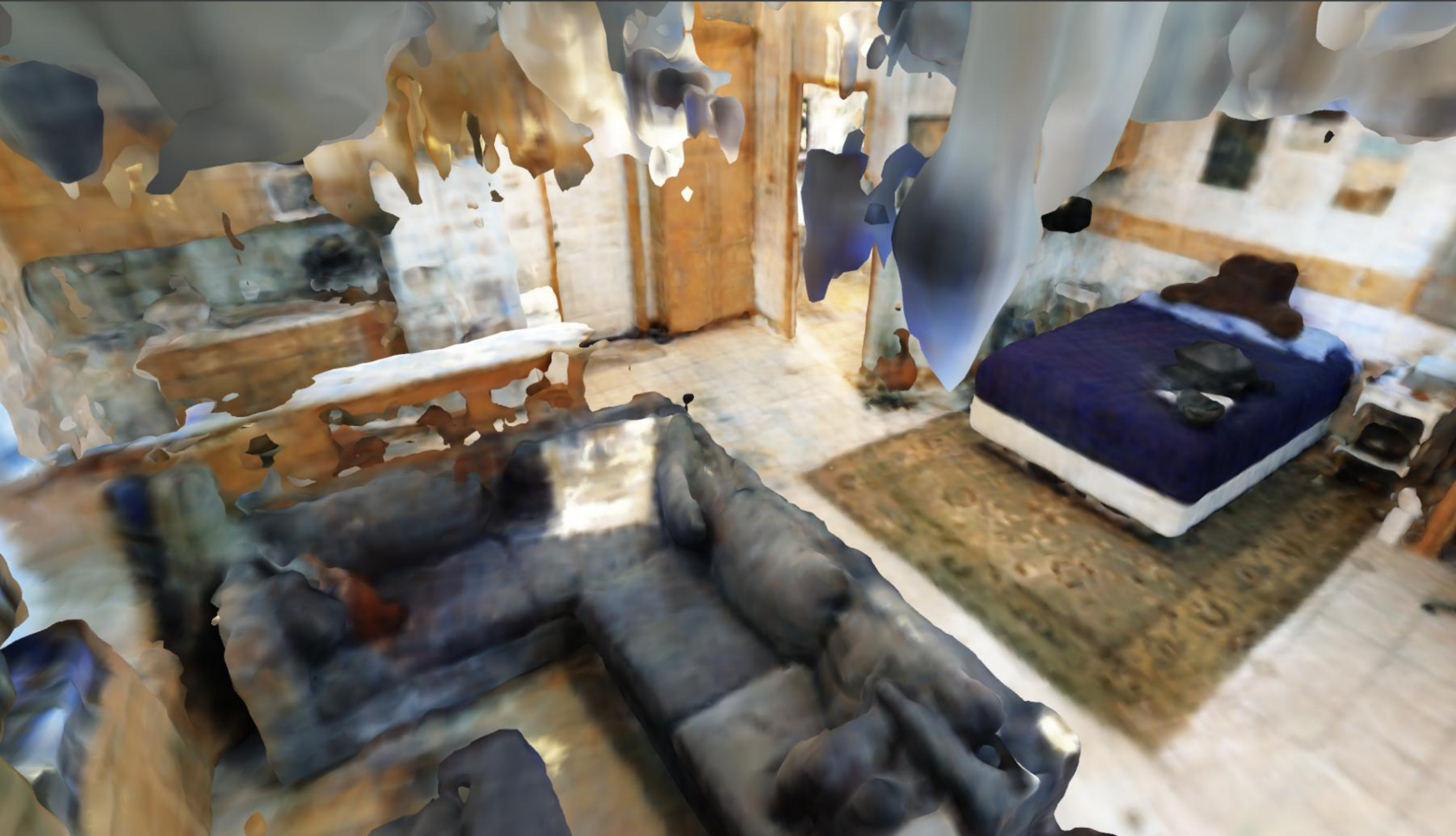}
      \end{subfigure}
      \begin{subfigure}{0.31\linewidth}
            \includegraphics[width=1\textwidth]{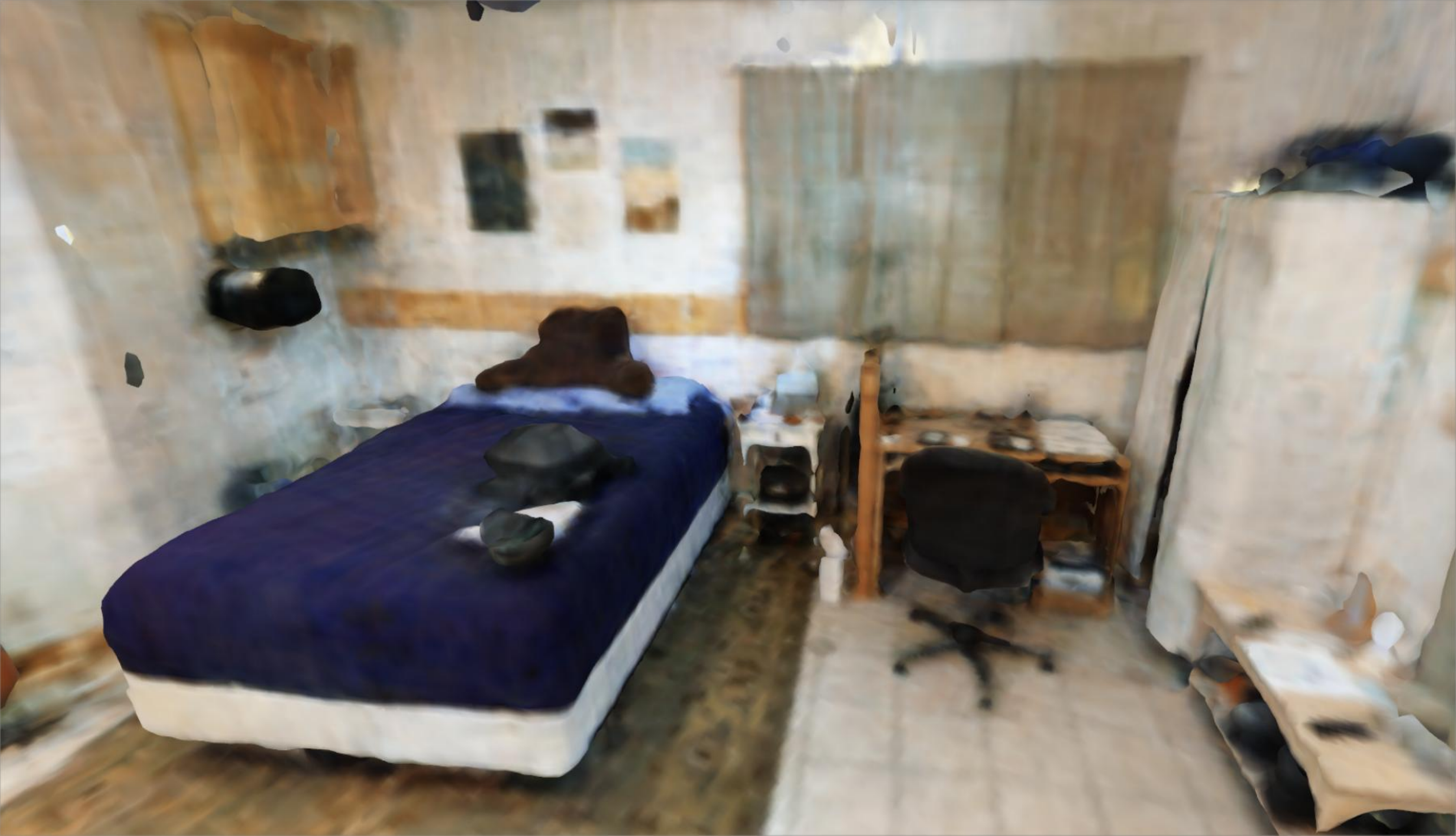}
      \end{subfigure}

      \rotatebox[origin=lb]{90}{\hspace{2mm} \scriptsize{Co-SLAM\cite{wang2023coslam}}}
      \begin{subfigure}{0.31\linewidth}
            \includegraphics[width=1\textwidth]{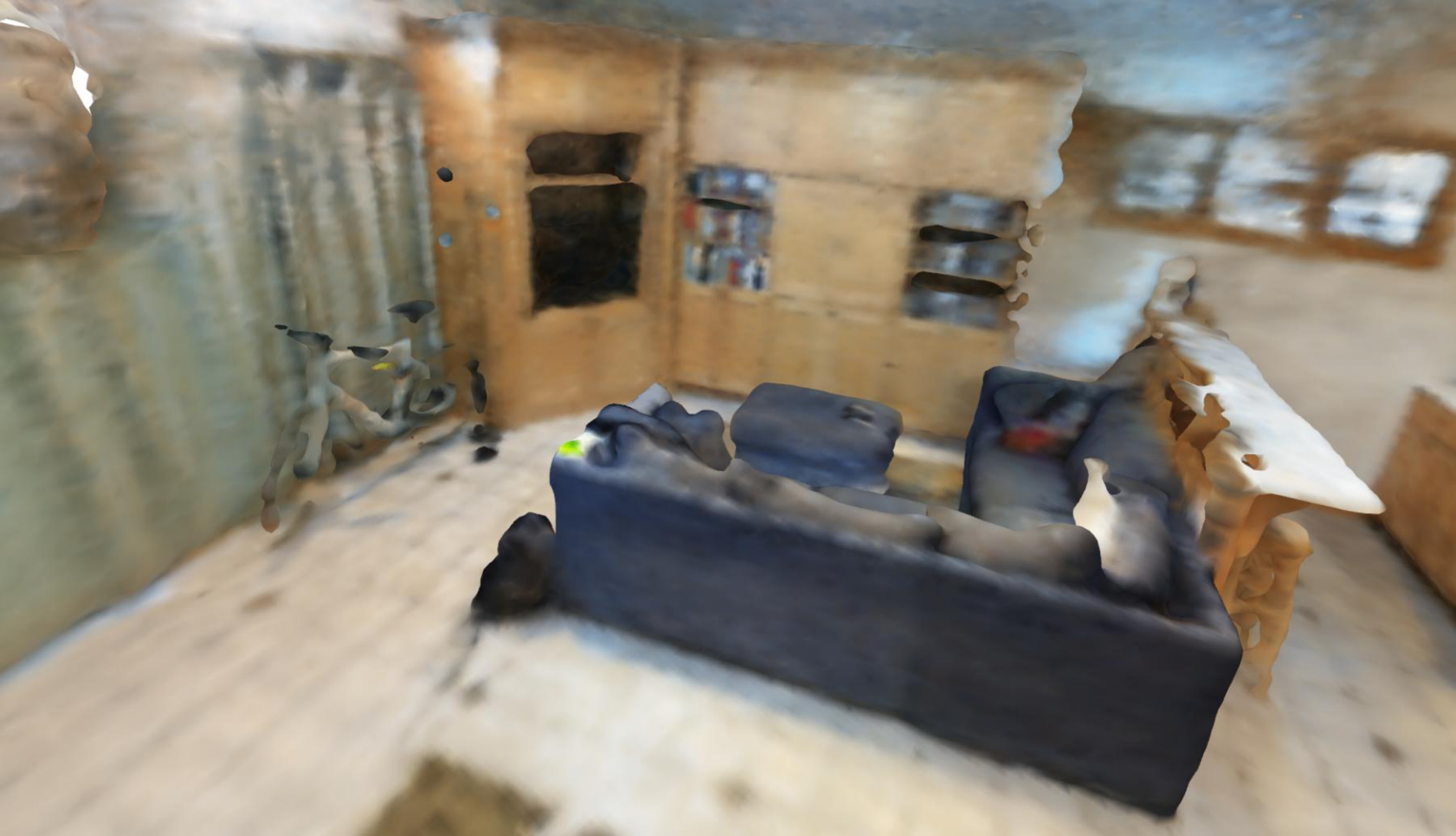}
      \end{subfigure}
      \begin{subfigure}{0.31\linewidth}
            \includegraphics[width=1\textwidth]{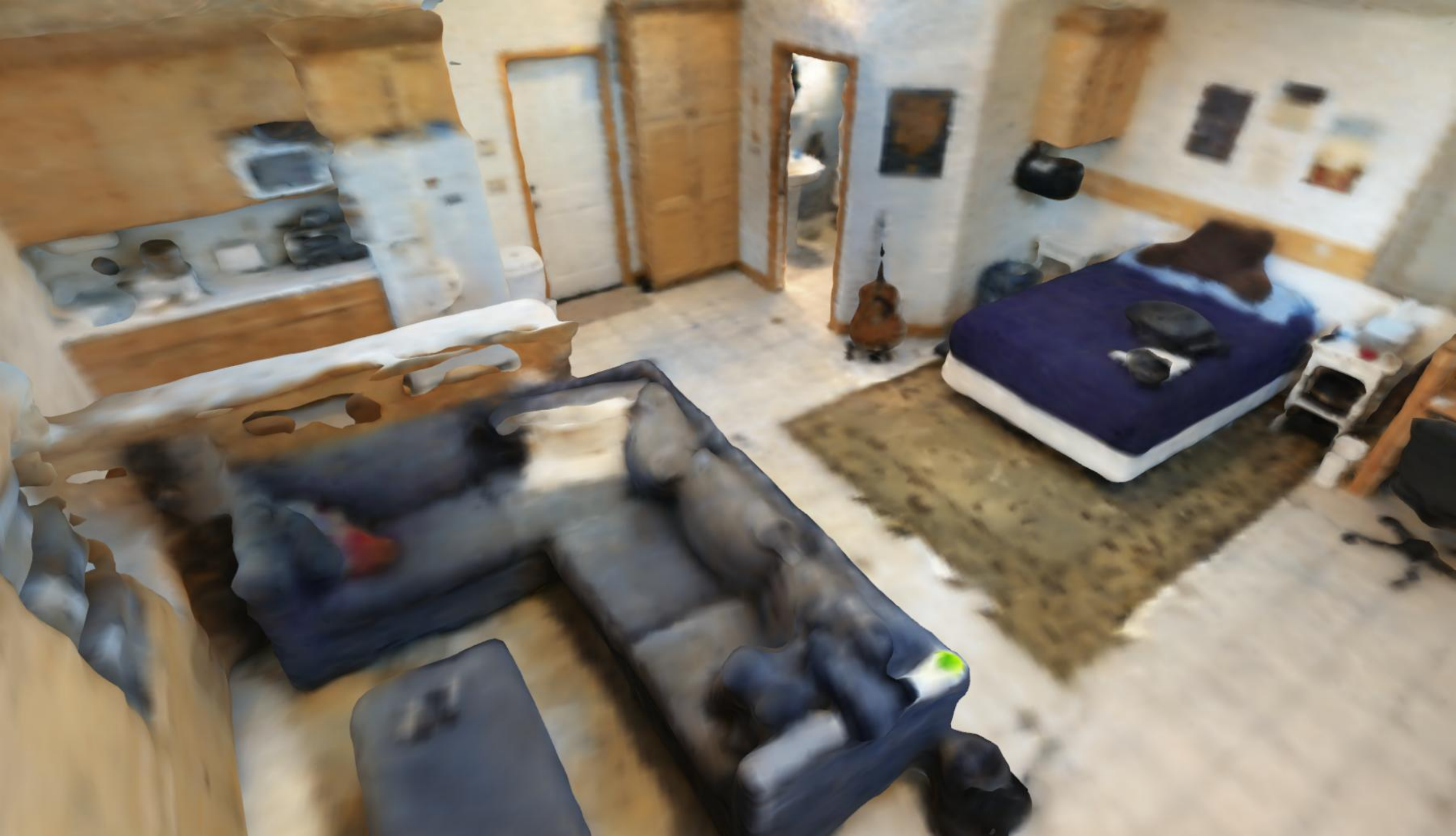}
      \end{subfigure}
      \begin{subfigure}{0.31\linewidth}
            \includegraphics[width=1\textwidth]{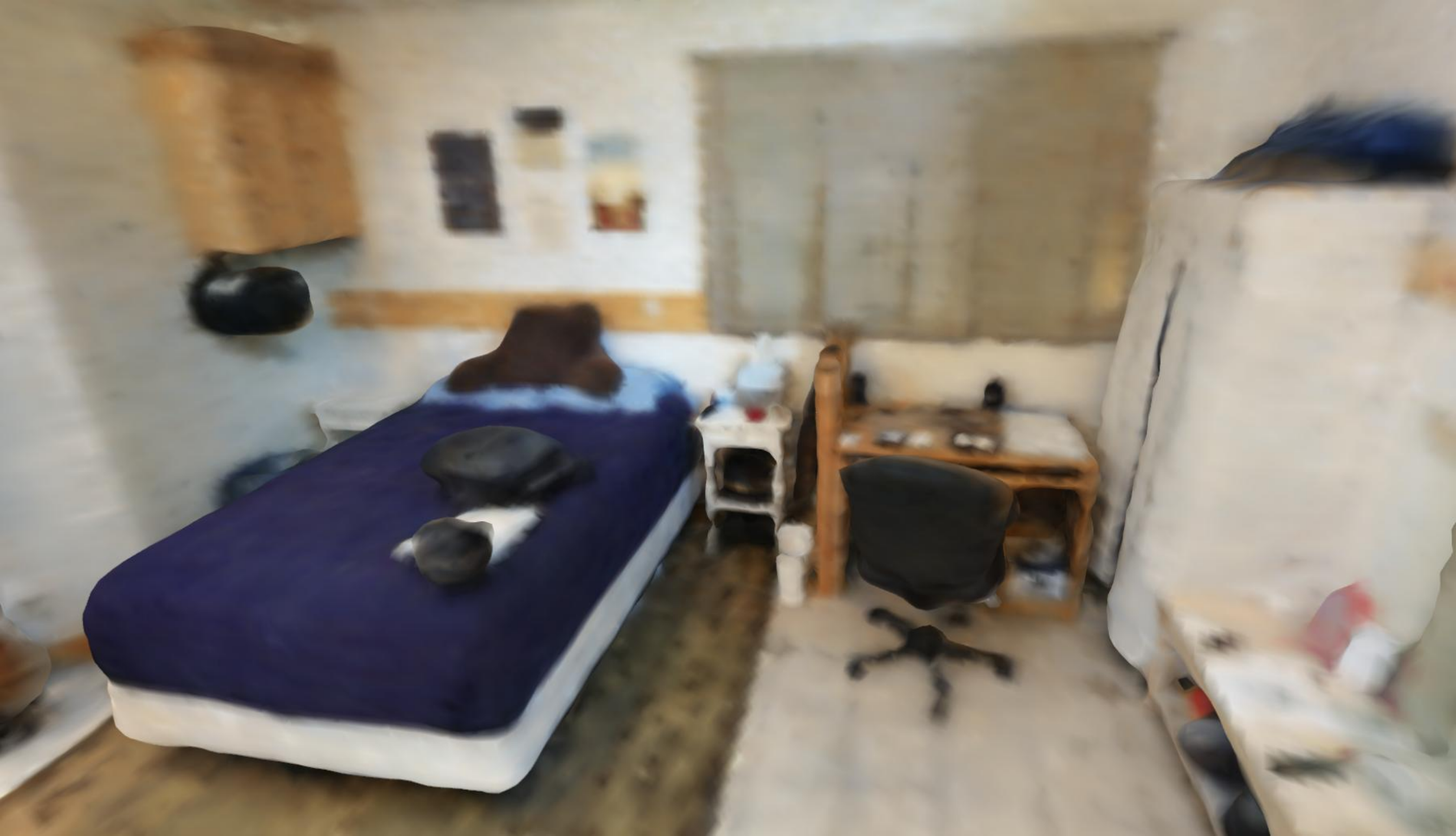}
      \end{subfigure}

      \rotatebox[origin=lb]{90}{\hspace{8mm} \footnotesize{Ours}}
      \begin{subfigure}{0.31\linewidth}
            \includegraphics[width=1\textwidth]{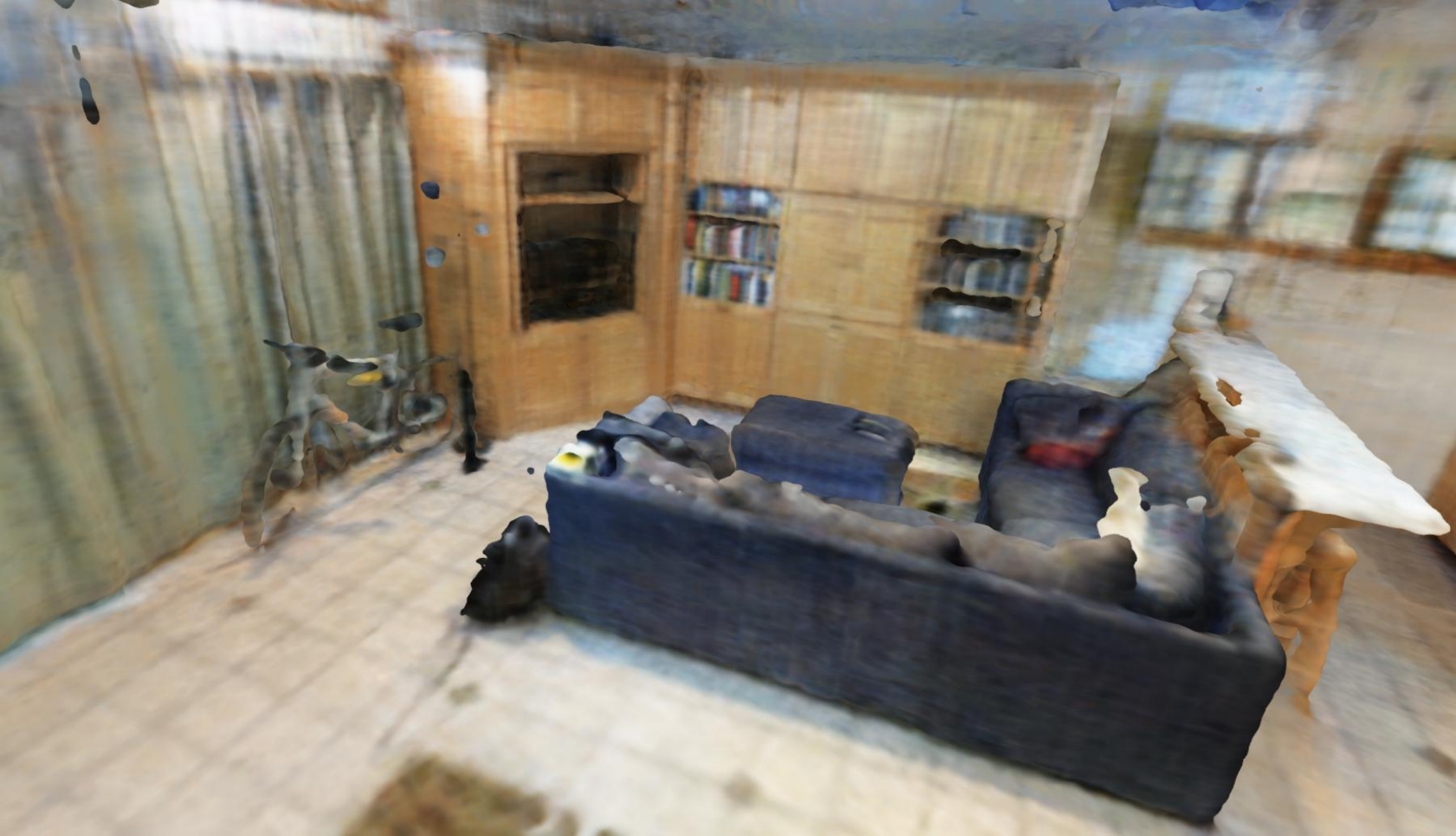}
      \end{subfigure}
      \begin{subfigure}{0.31\linewidth}
            \includegraphics[width=1\textwidth]{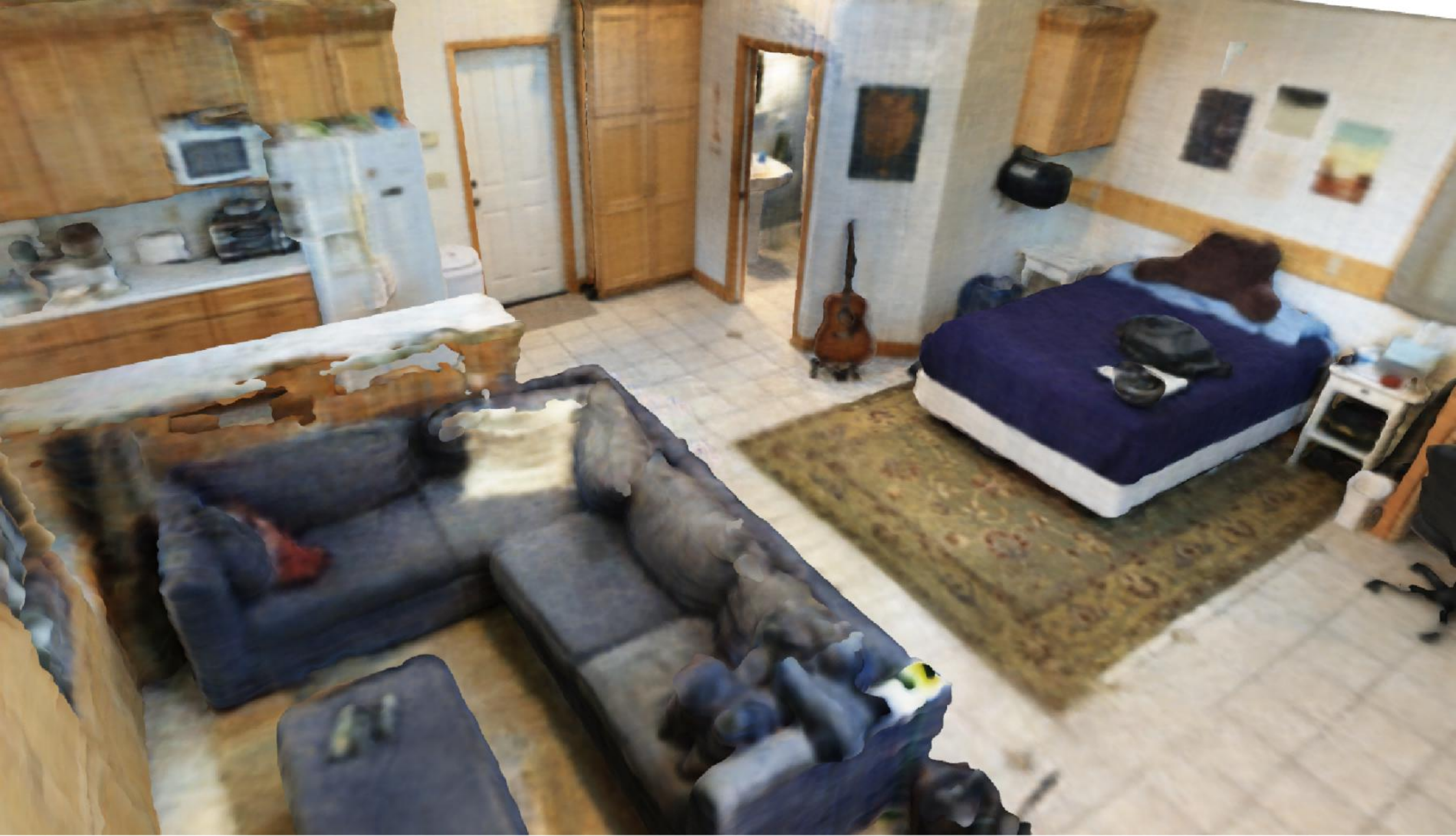}
      \end{subfigure}
      \begin{subfigure}{0.31\linewidth}
            \includegraphics[width=1\textwidth]{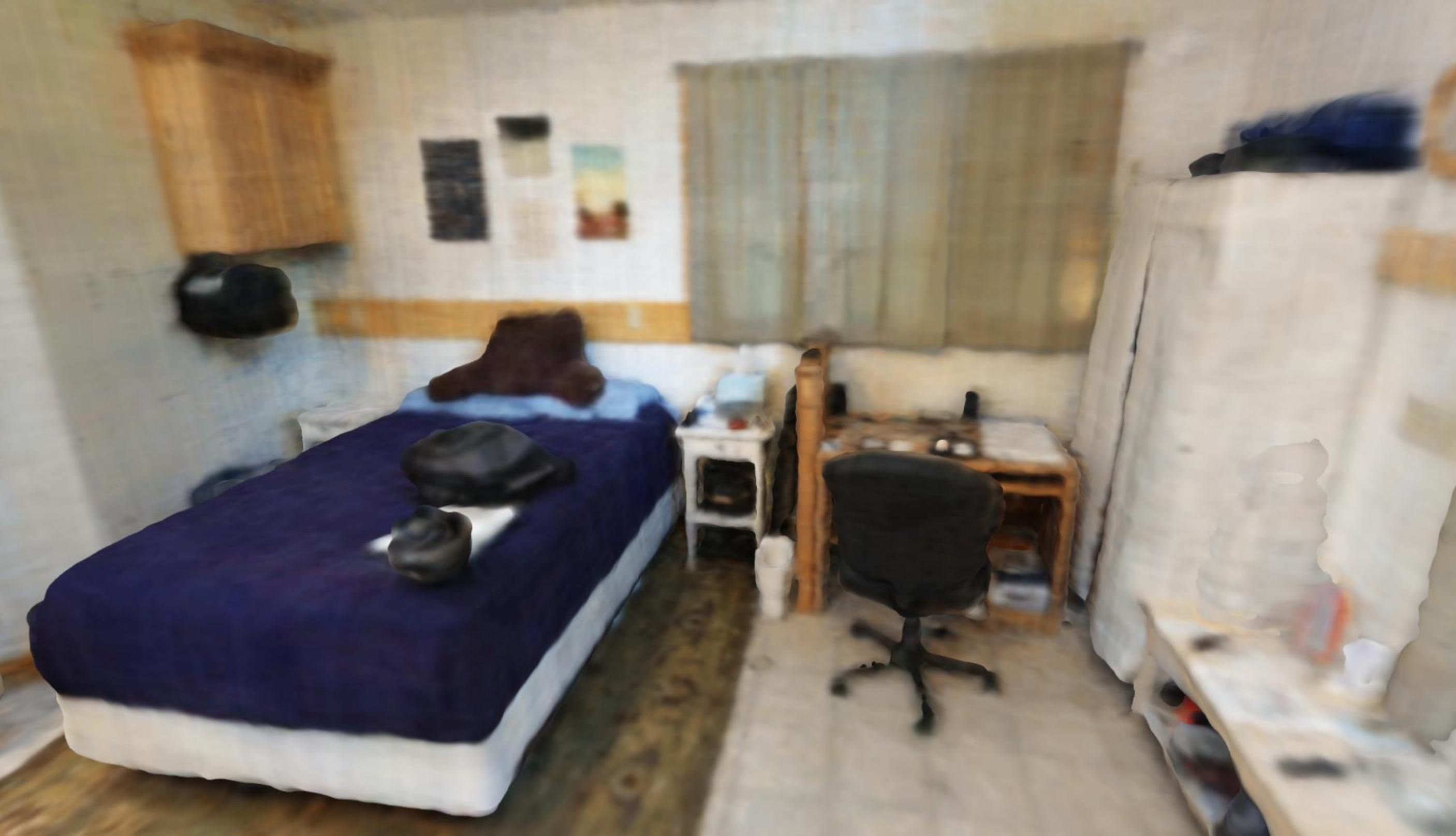}
      \end{subfigure}

      \rotatebox[origin=lb]{90}{\hspace{7mm} \footnotesize{GT}}
      \begin{subfigure}{0.31\linewidth}
            \includegraphics[width=1\textwidth]{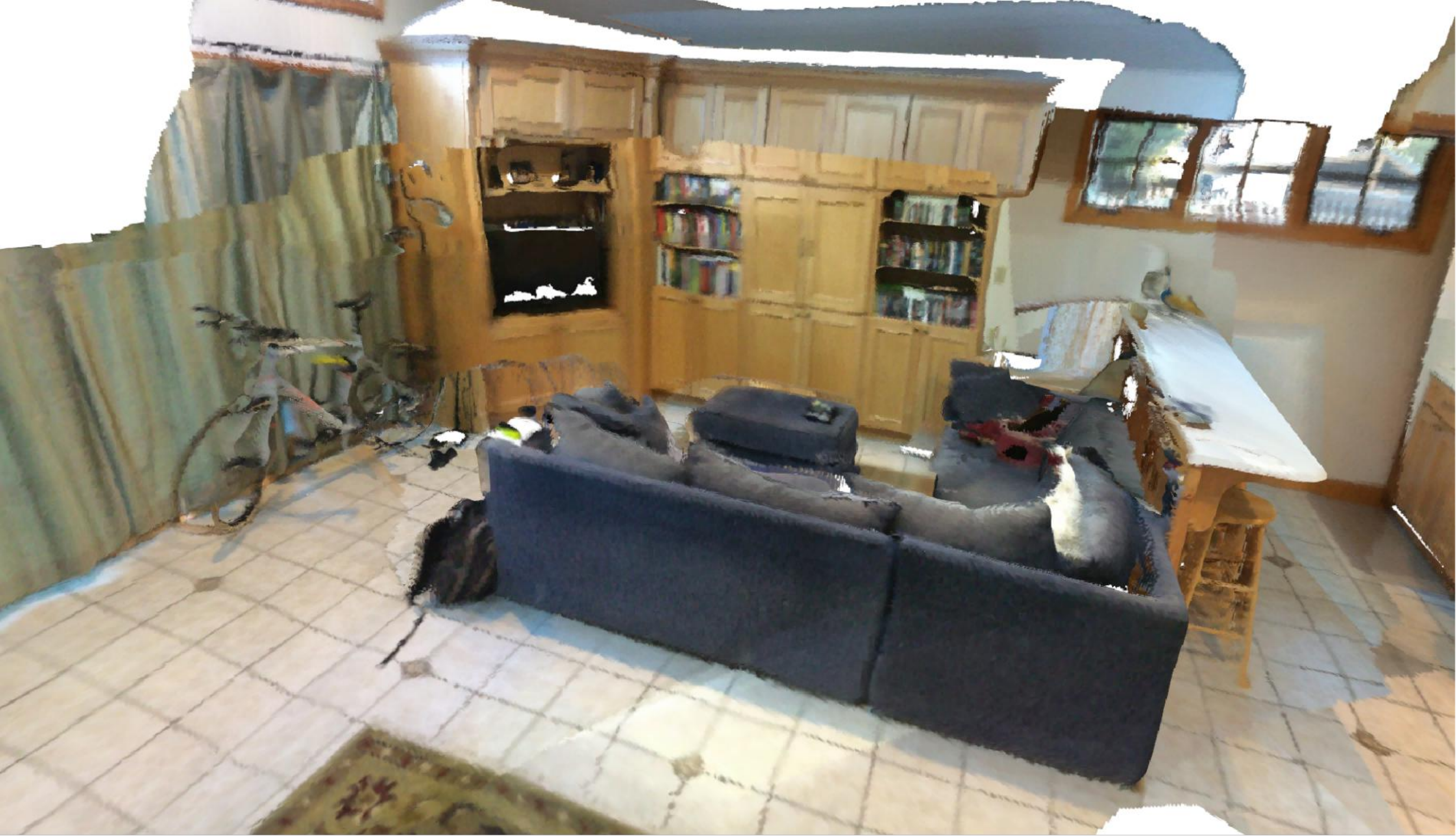}
      \end{subfigure}
      \begin{subfigure}{0.31\linewidth}
            \includegraphics[width=1\textwidth]{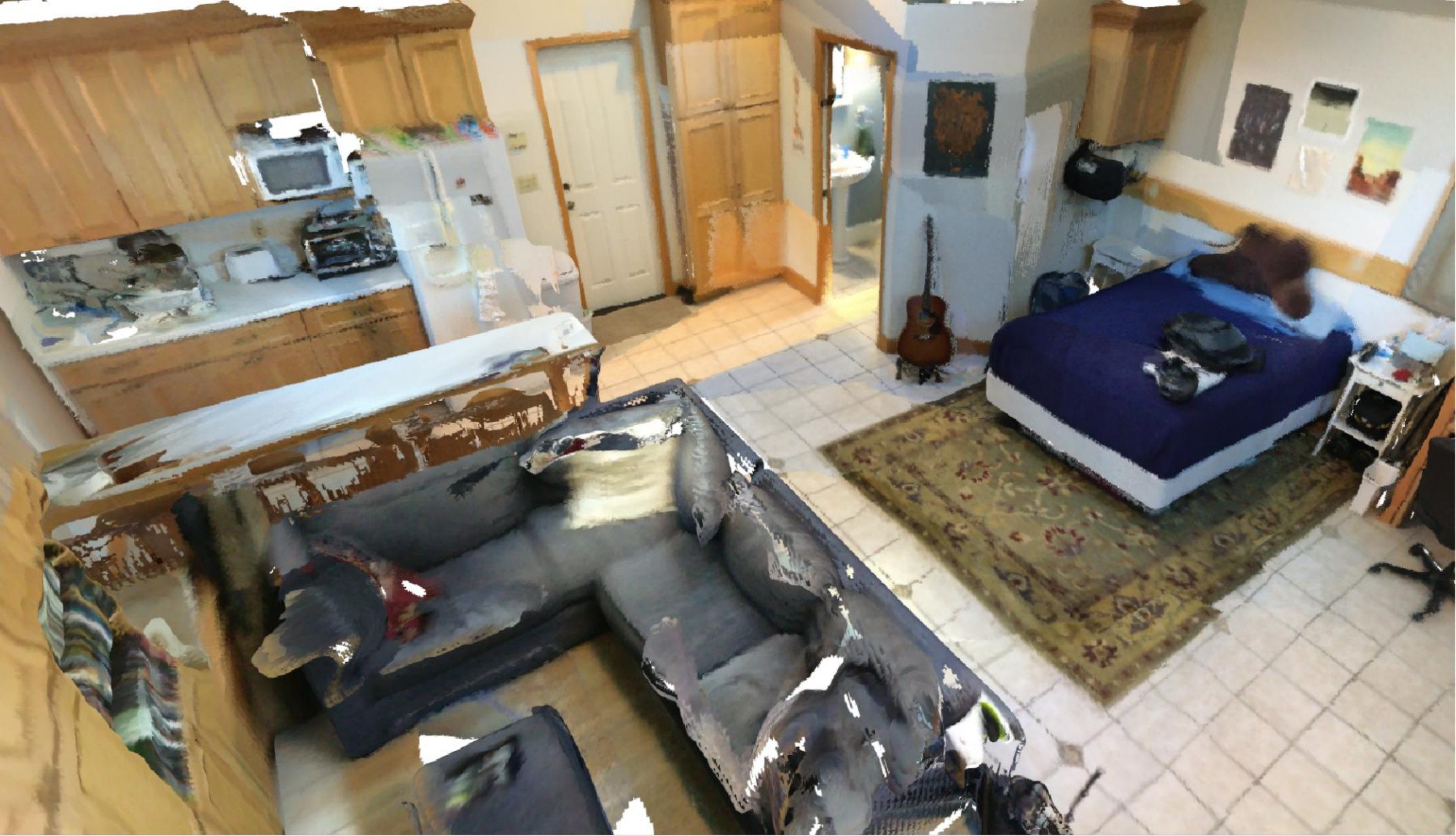}
      \end{subfigure}
      \begin{subfigure}{0.31\linewidth}
            \includegraphics[width=1\textwidth]{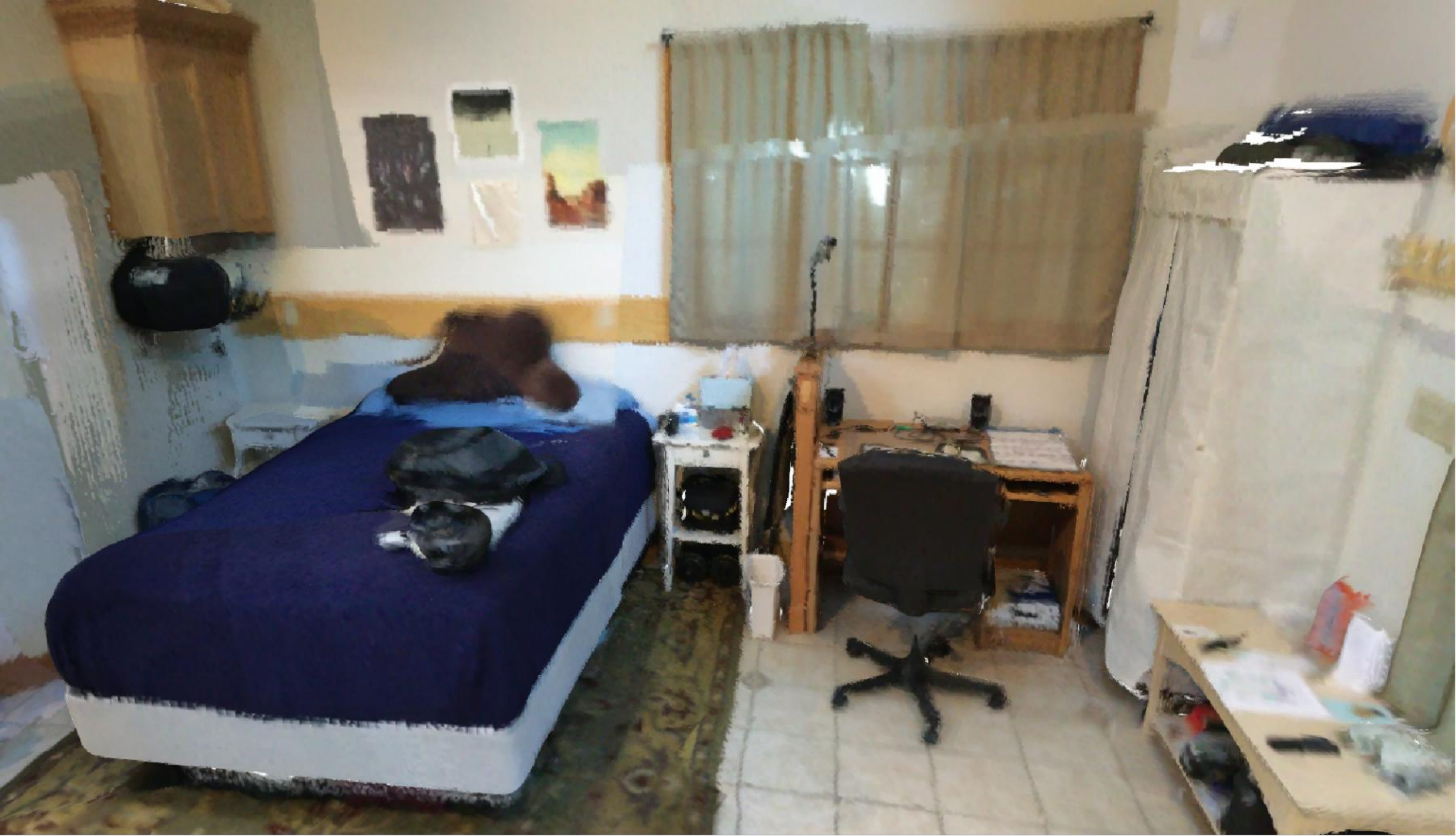}
      \end{subfigure}
      \caption{The qualitative results on ScanNet\cite{dai2017scannet} demonstrate that our method can accurately reconstruct large indoor rooms, capturing complex appearances and geometries. From different views, our method consistently produces competitive results in both geometry and appearance reconstruction.}
       \label{fig:scan_1}
\end{figure*}

\begin{figure*}[!t]
      \centering
      \captionsetup[subfigure]{labelformat=empty}

      \rotatebox[origin=lb]{90}{\scriptsize{NICE-SLAM\cite{zhu2022nice}}}
      \begin{subfigure}{0.31\linewidth}
            \centering
            \captionsetup{justification=raggedright}
            \includegraphics[width=1\textwidth]{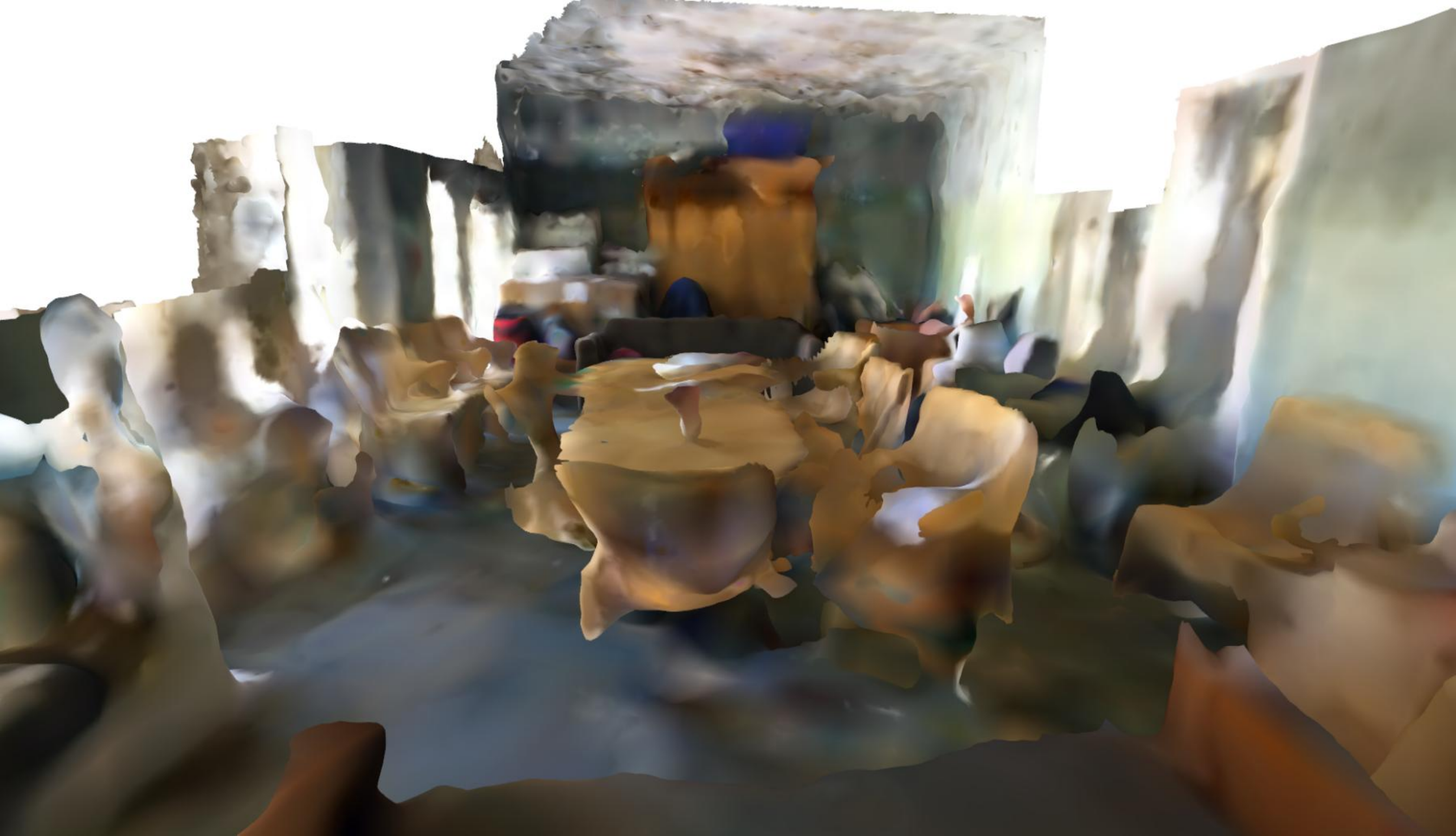}
      \end{subfigure}
      \begin{subfigure}{0.31\linewidth}
            \centering
            \captionsetup{justification=raggedright}
            \includegraphics[width=1\textwidth]{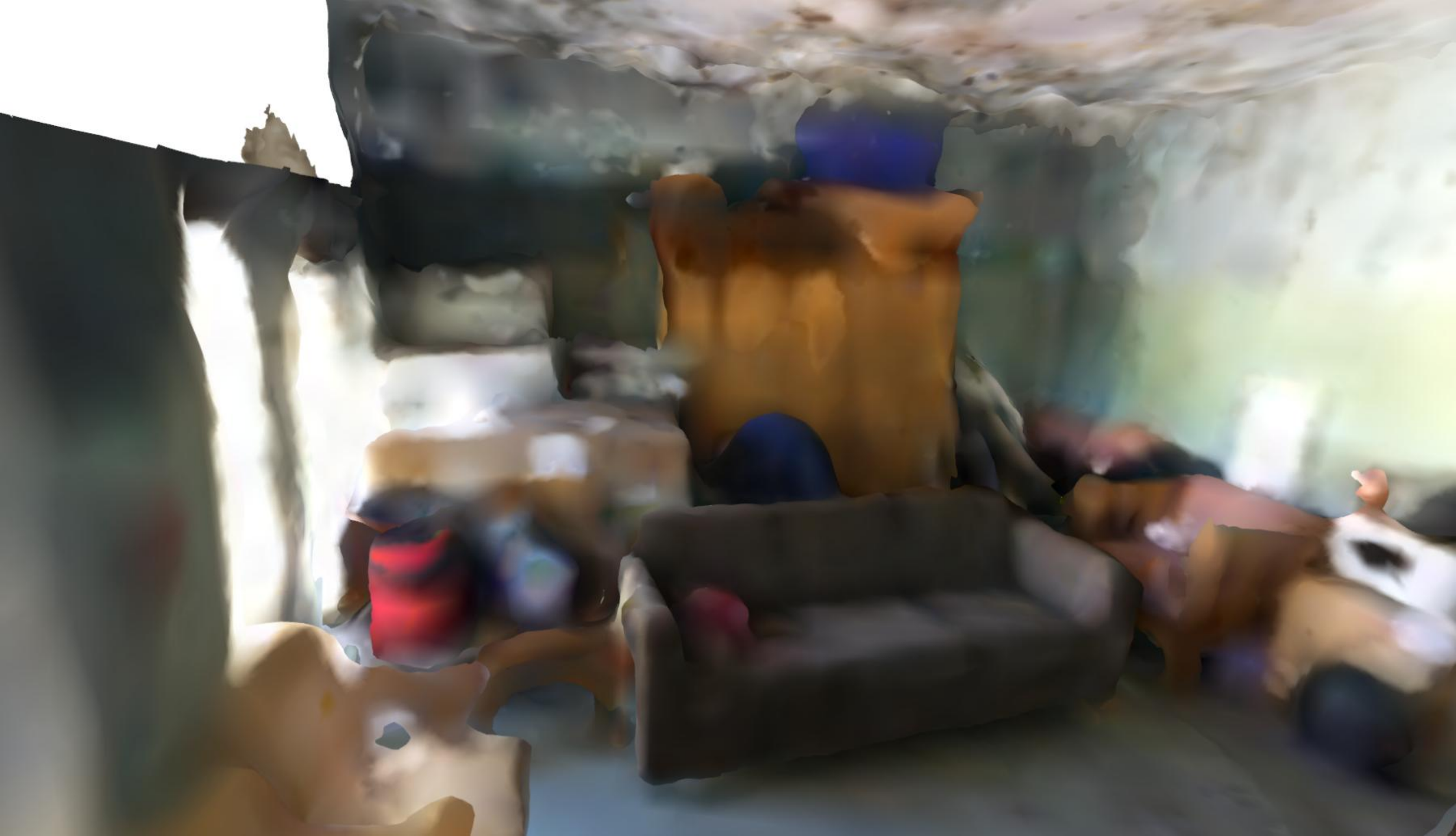}
      \end{subfigure}
      \begin{subfigure}{0.31\linewidth}
            \captionsetup{justification=raggedright}
            \includegraphics[width=1\textwidth]{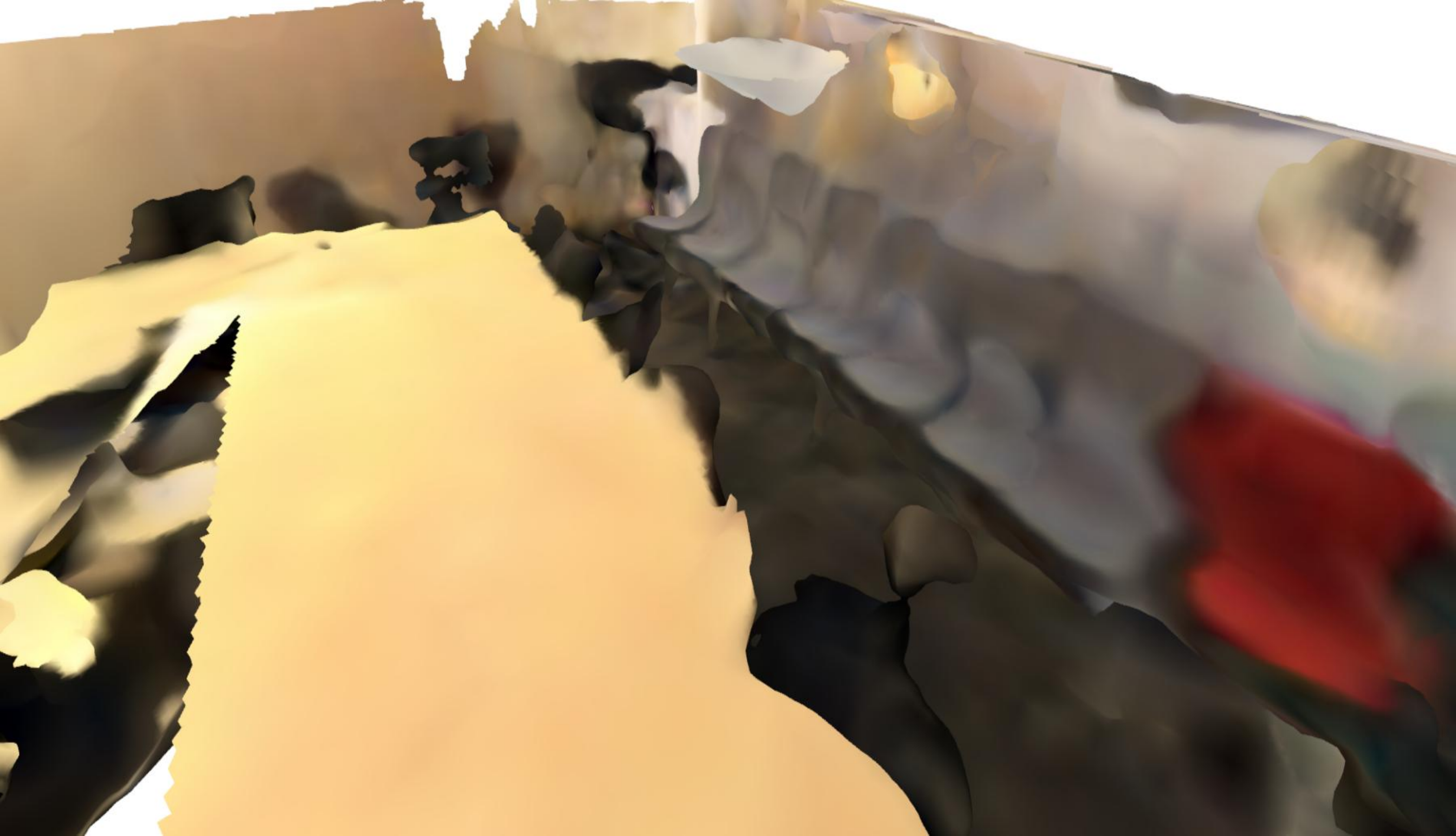}
      \end{subfigure}

      \rotatebox[origin=lb]{90}{\hspace{3mm} \scriptsize{ESLAM\cite{johari2023eslam}}}
      \begin{subfigure}{0.31\linewidth}
            \includegraphics[width=1\textwidth]{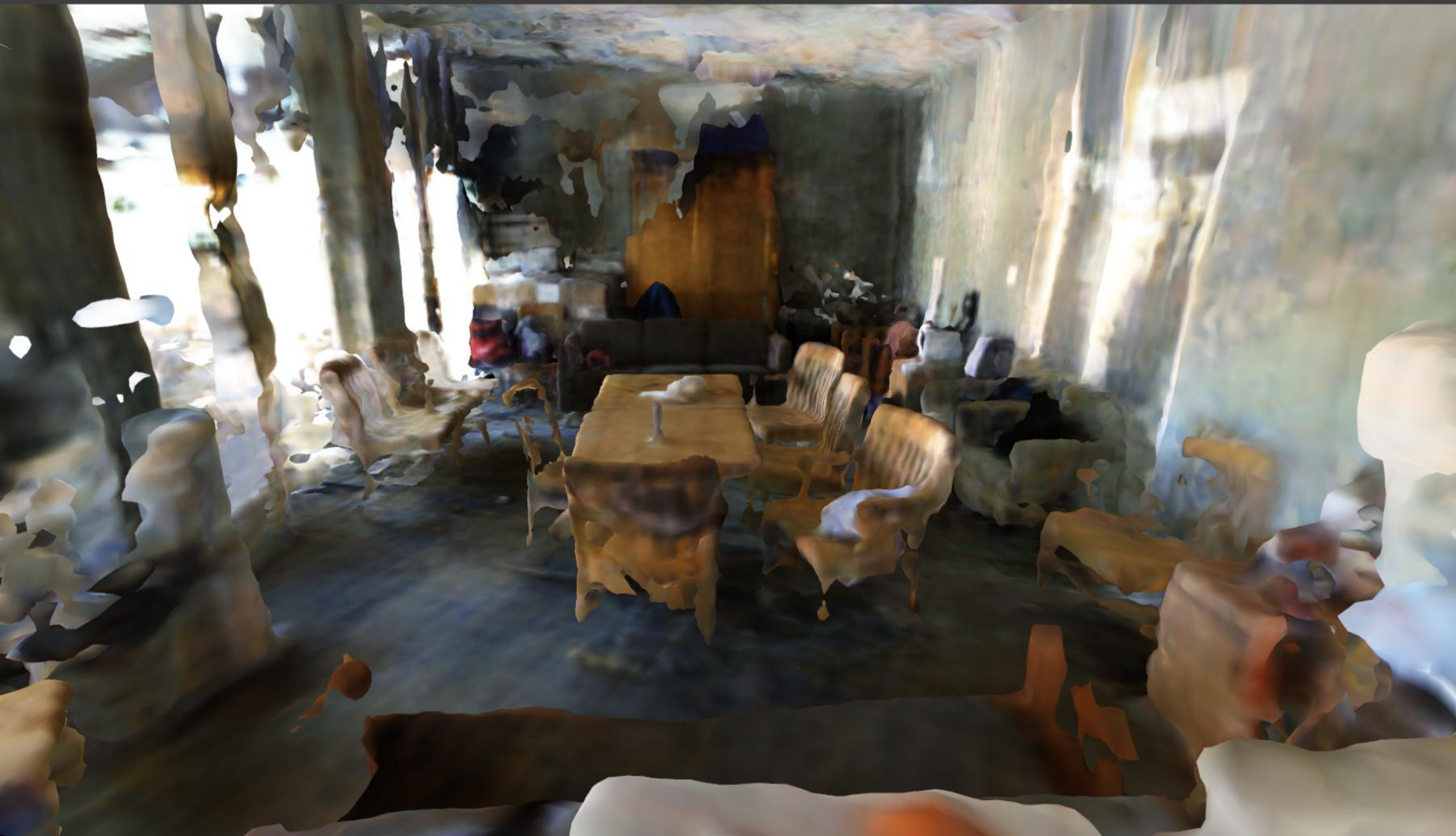}
      \end{subfigure}
      \begin{subfigure}{0.31\linewidth}
            \includegraphics[width=1\textwidth]{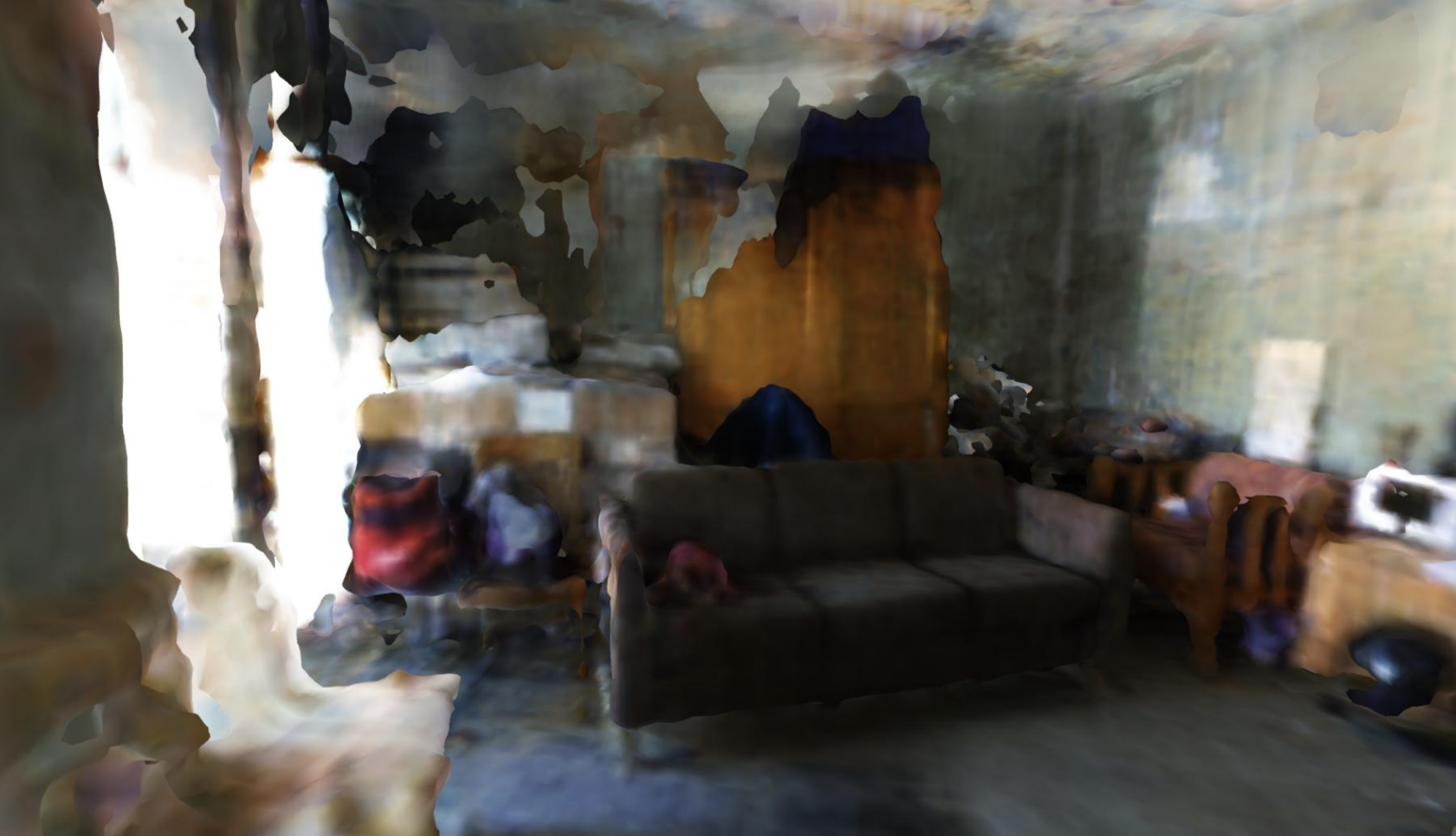}
      \end{subfigure}
      \begin{subfigure}{0.31\linewidth}
            \includegraphics[width=1\textwidth]{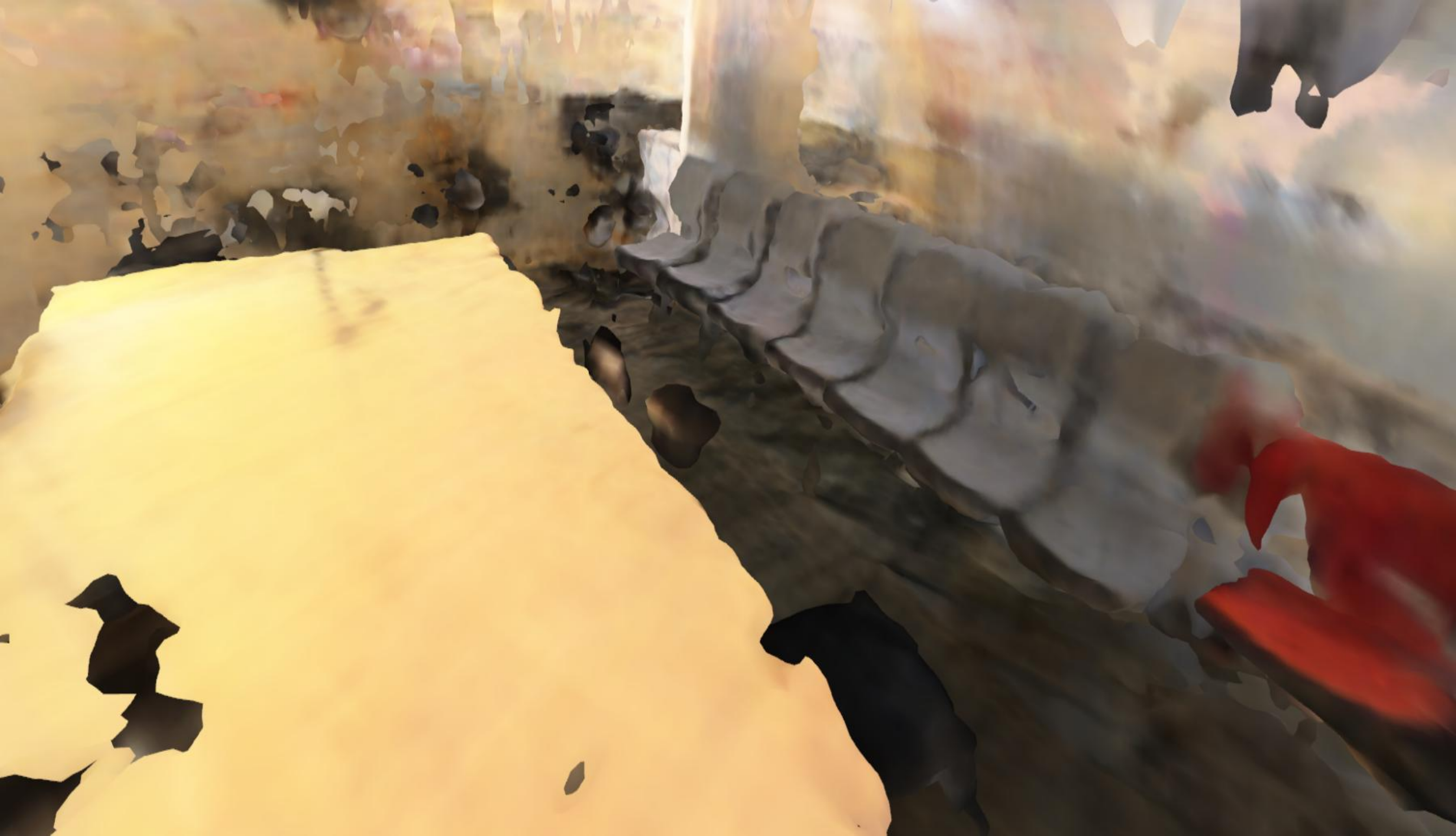}
      \end{subfigure}

      \rotatebox[origin=lb]{90}{\hspace{2mm} \scriptsize{Co-SLAM\cite{wang2023coslam}}}
      \begin{subfigure}{0.31\linewidth}
            \includegraphics[width=1\textwidth]{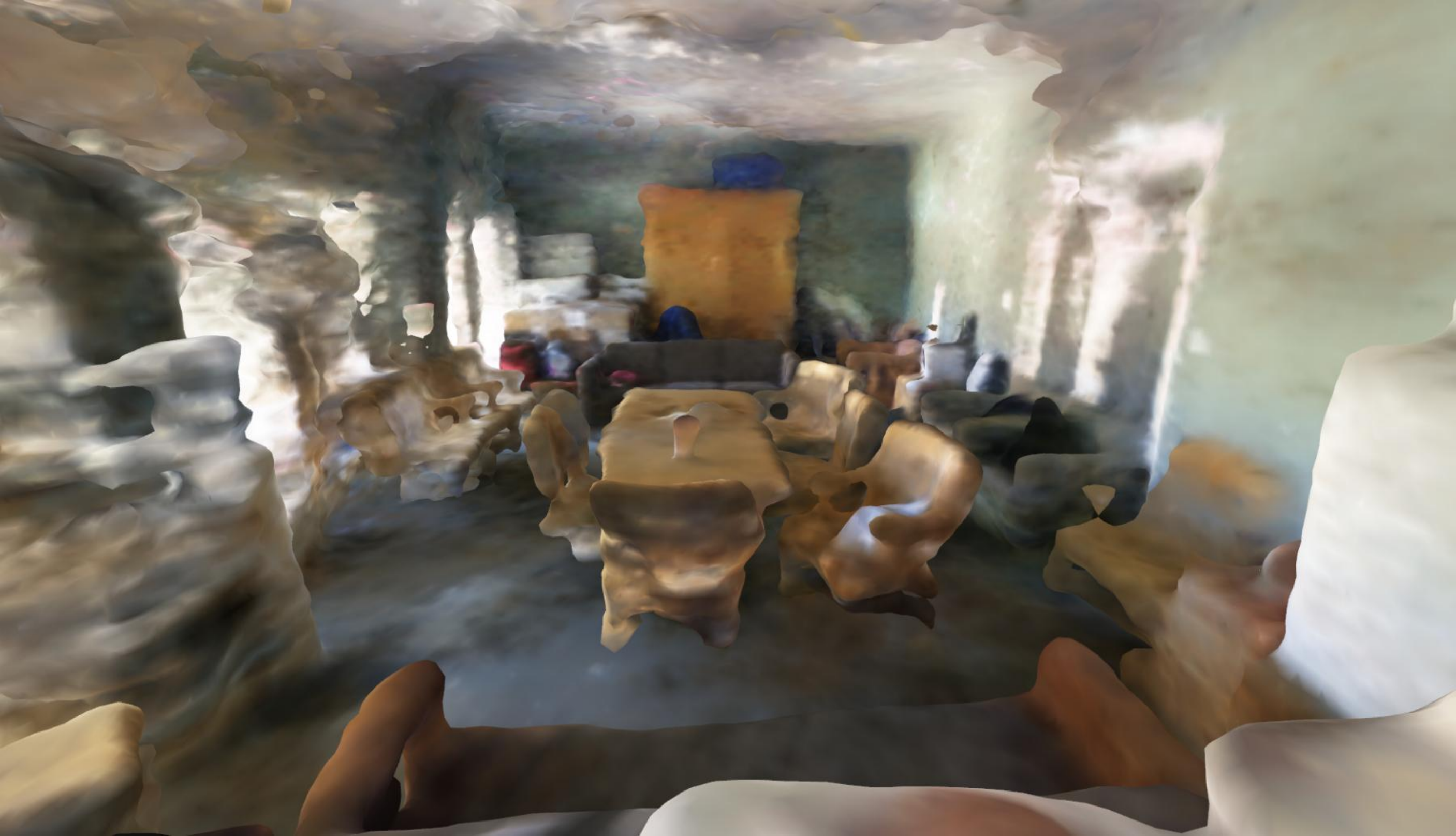}
      \end{subfigure}
      \begin{subfigure}{0.31\linewidth}
            \includegraphics[width=1\textwidth]{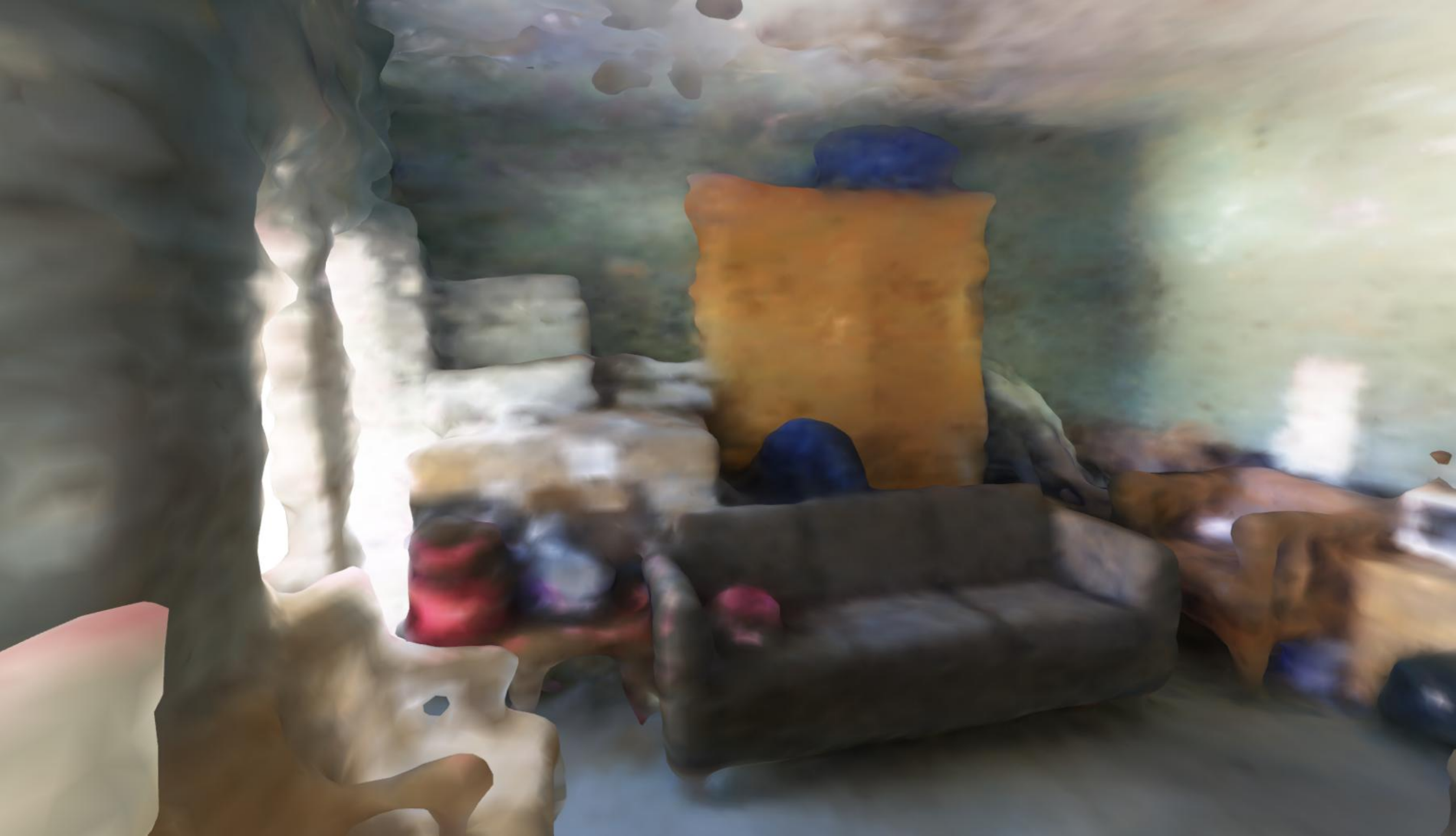}
      \end{subfigure}
      \begin{subfigure}{0.31\linewidth}
            \includegraphics[width=1\textwidth]{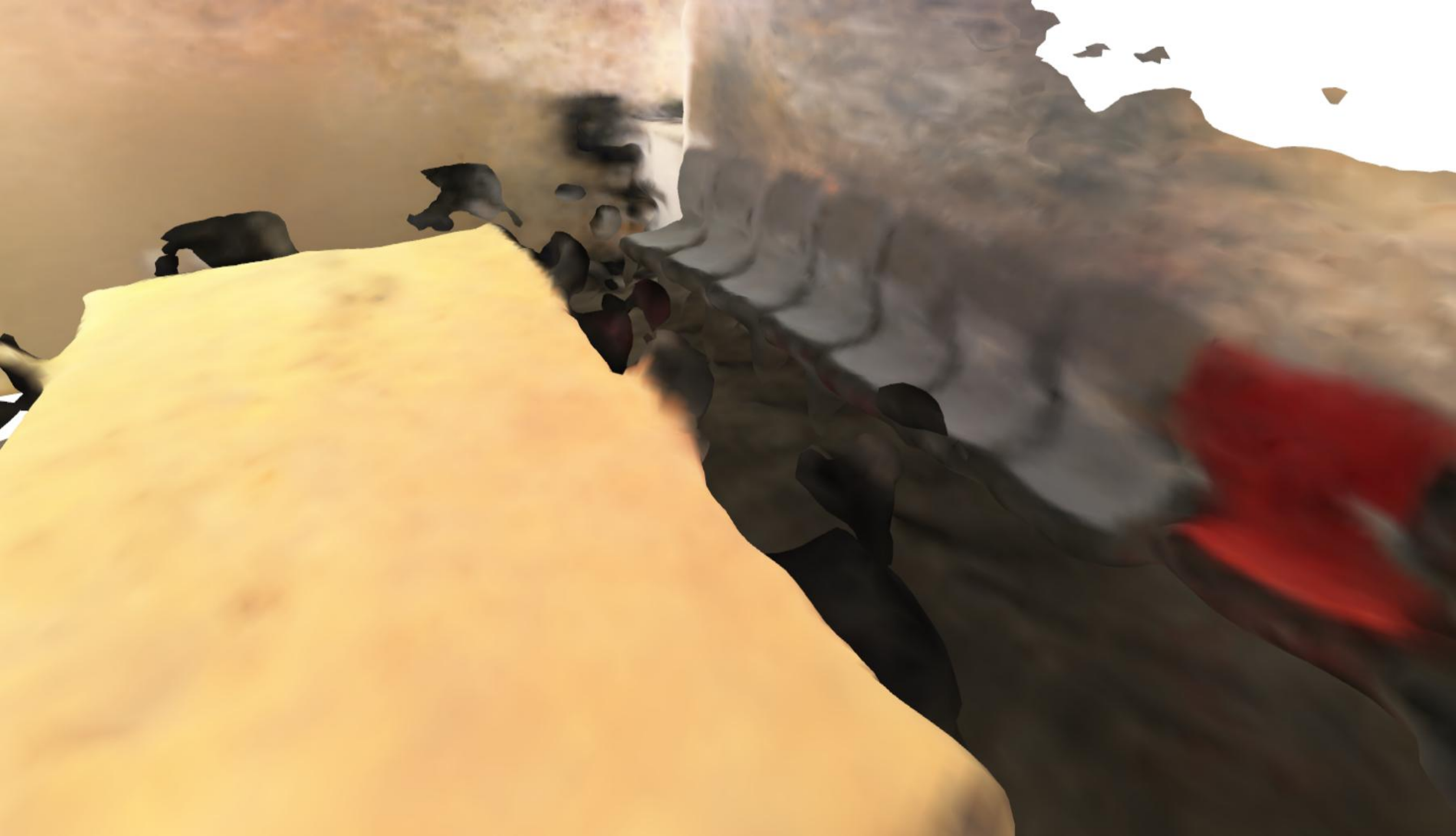}
      \end{subfigure}

      \rotatebox[origin=lb]{90}{\hspace{7mm} \footnotesize{Ours}}
      \begin{subfigure}{0.31\linewidth}
            \includegraphics[width=1\textwidth]{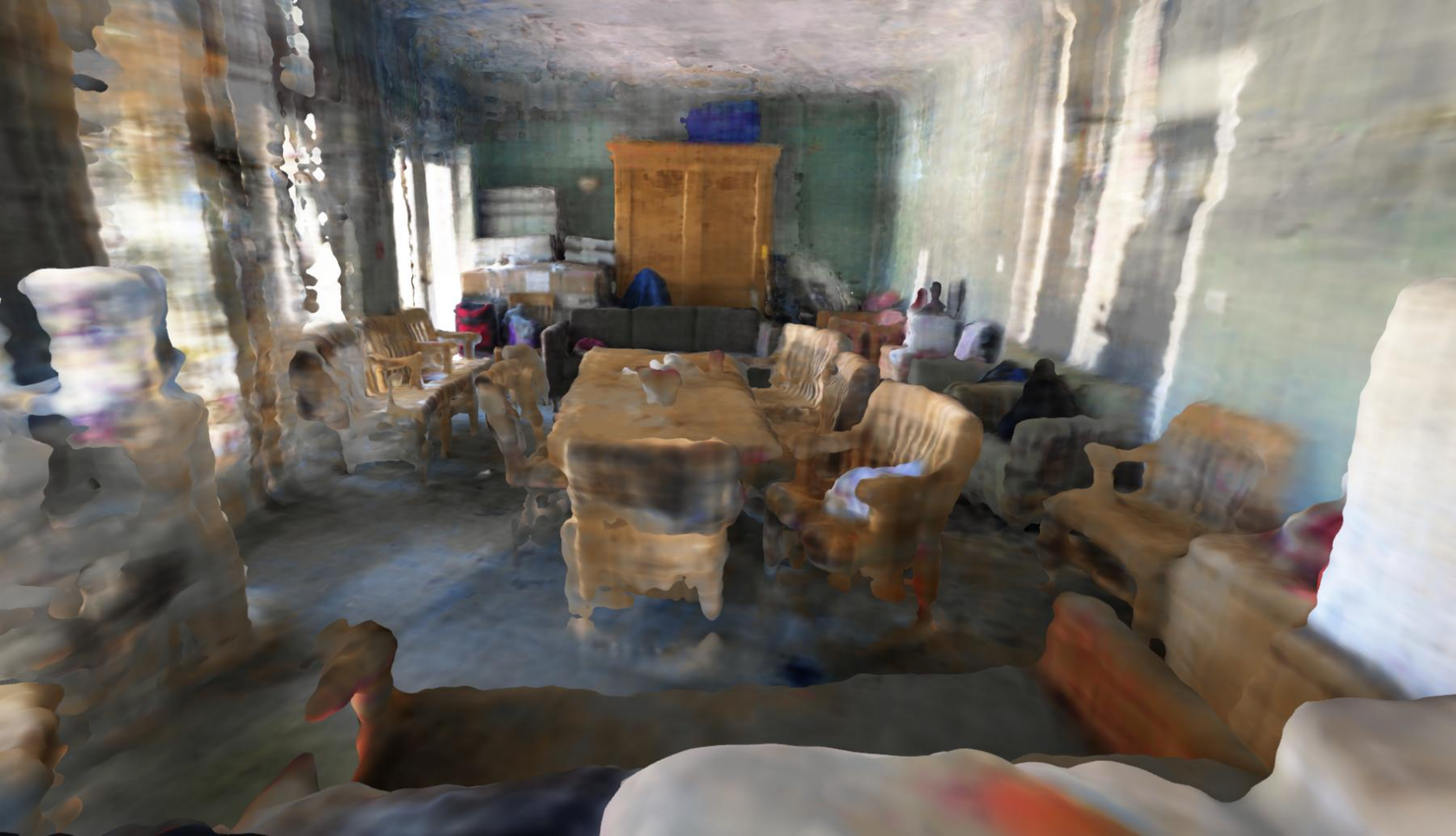}
      \end{subfigure}
      \begin{subfigure}{0.31\linewidth}
            \includegraphics[width=1\textwidth]{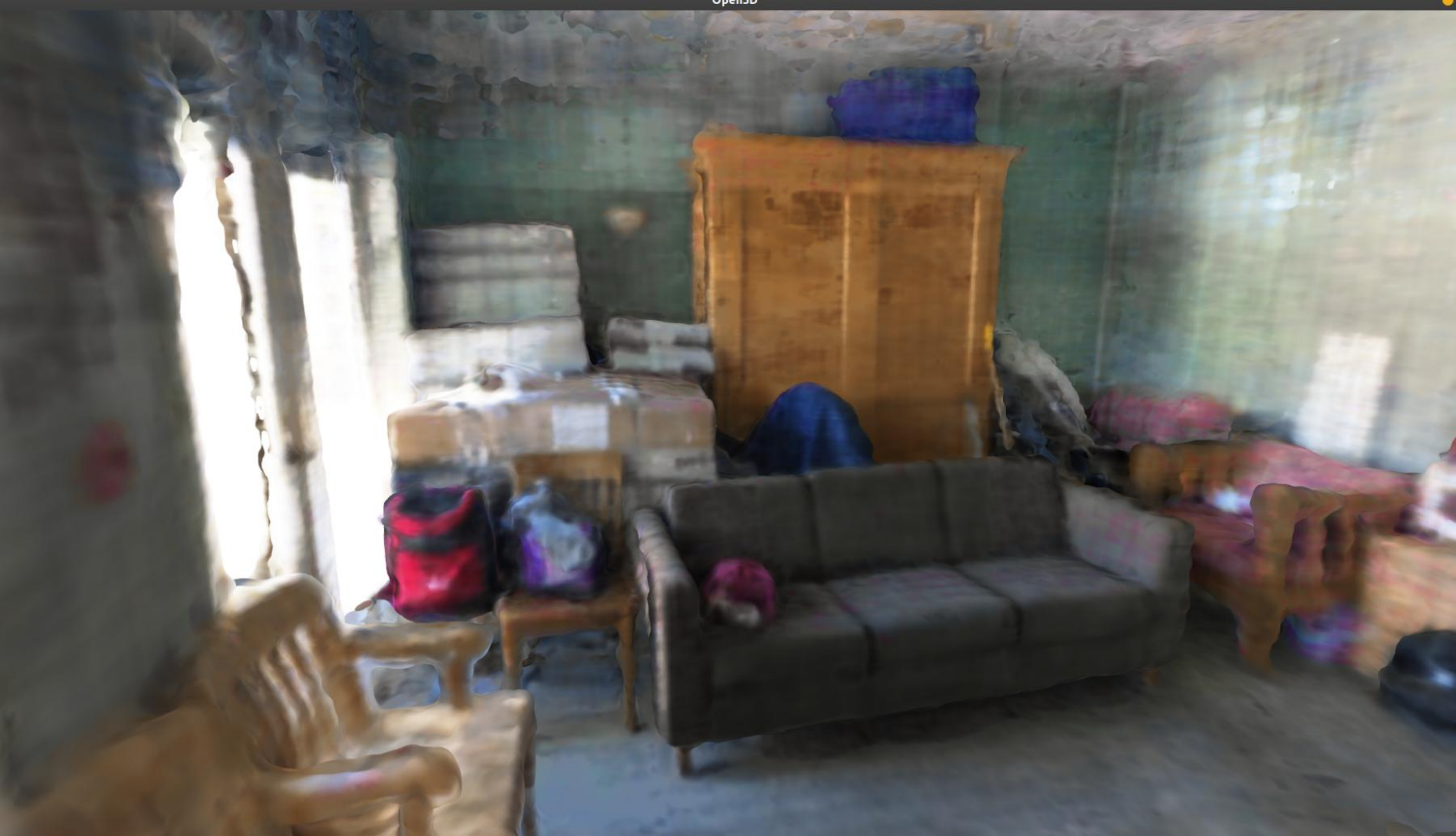}
      \end{subfigure}
      \begin{subfigure}{0.31\linewidth}
            \includegraphics[width=1\textwidth]{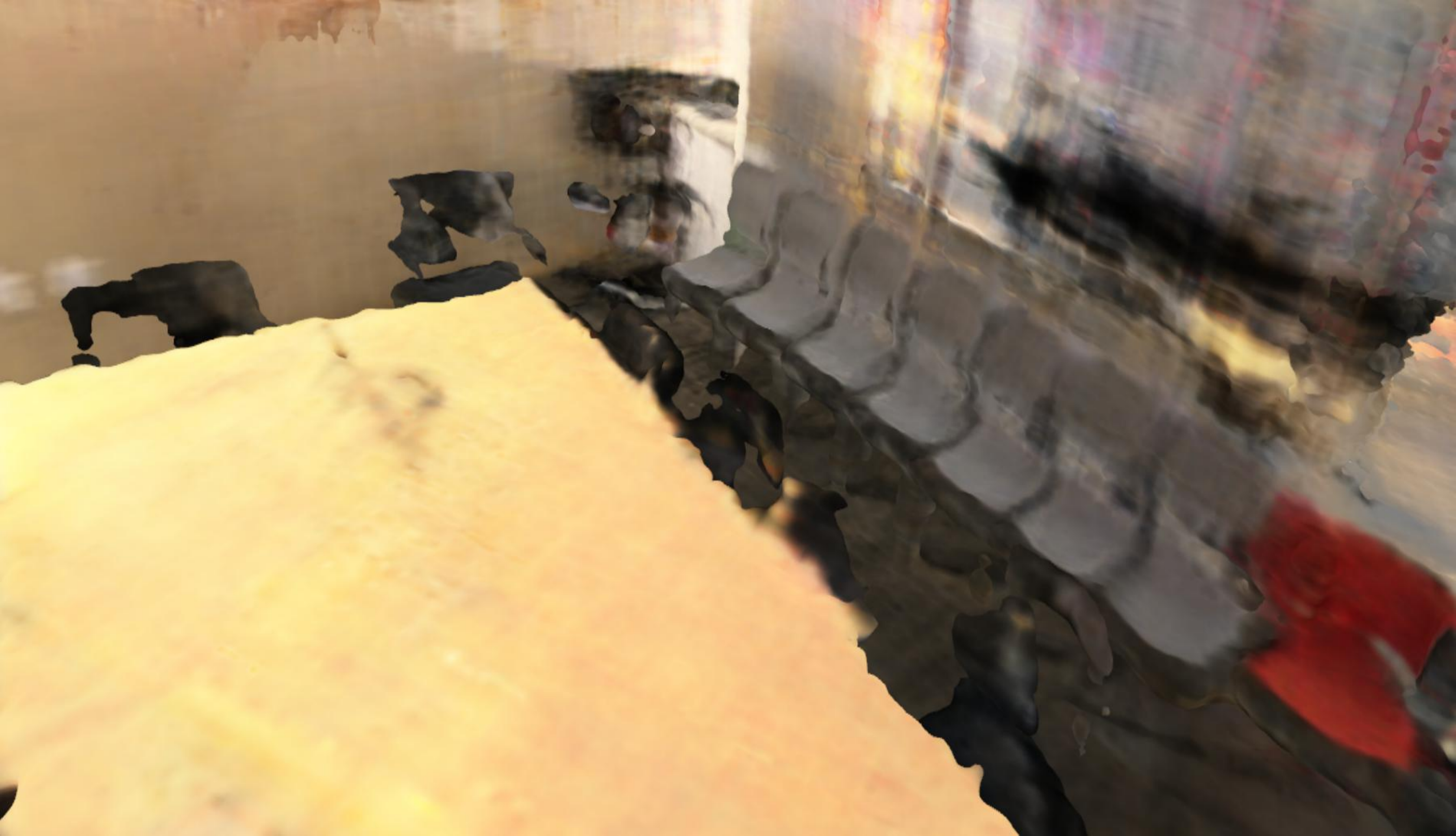}
      \end{subfigure}

      \rotatebox[origin=lb]{90}{\hspace{8mm} \footnotesize{GT}}
      \begin{subfigure}{0.31\linewidth}
            \includegraphics[width=1\textwidth]{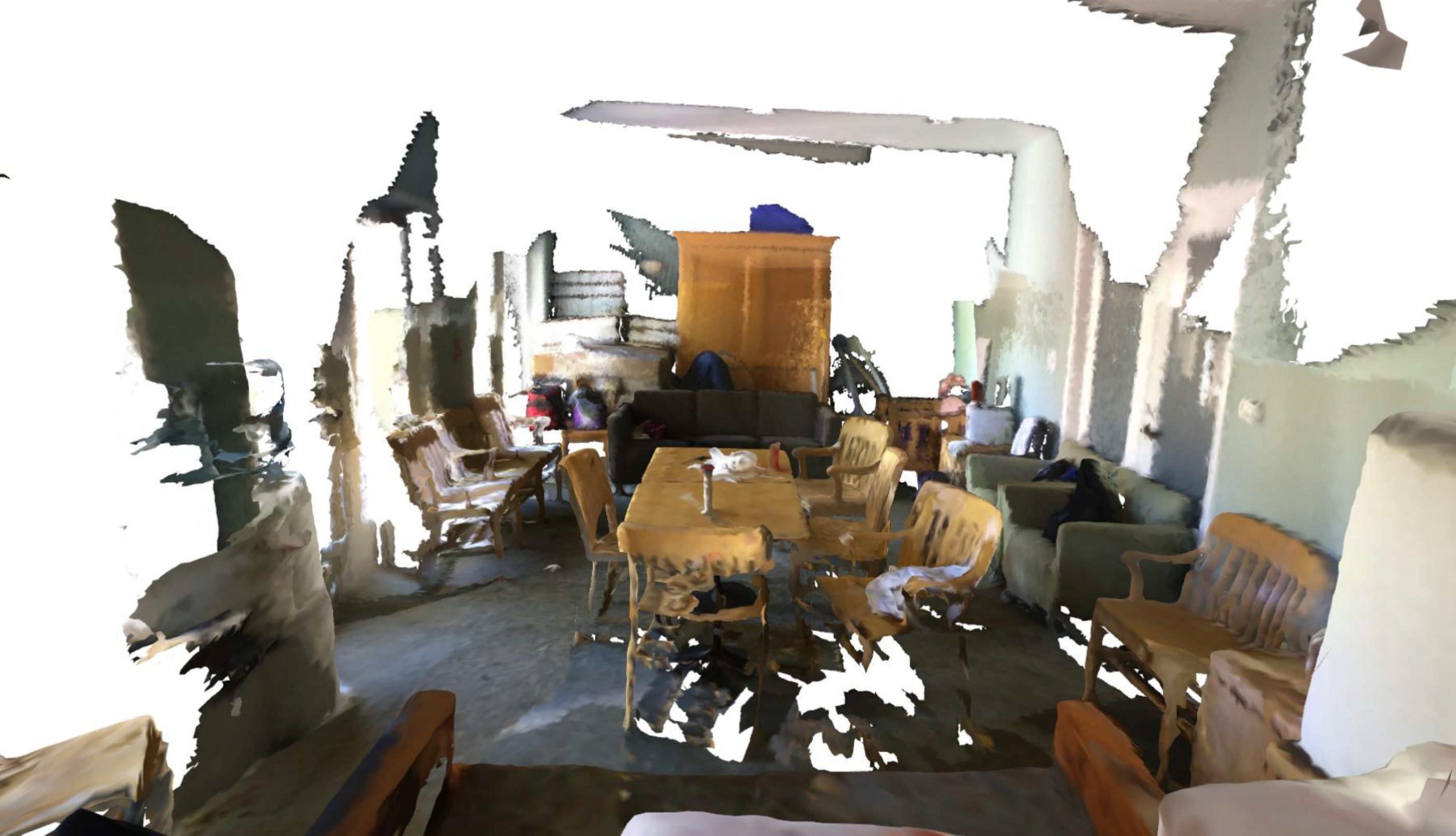}
      \end{subfigure}
      \begin{subfigure}{0.31\linewidth}
            \includegraphics[width=1\textwidth]{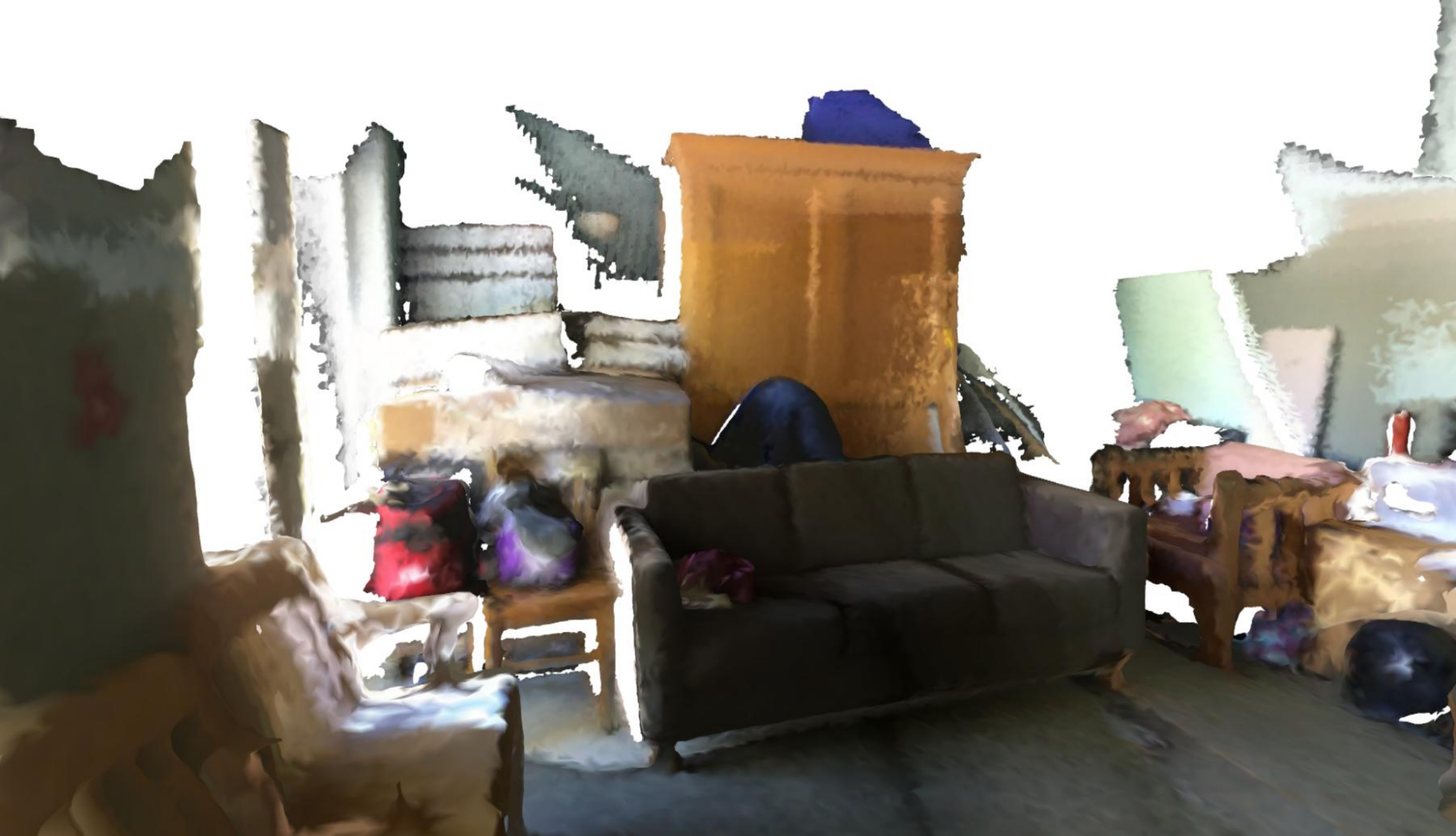}
      \end{subfigure}
      \begin{subfigure}{0.31\linewidth}
            \includegraphics[width=1\textwidth]{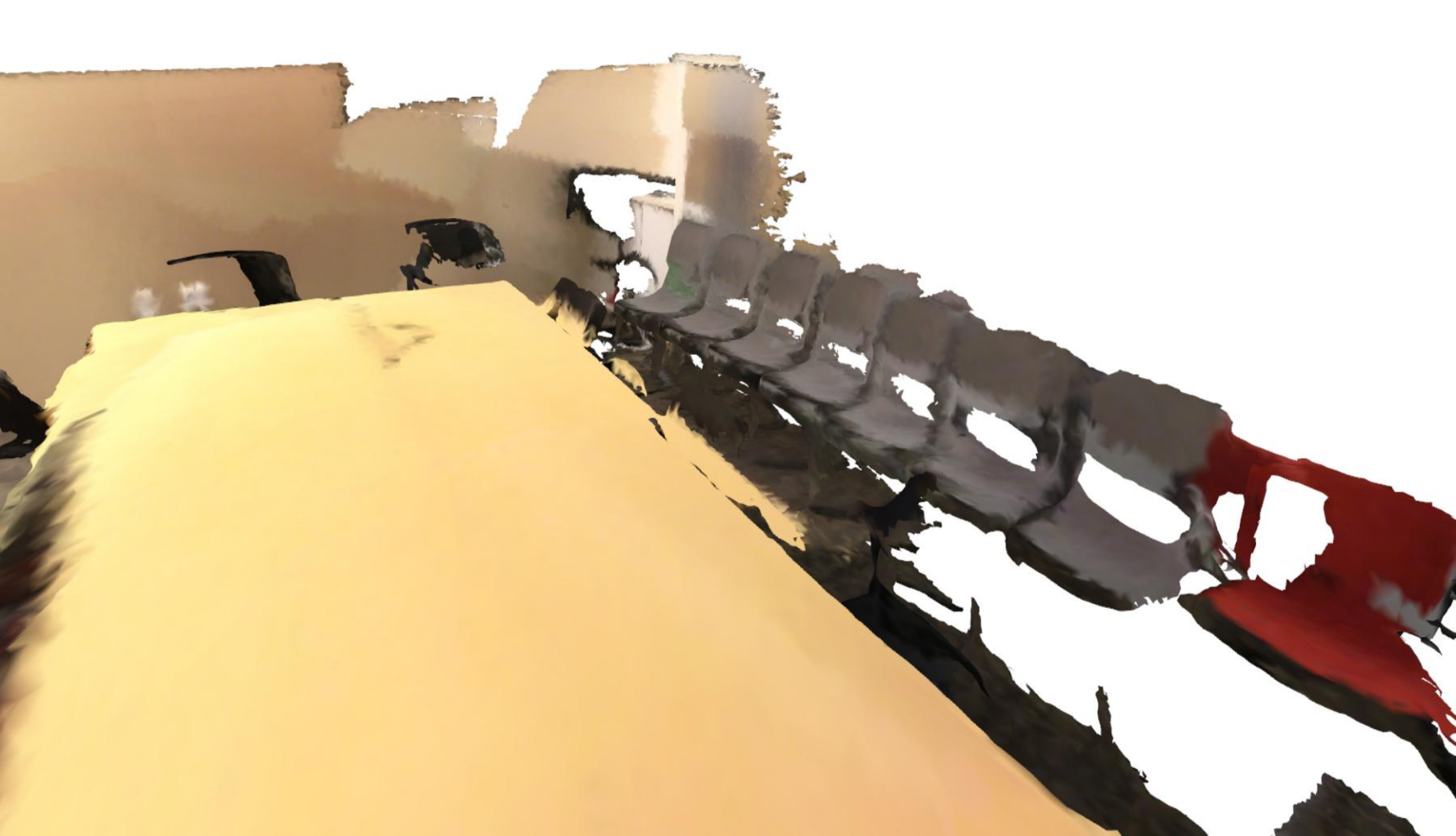}
      \end{subfigure}

      \caption{The qualitative results on ScanNet\cite{dai2017scannet} demonstrate that our method can reconstruct complex indoor scenes, including oversized items such as sofas, tables, chairs, and cabinets. Additionally, it produces high-quality reconstructions of small items like backpacks and other objects within the indoor scene.}
       \label{fig:scan_2}
\end{figure*}

\begin{figure*}[!t]
      \centering
      \captionsetup[subfigure]{labelformat=empty}

      \rotatebox[origin=lb]{90}{\hspace{10mm} \scriptsize{Co-SLAM\cite{wang2023coslam}}}
      \begin{subfigure}{0.46\linewidth}
            \centering
            \captionsetup{justification=raggedright}
            \includegraphics[width=1\textwidth]{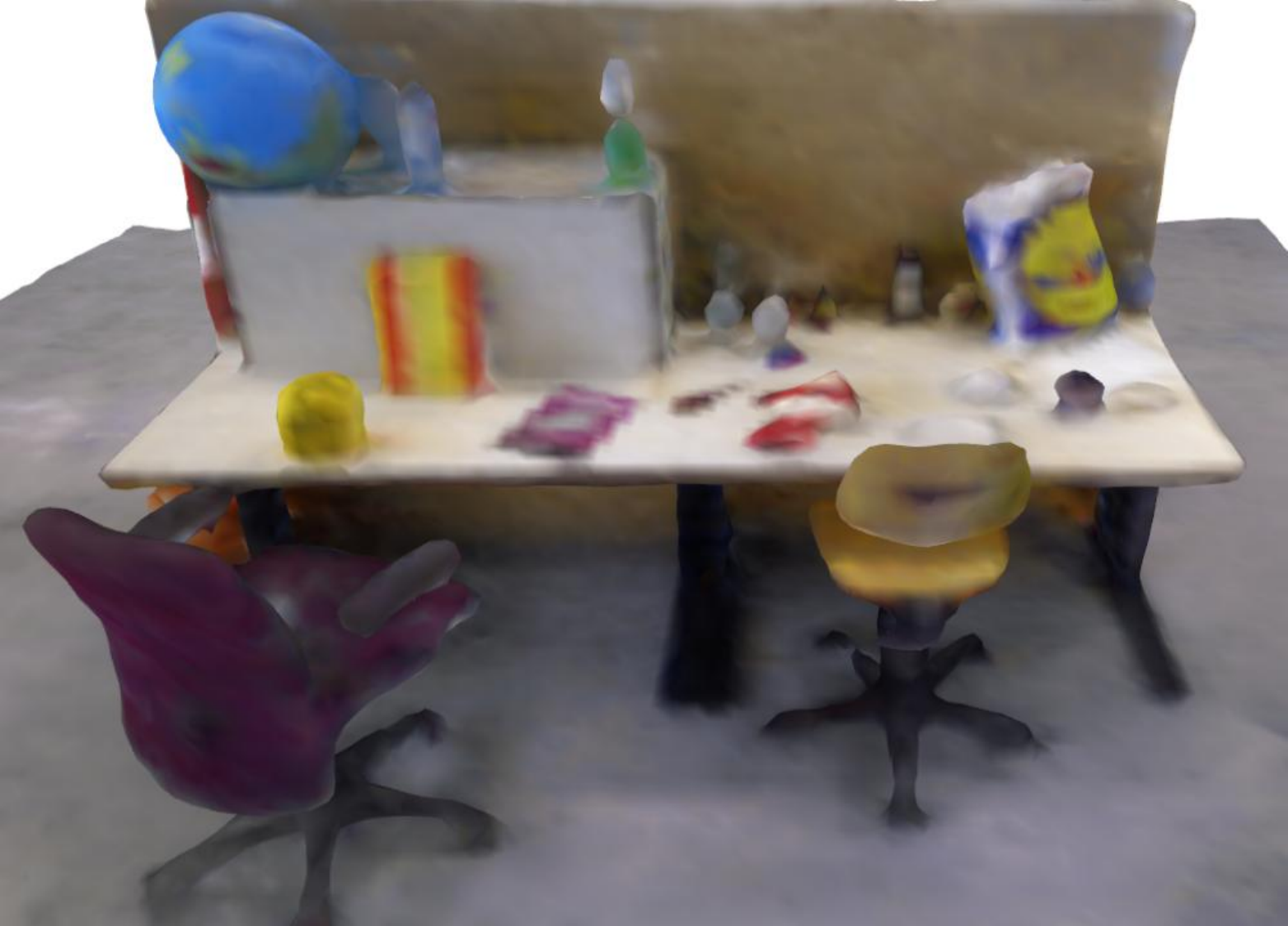}
      \end{subfigure}
      \begin{subfigure}{0.46\linewidth}
            \centering
            \captionsetup{justification=raggedright}
            \includegraphics[width=1\textwidth]{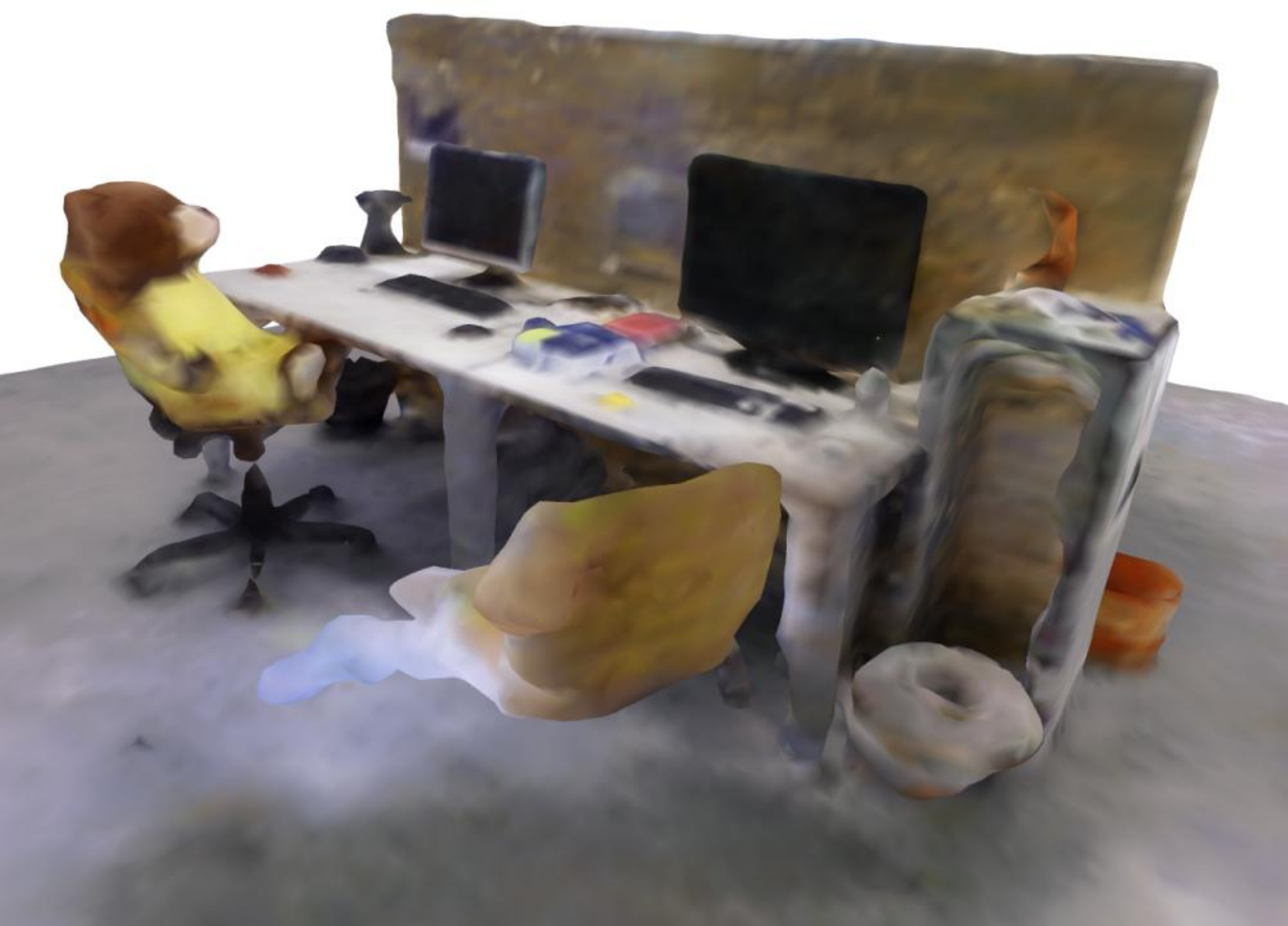}
      \end{subfigure}

      \rotatebox[origin=lb]{90}{\hspace{9mm} \scriptsize{Point-SLAM\cite{sandstrom2023pointslam}}}
      \begin{subfigure}{0.46\linewidth}
            \includegraphics[width=1\textwidth]{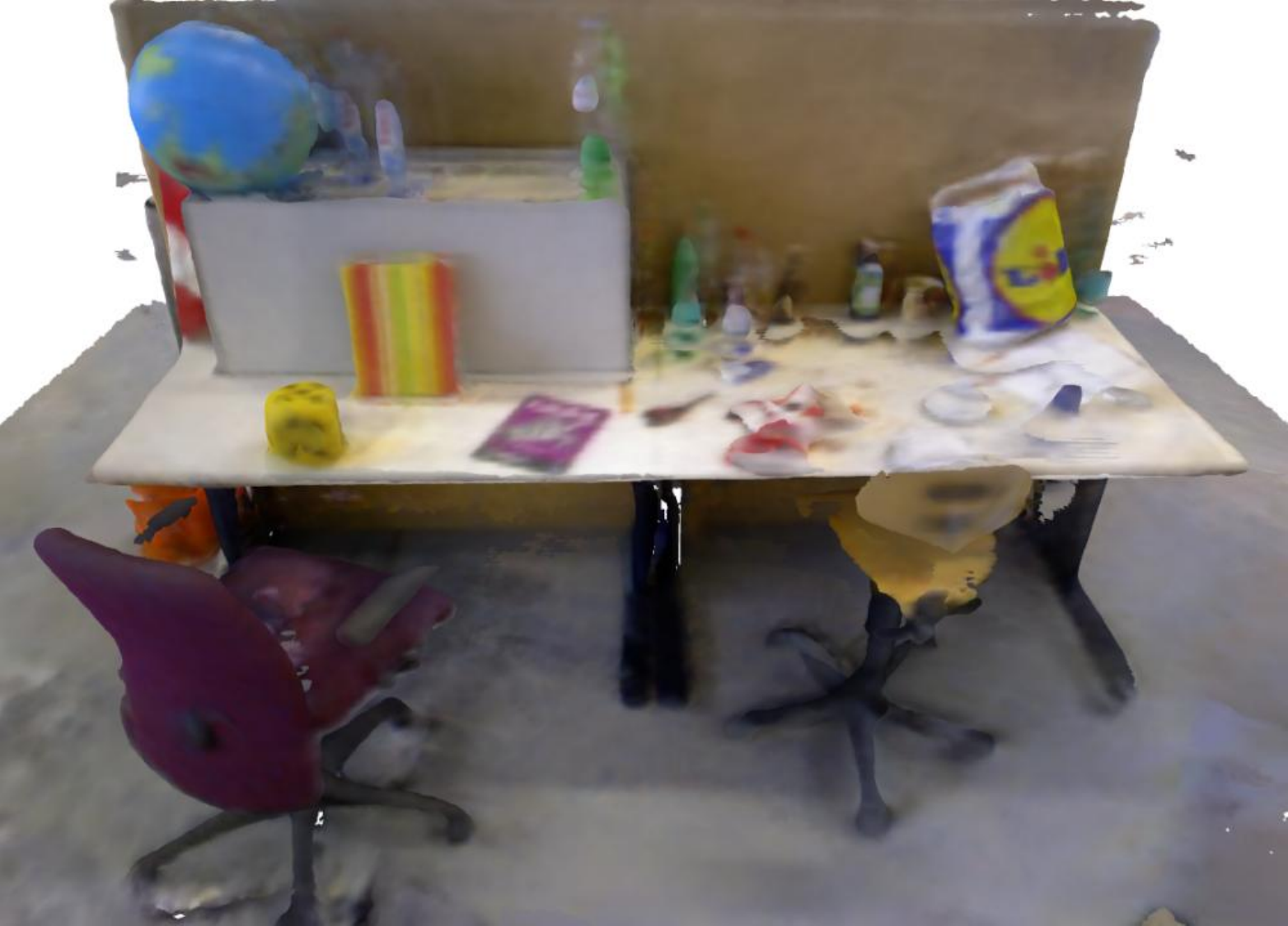}
      \end{subfigure}
      \begin{subfigure}{0.46\linewidth}
            \includegraphics[width=1\textwidth]{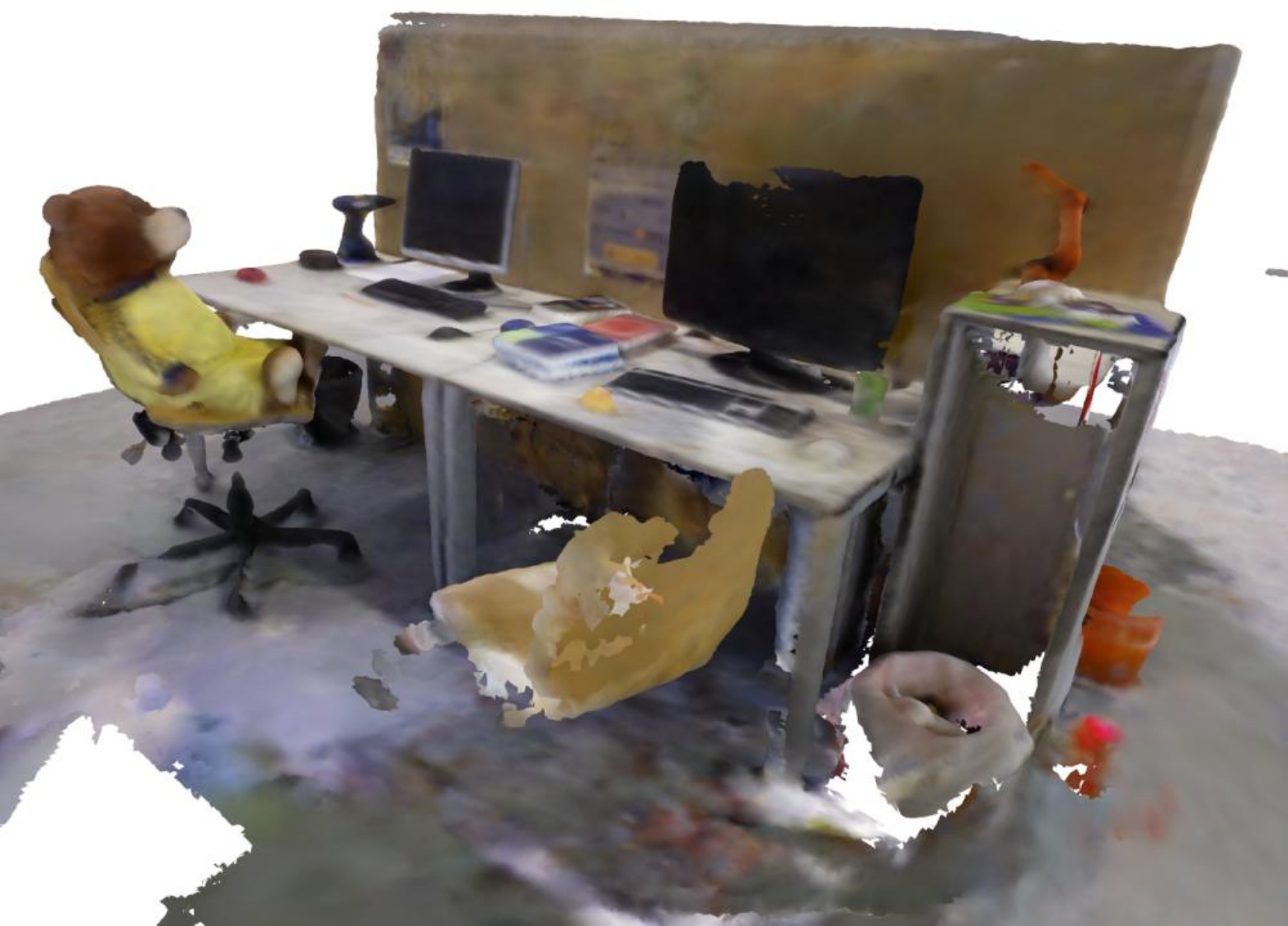}
      \end{subfigure}
     
      \rotatebox[origin=lb]{90}{\hspace{16mm} \footnotesize{Ours}}
      \begin{subfigure}{0.46\linewidth}
            \includegraphics[width=1\textwidth]{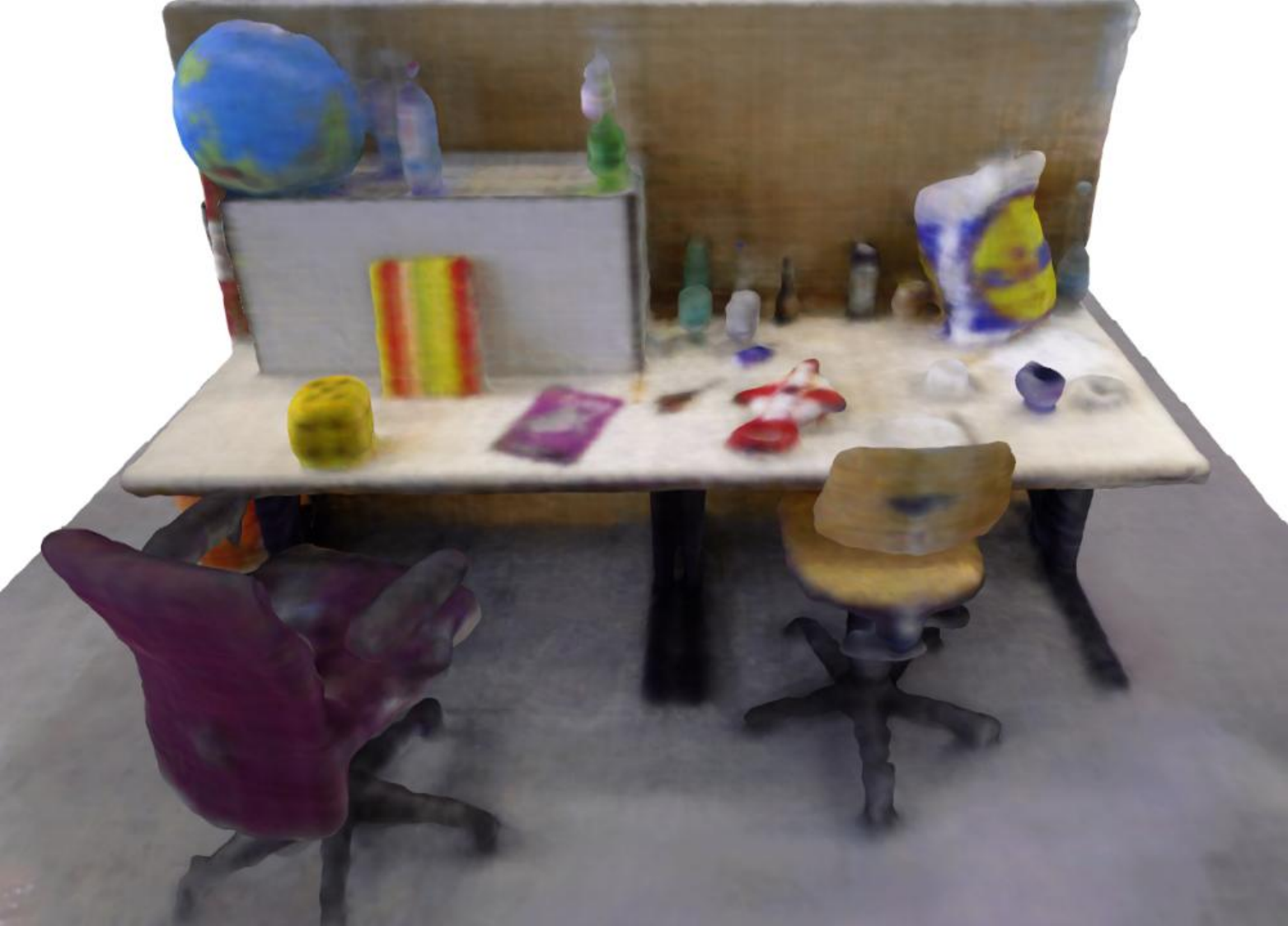}
      \end{subfigure}
      \begin{subfigure}{0.46\linewidth}
            \includegraphics[width=1\textwidth]{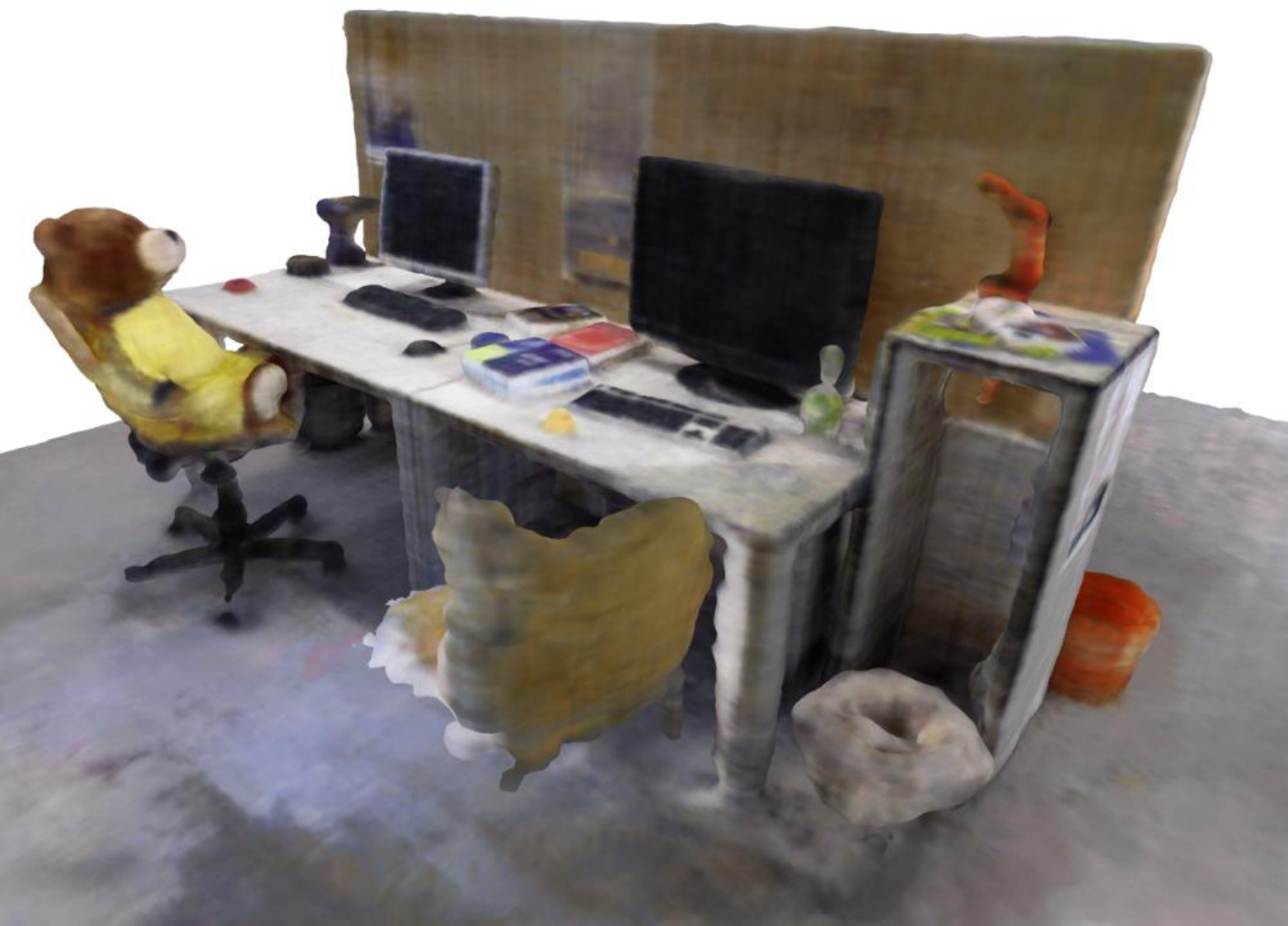}
      \end{subfigure}

      \caption{The qualitative results on TUM\cite{sturm12tum} demonstrate that our method can reconstruct high-quality tabletop-level scenes, including sharp edges of items such as keyboards, bottles, trash cans, and computers. Compared to other excellent methods, our reconstructed mesh is highly competitive.}
       \label{fig:tum_1}
\end{figure*}

\vspace{+4mm}
\noindent\noindent
\textit{\textbf{Qualitative Experimentation Supplement.}}
\indent\indent
To further illustrate the effectiveness of our method, as shown from \cref{fig:replica_1} to \cref{fig:tum_1} , we conduct additional qualitative experiments comparing our approach with various other methods, including NICE-SLAM\cite{zhu2022nice}, Co-SLAM\cite{wang2023coslam} and Point-SLAM\cite{sandstrom2023pointslam}.
ESLAM\cite{johari2023eslam} employs tri-planes, achieving high-quality geometric and appearance reconstruction. However, ESLAM fails to achieve a more compact representation for the tri-planes, so the parameter count remains relatively high, especially at high resolutions. Our approach adopts 2D hash grids to sparsify the parameters of the planes, aiming to reduce the overall parameter count. To further substantiate the objectivity of our parameter reduction, we conduct comparative experiments with ESLAM on Replica\cite{straub2019replica} and ScanNet\cite{dai2017scannet}.

\clearpage

\par\vfill\par

%
%
\end{document}